\documentclass{article}

 \usepackage[preprint]{neurips_2026}

\usepackage[utf8]{inputenc} 
\usepackage[T1]{fontenc}    
\usepackage{hyperref}       
\usepackage{url}            
\usepackage{booktabs}       
\usepackage{amsfonts}       
\usepackage{nicefrac}       
\usepackage{microtype}      
\usepackage{xcolor}         

\usepackage{multirow}
\usepackage[table]{xcolor}
\usepackage{pifont}
\usepackage{colortbl}
\usepackage{graphicx}
\usepackage{makecell}
\usepackage{subcaption}
\usepackage{amsmath}
\usepackage{algorithm}
\usepackage{algpseudocode}

\definecolor{oursbg}{RGB}{237,244,250}
\definecolor{avgbg}{RGB}{247,240,226}
\definecolor{posgreen}{RGB}{0,120,40}
\definecolor{negred}{RGB}{180,30,30}

\definecolor{oursrow}{RGB}{232,241,255}
\definecolor{deltarow}{RGB}{239,249,243}
\definecolor{posgreen}{RGB}{0,120,70}
\definecolor{negred}{RGB}{175,35,35}

\newcommand{\best}[1]{\textbf{#1}}
\newcommand{\pos}[1]{\textcolor{posgreen}{#1}}
\newcommand{\negv}[1]{\textcolor{negred}{#1}}

\newcommand{\cmark}{\ding{51}}
\newcommand{\xmark}{\ding{55}}
\newcommand{\yes}{\textcolor{posgreen}{\cmark}}
\newcommand{\no}{\textcolor{negred}{\xmark}}

\title{SD-LoRA: Training-Time Structural Distillation into LoRA for Few-Shot Vision-Language Adaptation}

\author{%
  Mohammed Rahman Sherif Khan Mohammad \\
  Edge Hill University, UK \\
  \texttt{Mohamm@edgehill.ac.uk}
  \And
  Ardhendu Behera \thanks{Corresponding author.}\\   
  Edge Hill University, UK \\
  \texttt{Beheraa@edgehill.ac.uk}
  \AND
  Sandip Pradhan \\
  Edge Hill University, UK \\
  \texttt{Pradhans@edgehill.ac.uk}
  \And
  Swagat Kumar \\
  Edge Hill University, UK \\
  \texttt{Kumars@edgehill.ac.uk}
  \And
  Yonghuai Liu \\
  Edge Hill University, UK \\
  \texttt{Liuyo@edgehill.ac.uk}
}

\begin{document}

\maketitle

\begin{abstract}
Few-shot adaptation of vision--language models must balance two competing goals: learning fine-grained visual distinctions from very limited supervision, and preserving the efficiency that makes frozen CLIP-like models attractive for deployment. Existing lightweight adaptation methods, including prompt tuning, LoRA, and cache-based adapters, typically supervise global image embeddings and therefore underuse token-level structure. Conversely, structure-aware methods can exploit local evidence but often introduce additional inference-time computation. We propose \textit{SD-LoRA}, a structural distillation framework that uses token-level graph reasoning only during training and absorbs its effect into a lightweight LoRA-adapted CLIP model. During training, a heterogeneous graph teacher models patch--patch, global--local, and visual--text relations over high-resolution ViT tokens. Rather than keeping this graph at test time, SD-LoRA transfers its structural bias into the shared CLIP-LoRA backbone through implicit distillation and Prototype Predictive Alignment, which aligns graph-conditioned representations with support-derived class prototypes. At inference, the entire graph teacher is discarded; prediction uses only a single CLIP-LoRA forward pass and a cache-based classifier. Across 11 few-shot benchmarks, SD-LoRA improves over a strong CLIP-LoRA baseline by 2.4/1.8/3.0/2.4/2.0 percentage points in the 1/2/4/8/16-shot settings, while adding no graph computation at test time. Further results on base-to-novel generalization, cross-dataset transfer, OOD robustness, and calibration show that training-time structural supervision improves not only accuracy but also transfer reliability. Code is available at \url{https://github.com/MR-Sherif/SD-LoRA.git}.

\end{abstract}

\section{Introduction}
\label{sec:intro}

\begin{figure}[t]
    \centering
    \includegraphics[width=\linewidth]{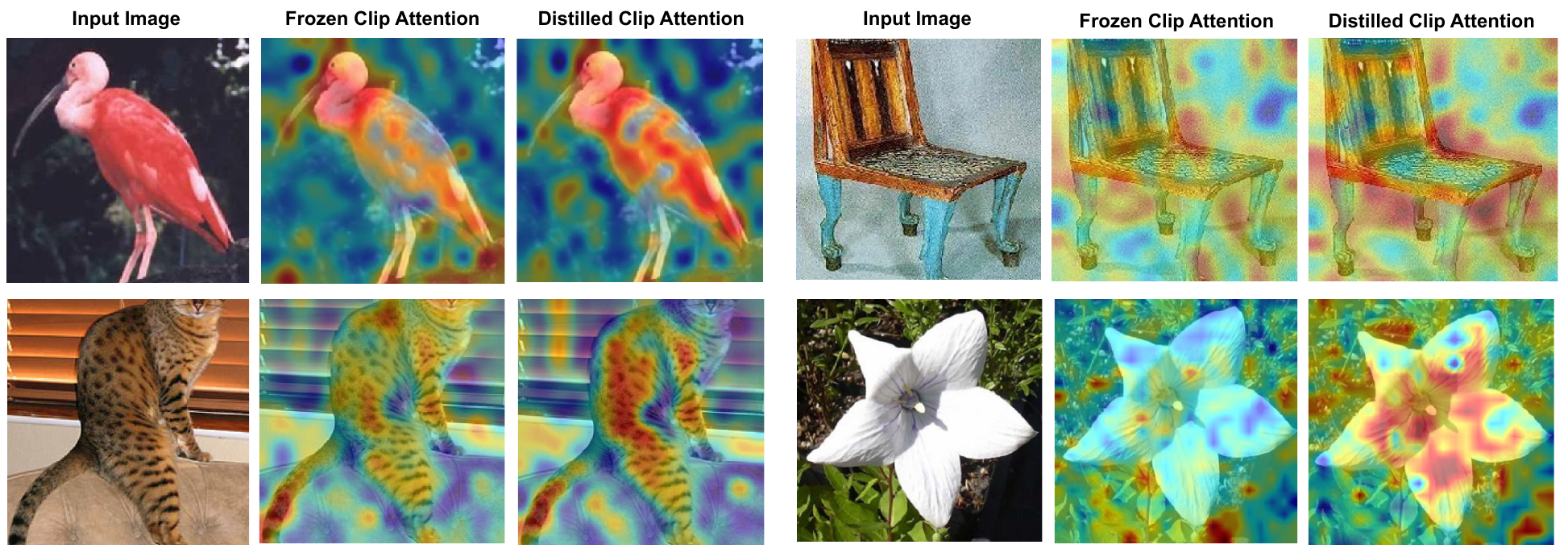}
    \caption{
        \textbf{Structural distillation sharpens token-level evidence.}
We visualize patch-level attention maps for frozen CLIP and SD-LoRA using cosine similarity between the final \texttt{[CLS]} token and patch tokens. Frozen CLIP attends strongly to background foliage and misclassifies the image as \emph{clematis}, while SD-LoRA shifts focus toward discriminative flower regions and correctly predicts \emph{balloon flower}. The graph teacher is used only during training; the visualization is produced from the deployed LoRA backbone without test-time graph computation.
    }
    \label{fig:Spatial_Heatmap_abl}
    \vspace{-1.0em}
\end{figure}


Vision--language models (VLMs) such as CLIP~\cite{radford2021learningclip} and ALIGN~\cite{jia2021scaling} are now standard foundations for open-vocabulary and few-shot recognition. Their strength comes from large-scale image--text pretraining, which produces transferable global representations and strong zero-shot performance. However, this same global representation that makes CLIP efficient can become a limitation when only a few labeled examples are available. In many few-shot tasks, especially fine-grained recognition, class distinctions depend not just on coarse semantic alignment but on subtle local cues such as texture, part shape, object configuration, or small discriminative regions.

Recent CLIP adaptation methods advance this setting along several axes. Prompt-learning adapts CLIP’s text interface~\cite{zhou2022learningcoop,zhou2022conditionalCoCoOp,khattak2023maple,khattak2023self}, distillation-based prompts improve transfer from stronger teachers~\cite{li2024promptkd}, and base-to-novel approaches explicitly manage the tradeoff between fitting base classes and preserving novel-class generalization~\cite{li2025dpc}. In parallel, adapter and cache-based methods such as CLIP-Adapter and Tip-Adapter~\cite{gao2024clip-adapter,zhang2022tip}, along with recent kernel views of cache adaptation~\cite{bendou2025proker}, provide lightweight alternatives to full backbone fine-tuning. LoRA-based adaptation likewise achieves parameter efficiency by updating only low-rank components~\cite{hu2022lora,zanella2024lowclip-lora}. Most methods still use a single global image embedding for supervision or retrieval, which improves efficiency but does not explicitly model interactions among local visual tokens or between local tokens and class semantics.

This creates a \emph{pooled-supervision bottleneck}: the adapted model maps an image to the right coarse region of CLIP's embedding space, but neglects local token interactions needed to separate fine-grained classes. Although CLIP's ViT backbone encodes dense token-level information, standard adaptation collapses these tokens into a pooled feature before supervision, weakening or discarding cues such as part shape, texture, and spatial layout. Region-level alignment, dense prediction, multi-crop reasoning, or graph structure~\cite{rao2022denseclip,zhong2022regionclip,li2022grounded,ahmad2025tip} can recover such cues, but usually require extra detectors, dense computation, multiple crops, or graph modules at inference. Consequently, existing methods fall into two regimes: lightweight approaches that adapt global representations but underuse structure, and structure-aware approaches that better exploit local evidence but increase deployment cost.

This paper asks a simple question: \emph{can token-level structural reasoning improve few-shot CLIP adaptation without being used at test time?} We answer this via proposing \textbf{SD-LoRA}, a structural distillation framework for few-shot vision--language adaptation. SD-LoRA uses graph reasoning only as \emph{training-time supervision}, not for inference. During training, it builds a heterogeneous graph over high-resolution visual tokens and class-text embeddings, modeling patch--patch, global--local, and visual--text relations. The teacher is optimized to reveal a discriminative local structure that is never deployed. Its structural signal is distilled into a shared CLIP-LoRA backbone, so the final model is just a lightweight LoRA-adapted CLIP encoder with a cache-based classifier.

The core idea is \textbf{implicit structural distillation}. Instead of conventional logit-based distillation~\cite{hinton2015distillingkd}, SD-LoRA transfers structure via shared-backbone optimization: the graph teacher and the deployed student use the same LoRA-adapted visual backbone, so gradients from the graph objective directly shape the low-rank updates used in inference. To stabilize this in the low-shot regime, we introduce \textbf{Prototype Predictive Alignment} (PPA), aligning graph-conditioned representations with support-based class prototypes, pushing the LoRA-updated features to form compact, structure-aware class regions instead of overfitting to individual support examples.
Unlike recent transductive/test-time adaptation methods that rely on unlabeled test batches or update the model during deployment~\cite{zanella2024boosting,zanella2025realistic,yoon2024ctpt}, we target the inductive few-shot setting: after training in the support set, it performs single-image inference, with no test-time graph construction, unlabeled-batch assumptions or online adaptation. 


%
\noindent\textbf{Contributions: }
(1) We propose \textbf{SD-LoRA}, a training-time structural distillation framework that injects token-level graph reasoning into a CLIP-LoRA backbone while keeping inference as lightweight LoRA$+$cache adaptation.
(2) We introduce a heterogeneous graph teacher over ViT tokens and class-text embeddings, enabling patch--patch, global--local, and visual--text relational supervision without test-time graph computation.
(3) We introduce Prototype Predictive Alignment as a complementary regularizer of graph-conditioned supervision, boosting few-shot accuracy, base-to-novel generalization, cross-dataset transfer, and OOD robustness in the same lightweight inference pathway. 
\section{Related Work}
\label{sec:related-works}
\vspace{-0.3em}
\textbf{Global adaptation of CLIP representations.}
Most few-shot CLIP adaptation methods tweak text prompts, image features, or the classifier while keeping the pretrained model fixed. Prompt-learning approaches like CoOp and CoCoOp~\cite{zhou2022learningcoop,zhou2022conditionalCoCoOp} replace hand-crafted templates with learnable context tokens, and MaPLe~\cite{khattak2023maple} extends prompting to both visual and textual branches. Later work enhances generalization via regularization, prompt distillation, or base-to-novel balancing, including PromptSRC~\cite{khattak2023self}, CoPrompt~\cite{roy2024coprompt}, PromptKD~\cite{li2024promptkd}, and DPC~\cite{li2025dpc}. Adapter- and cache-based methods instead modify the visual representation or decision rule: CLIP-Adapter~\cite{gao2024clip-adapter} adds residual adapters, Tip-Adapter~\cite{zhang2022tip} builds a support-set key--value cache, and ProKeR~\cite{bendou2025proker} interprets cache adaptation as kernel regularization. These methods provide strong lightweight adaptation but operate mostly on image-level embeddings, prompt logits, or support-set similarities. SD-LoRA is complementary: it remains lightweight but changes the training signal by adding token-level structural supervision.

\textbf{LoRA and parameter-efficient VLM adaptation.}
Parameter-efficient fine-tuning updates only a small subset of parameters, keeping most of the pretrained model fixed. LoRA~\cite{hu2022lora} does this via low-rank updates to selected linear layers, and CLIP-LoRA~\cite{zanella2024lowclip-lora} shows it is a strong few-shot CLIP baseline. Robustness-focused extensions like AdvCLIP-LoRA~\cite{ghiasvand2025few} study LoRA under distribution shift. SD-LoRA uses LoRA differently: LoRA is not the main novelty but the target of structural supervision. The graph teacher and student share the same LoRA-adapted backbone, so teacher-side relational losses directly shape the low-rank updates that will later be used without the teacher, unlike standard LoRA, where low-rank parameters are trained mainly from global classification losses.

\textbf{Region-, token-, and graph-level reasoning.}
Another research line augments CLIP with localized or structured visual evidence. DenseCLIP~\cite{rao2022denseclip} targets dense prediction, while RegionCLIP~\cite{zhong2022regionclip} and GLIP-style models~\cite{li2022grounded} align text with object regions. Patch- and crop-based methods further use localized views to refine predictions~\cite{ahmad2025tip}. Graph-based approaches such as HeGraphAdapter~\cite{zhao2024hegraphadapter} and GraphAdapter~\cite{li2023graphadapter} encode relations among visual, textual, or class-level nodes. TOGA~\cite{Mohammad_2026_CVPR} uses a training-only heterogeneous image--patch--text graph teacher to supervise a cache-based adapter while keeping CLIP frozen. SD-LoRA instead transfers structural supervision into shared LoRA parameters using high-resolution ViT tokens, while likewise discarding the graph teacher at inference.

\textbf{Distillation, transduction, and test-time adaptation.}
Knowledge distillation transfers information from a teacher model to a student~\cite{hinton2015distillingkd}. In VLM adaptation, PromptKD~\cite{li2024promptkd} distills prompt knowledge from a larger CLIP teacher. SD-LoRA differs by distilling structure rather than only predictions: the teacher is a token-level graph reasoner, and the transfer occurs implicitly through shared LoRA parameters and Prototype Predictive Alignment. Our setting is also distinct from transductive or test-time adaptation methods such as TransCLIP~\cite{zanella2024boosting}, C-TPT~\cite{yoon2024ctpt}, and StatA~\cite{zanella2025realistic}, which exploit unlabeled test batches or update the model during deployment. SD-LoRA considers the standard inductive few-shot protocol: after support-set adaptation, each test image is processed independently with a fixed LoRA$+$cache model. Following recent concerns about standard CLIP few-shot evaluation~\cite{kravets2025rethinking}, we therefore report not only few-shot accuracy but also base-to-novel generalization, cross-dataset transfer, out-of-distribution (OOD) robustness, and calibration.

\begin{figure*}[t]
\centering
\includegraphics[width=0.99\linewidth]{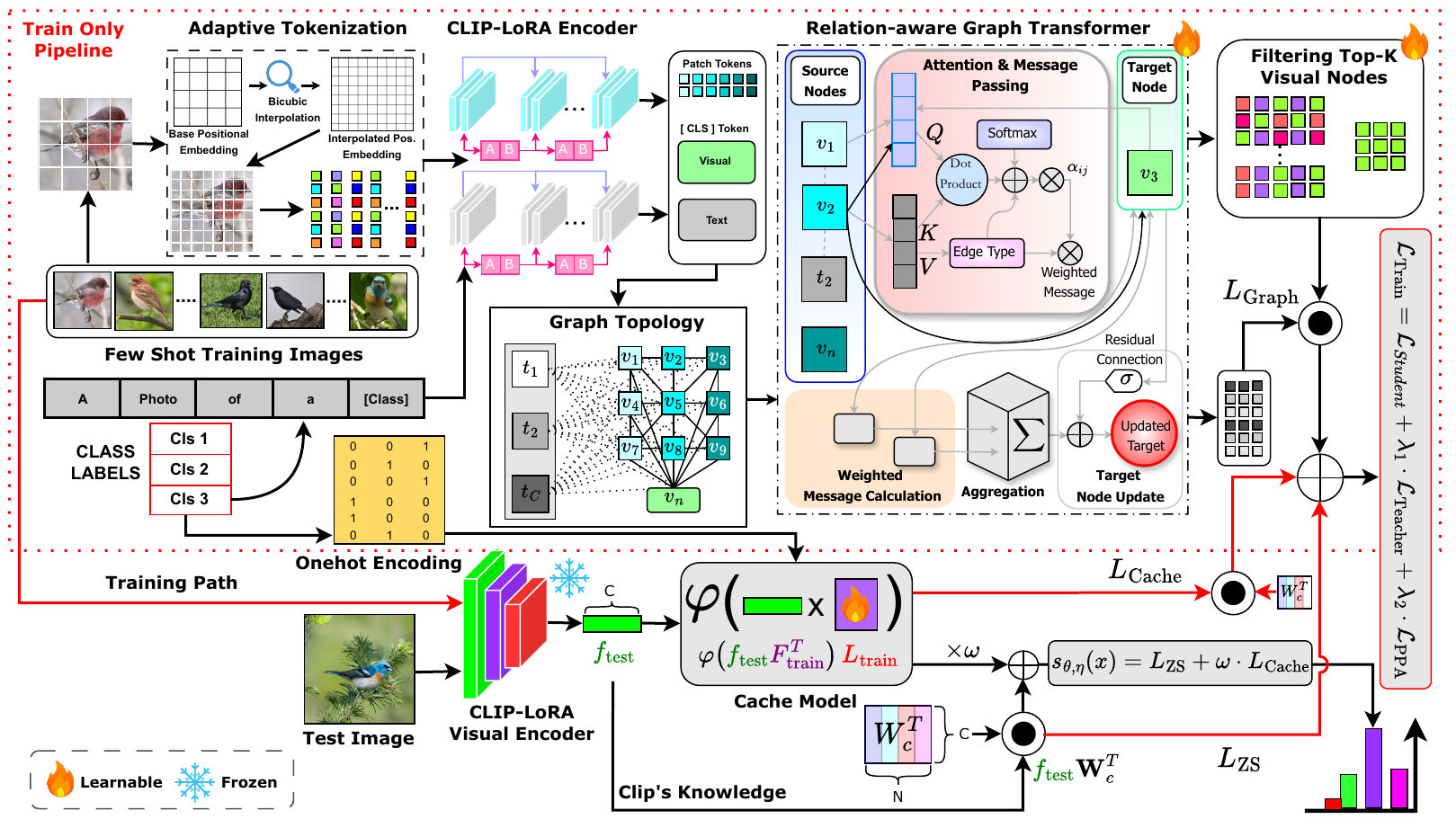}
\caption{
\textbf{SD-LoRA framework.}
It trains a deployable CLIP-LoRA$+$cache student $s_{\theta,\eta}$ using an auxiliary structural teacher $g_{\theta,\psi}$ present only during adaptation. Teacher and student share the LoRA backbone, but the teacher reasons over high-resolution ViT tokens and class-text embeddings with a relation-aware graph transformer. Structural supervision from the graph branch plus Prototype Predictive Alignment updates the shared LoRA parameters $\theta$. At inference, the teacher parameters $\psi$ and graph computation are discarded, and only the lightweight student $s_{\theta,\eta}$ makes predictions.
}
\label{fig:architecture}
\vspace{-4mm}
\end{figure*}

\vspace{-0.5em}\section{Structural Distillation into Low-Rank Adaptation}
\label{sec:method}\vspace{-0.5em}
We formulate \textbf{SD-LoRA} as learning a deployable low-rank predictor within a richer training-time hypothesis class. The inference model $s_{\theta,\eta}$ combines a LoRA-adapted CLIP image encoder with parameters $\theta$ and a cache head with parameters $\eta$. During training, we introduce a teacher $g_{\theta,\psi}$ that shares the LoRA backbone but incorporates graph-reasoning parameters $\psi$ over visual tokens and class-text embeddings. Student and teacher share $\theta$: student losses optimize deployment-time discrimination, while teacher losses enforce structural constraints on the representation that persist at inference. After training, we discard $\psi$ and use only $s_{\theta,\eta}$ for prediction.

\vspace{-0.25em}\subsection{Deployable Student and Training-Time Teacher}
\label{sec:student_teacher}\vspace{-0.25em}

We consider $C$-way $K$-shot few-shot classification problem with support set
$\mathcal{S}=\{(x_i,y_i)\}_{i=1}^{CK}$, where each of the $C$ classes has $K$ labeled support examples from the training split following standard few-shot CLIP adaptation protocols~\cite{zhou2022learningcoop,zhang2022tip,zanella2024lowclip-lora}. For each class $c$, we construct a prompt $\tau_c$ and compute the corresponding CLIP text embedding $t_c=f_t(\tau_c)$, giving the class-text set $T=\{t_c\}_{c=1}^{C}$.

The deployed student $s_{\theta,\eta}$ uses a LoRA-adapted CLIP image encoder. Starting from pretrained CLIP~\cite{radford2021learningclip}, the original image encoder weights are frozen and only low-rank updates are trained. For a pretrained weight matrix $W_0$, LoRA defines
\(
    W = W_0 + \Delta W, 
    \ 
    \Delta W = BA,
\)
with $A$ and $B$ low-rank trainable matrices. 
Let $f^\theta_v:\mathcal{X}\rightarrow\mathbb{R}^d$ be the pooled image embedding from the LoRA-adapted CLIP visual encoder, and define the normalized feature
\(
z_\theta(x)=f^\theta_v(x)/\|f^\theta_v(x)\|_2.
\)
The zero-shot CLIP logits are
\(
L_{\mathrm{ZS}}(x)=\mathrm{sim}(z_{\theta}(x),T),
\)
the cosine similarities between the image feature and all class-text embeddings.  
To use the few-shot support set at deployment, the student adds a cache-based head: support features give cache keys
\(
a_i^{\mathrm{cache}} = z_{\theta}(x_i),
\)
which form the rows of $A_{\mathrm{cache}}\in\mathbb R^{CK\times d}$ for $CK$ support examples ($K$ shots per class), and the corresponding one-hot labels give cache values $V\in\mathbb{R}^{CK\times C}$. 
Given a query feature $z_{\theta}(x)$, the cache head computes support affinities
\(
    a_{\eta}(x)=
    \exp\!\left[-\beta\big(1-\mathcal{A}_{\eta}(z_{\theta}(x),A_{\mathrm{cache}})\big)\right],
\)
where $\mathcal{A}_{\eta}$ is a learnable cache matcher, $\beta$ controls affinity sharpness, and $a_{\eta}(x)\in\mathbb{R}^{CK}$ gives one score per support example. Multiplying these affinities by the one-hot cache values $V$ yields cache logits over the $C$ classes:
\begin{equation}
    L_{\mathrm{Cache}}(x)=a_{\eta}(x)^{\top}V, \qquad
    s_{\theta,\eta}(x)
    =
    L_{\mathrm{ZS}}(x)
    +
    \omega L_{\mathrm{Cache}}(x),
    \label{eq:student_logits}
\end{equation}
where $L_{\mathrm{Cache}}(x)\in\mathbb{R}^{C}$ is the cache-branch logit vector, $s_{\theta,\eta}(x)\in\mathbb{R}^{C}$ is the final deployed student logits, and $\omega$ controls the cache contribution. Eq.~\eqref{eq:student_logits} defines the inference-time student model.

During training, we introduce a structural teacher $g_{\theta,\psi}$ that, unlike the student operating on the pooled image representation $z_{\theta}(x)$, processes the full token sequence from the same LoRA-adapted visual backbone,
\(
    \mathcal{U}_{\theta}(x)
    =
    \{u_{\mathrm{cls}},u_1,\ldots,u_M\}
\),
consisting of the \texttt{[CLS]} token $u_{\mathrm{cls}}$ and $M$ patch tokens. The teacher fuses these visual tokens with the class-text embeddings $T$ via graph reasoning to obtain a structure-aware representation $h_g$ and graph logits $L_{\mathrm{Graph}}$. Thus,
\begin{equation}
    g_{\theta,\psi}(x,T)
    =
    \big(h_g, L_{\mathrm{Graph}}\big).
    \label{eq:graph_teacher}
\end{equation}
The important point is that both $s_{\theta,\eta}$ and $g_{\theta,\psi}$
depend on the same LoRA parameters $\theta$. Consequently, structural
supervision applied to the teacher updates the representation used by the
deployed student.
\vspace{-0.25em}\subsection{Relation-Aware Structural Supervision}
\label{sec:structural_supervision}
\vspace{-0.25em}
The role of $g_{\theta,\psi}$ is to define a richer training-time hypothesis class that captures local token interactions and class semantics. To obtain fine-grained evidence, we extract ViT tokens at higher spatial resolution and interpolate the pretrained positional embeddings to the new token grid. Let $e_{\mathrm{cls}}$ be the pretrained positional embedding of the \texttt{[CLS]} token, and let $E_{\mathrm{pos}}^{\mathrm{patch}}$ be the pretrained patch positional embeddings. The adapted positional encoding is
\begin{equation}
    E'_{\mathrm{pos}}
    =
    \mathrm{Concat}
    \left(
    e_{\mathrm{cls}},
    \mathrm{Resize}_{\mathrm{bicubic}}
    (E_{\mathrm{pos}}^{\mathrm{patch}},(h',w'))
    \right),
\end{equation}
where $(h',w')$ is the new token grid size. This yields denser visual tokens in a single backbone forward pass, without multi-crop inference or external region proposals. Given the visual tokens $\mathcal{U}_{\theta}(x)$ and text embeddings $T$, the teacher builds a heterogeneous graph
\begin{equation}
    \mathcal{G}=(\mathcal{N},\mathcal{E}),
    \qquad
    \mathcal{N}=\mathcal{N}_{v}\cup\mathcal{N}_{t},
    \qquad
    \mathcal{N}_{v}=\{u_{\mathrm{cls}},u_1,\ldots,u_M\},
\end{equation}
$\mathcal{N}_{v}$ is the set of visual nodes and $\mathcal{N}_{t}=T$ the class-text nodes. The edge set includes three relation types: \emph{patch--patch} spatial edges between neighboring visual tokens, \emph{global--local} edges between the CLS and patch tokens, and \emph{visual--text} semantic edges between visual tokens and class-text nodes. These relations let the teacher jointly capture local visual consistency, image-level context, and class-conditioned token alignment in a single structural representation.

Patch–patch edges use 8-connected adjacency on the 2D ViT token grid: each patch token connects to horizontal, vertical, and diagonal neighbors within one grid step, and boundary tokens connect only to in-grid neighbors. Global–local edges are bidirectional between the \texttt{[CLS]} node and every patch node. Visual–text edges are bidirectional between visual nodes and class-text nodes. The graph is updated by \emph{relation-aware message passing}. Let $\phi(i)\in\{\mathrm{vis},\mathrm{text}\}$ be the type of node $i$, and let $r=\mathrm{rel}(j,i)$ be the relation type of edge $j\rightarrow i$. Let $f_i^{(l)}\in\mathbb{R}^{d}$ be the hidden state of node $i\in\mathcal{N}$ at layer $l$, with $f_i^{(0)}$ initialized from the corresponding visual-token or class-text embedding. 
The neighborhood of node $i$ is defined by its incoming edges, $\Gamma(i)=\{j\mid (j,i)\in\mathcal{E}\}$. In our heterogeneous graph, $\Gamma(i)$ may contain spatially adjacent patch nodes (patch–patch relations), the \texttt{[CLS]} node or patch nodes (global–local relations), and connected class-text nodes (visual–text semantic relations). Attention normalization is thus performed over neighbors of the target node, with the edge type $r=\mathrm{rel}(j,i)$ determining the relation-specific bias and message transformation.

For attention head $h$, we construct a query vector $\mathbf{q}_i^{(h)}$, a relation-conditioned key vector $\mathbf{k}_{j\rightarrow i}^{(h)}$, and a relation-conditioned value vector $\mathbf{v}_{j\rightarrow i}^{(h)}$ in a $d_h$-dimensional space, where $d_h = d/\mathcal{H}$ for $\mathcal{H}$ heads. The unnormalized attention score $e_{j\rightarrow i}^{(h)}$, normalized attention weight $\alpha_{j\rightarrow i}^{(h)}$, and aggregated message $m_i^{(h)}$ are then computed as
\begin{equation}
e_{j\rightarrow i}^{(h)}
=
\frac{
(\mathbf{q}_i^{(h)})^{\top}
\mathbf{k}_{j\rightarrow i}^{(h)}
}{\sqrt{d_h}}
+
b_{r}^{(h)},
\quad
\alpha_{j\rightarrow i}^{(h)}
=
\rho_{j\in\Gamma(i)}
\left(e_{j\rightarrow i}^{(h)}\right),
\quad
m_i^{(h)}
=
\sum_{j\in\Gamma(i)}
\mu_{r}^{(h)}
\alpha_{j\rightarrow i}^{(h)}
\mathbf{v}_{j\rightarrow i}^{(h)} .
\label{eq:relation_attention}
\end{equation}
Here, $e_{j\rightarrow i}^{(h)}$ is the relation-aware compatibility score, $\rho$ is the softmax over neighbors $\Gamma(i)$, $b_r^{(h)}$ is the relation-dependent attention bias, and $\mu_r^{(h)}$ is the relation-specific message scaling factor. 
The concatenated multi-head message is projected back to $\mathbb{R}^{d}$ by a type-specific learnable output matrix $W_O^{(\phi(i))}\in\mathbb{R}^{\mathcal{H}d_h\times d}$. We use separate output projections for visual and text nodes, and optimize these matrices as part of the graph-teacher parameters $\psi$, giving

\begin{equation}
f_i^{(l+1)}
=
\mathrm{LN}
\left(
f_i^{(l)}
+
\mathrm{FFN}^{(\phi(i))}
\left(
\mathrm{Concat}_{h=1}^{\mathcal{H}}(m_i^{(h)})
W_O^{(\phi(i))}
\right)
\right),
\label{eq:relation_message_passing}
\end{equation}
Here, $\mathrm{LN}$ is layer normalization and $\mathrm{FFN}^{(\phi(i))}$ is a type-specific feed-forward network. Thus, $\phi$ controls modality-specific transformations and $r$ defines relation-specific interaction patterns, allowing the teacher to assign distinct roles to \emph{patch--patch}, \emph{global--local}, and \emph{visual--text} evidence. After $L$ message-passing layers, the teacher obtains updated visual and text nodes. 
Instead of averaging all patch tokens, which mixes informative and background regions, we use a learned token-selection readout. Here, “saliency’’ is the scalar importance score predicted by the graph teacher for each refined patch node, not an external saliency annotation. Each refined patch node $f_i$ is assigned a score
\(
\sigma_i = w_s^\top f_i,\ w_s\in\mathbb{R}^d,
\)
where $w_s$ is a learnable scoring vector included in the teacher parameters $\psi$. We keep the top $\lceil \kappa M\rceil$ patch nodes ranked by $\sigma_i$, denoted $\mathcal{N}_{\mathrm{top}}$, where $0<\kappa\leq 1$ is a token-retention ratio tuned on validation. These selected nodes, together with the refined \texttt{[CLS]} token, form the graph representation used to compute teacher logits by comparing it with the refined text nodes:

\begin{equation}
    h_g
    =
    f_{\mathrm{cls}}
    +
    \frac{1}{|\mathcal{N}_{\mathrm{top}}|}
    \sum_{f_j\in\mathcal{N}_{\mathrm{top}}} f_j,
   \qquad
    L_{\mathrm{Graph}}(x)
    =
    \mathrm{sim}(h_g,F_t),
    \label{eq:structural_readout}
\end{equation}
where $F_t$ denotes the graph-refined class-text embeddings. The teacher, therefore, provides class-discriminative supervision based on token-level visual evidence and visual--text relations.

\subsection{Implicit Distillation Objective}
\label{sec:implicit_distillation}
\vspace{-.5em}
The teacher is not distilled into the student by matching logits. Instead, it performs \emph{implicit structural distillation} via the shared LoRA backbone: because both the deployed student $s_{\theta,\eta}$ and structural teacher $g_{\theta,\psi}$ depend on $\theta$, teacher-side losses directly update the low-rank inference parameters.
\begin{equation}
\frac{\partial \mathcal{L}_{\mathrm{struct}}}{\partial \theta}
=
\frac{\partial \mathcal{L}_{\mathrm{struct}}}{\partial h_g}
\frac{\partial h_g}{\partial \mathcal{U}_{\theta}(x)}
\frac{\partial \mathcal{U}_{\theta}(x)}{\partial \theta}.
\label{eq:implicit_gradient}
\end{equation}
Thus, the graph teacher shapes the deployed representation without becoming part of the deployed model. To stabilize this transfer in the few-shot regime, we introduce Prototype Predictive Alignment (PPA). For each class $c$, let $\mathcal{S}_c=\{i\mid y_i=c\}$ denote the indices of support examples belonging to class $c$. We compute the class prototype from the cache keys $a_i^{\mathrm{cache}}=z_\theta(x_i)$ as
\begin{equation}
\bar{\pi}_c
=
\frac{1}{|\mathcal{S}_c|}
\sum_{i\in\mathcal{S}_c}
a_i^{\mathrm{cache}},
\qquad
\pi_c
=
\frac{\bar{\pi}_c}{\|\bar{\pi}_c\|_2}.
\label{eq:ppa_prototype}
\end{equation}
Here, $\bar{\pi}_c$ is the unnormalized support mean and $\pi_c$ is the $\ell_2$-normalized class prototype used as the PPA target.
Given a training image with label $y$, the graph representation $h_g$ is mapped to the prototype space by a lightweight head $\mathcal{P}$:
\(
    \hat{\pi}=\mathcal{P}(h_g).
\)
The PPA loss aligns this prediction with the ground-truth class prototype:
\(
    \mathcal{L}_{\mathrm{PPA}}
    =
    1-\mathrm{sim}(\hat{\pi},\pi_y).
    \label{eq:ppa_loss}
\)
Instead of fitting only instance-level labels, PPA encourages the structural branch to recover class-level visual concepts from token interactions. Since $h_g$ depends on $\mathcal{U}_{\theta}(x)$, this alignment also regularizes the LoRA feature space used by the deployed student. The complete training objective is
\begin{equation}
\mathcal{L}_{\mathrm{train}}
=
\mathcal{L}_{\mathrm{Student}}
+
\lambda_1\mathcal{L}_{\mathrm{Teacher}}
+
\lambda_2\mathcal{L}_{\mathrm{PPA}},
\label{eq:training_objective}
\end{equation}
where $\lambda_1$ and $\lambda_2$ balance the structural losses. The student loss is cross-entropy on the deployed logits and the teacher loss applies focal loss~\cite{lin2017focal} to the graph logits:
\begin{equation}
    \mathcal{L}_{\mathrm{Student}}
    =
    \mathrm{CE}\big(s_{\theta,\eta}(x),y\big),
\qquad
    \mathcal{L}_{\mathrm{Teacher}}
    =
    -(1-p_y)^\gamma \log p_y,
    \qquad
    p=\mathrm{softmax}(L_{\mathrm{Graph}}),
\end{equation}
where $p_y$ is the probability of the ground-truth class and $\gamma\geq0$ is the focusing parameter. The factor $(1-p_y)^\gamma$ down-weights confident teacher predictions and keeps larger gradients for low-confidence or misclassified examples. 

This objective makes parameter coupling explicit. The student term updates $(\theta,\eta)$ for deployment-time discrimination. The teacher and PPA terms update $(\theta,\psi)$, but only $\theta$ is shared with the deployed model. After training, $\psi$ and $\mathcal{P}$ are discarded, and inference uses only $s_{\theta,\eta}$ from Eq. \eqref{eq:student_logits}.
\vspace{-0.8em}\section{Evaluation of Structural Distillation}
\label{sec:experiments}
\vspace{-0.5em}
We evaluate on 11 standard CLIP few-shot benchmarks, 
spanning general object recognition, fine-grained categories, textures, scenes, remote sensing, and action recognition. 
We follow the standard inductive few-shot protocol~\cite{zhou2022learningcoop,zhang2022tip,zanella2024lowclip-lora}: for each dataset, we randomly sample $K\in\{1,2,4,8,16\}$ labeled examples per class from the training split and evaluate on the official validation or test split. We report top-1 accuracy. Unless stated otherwise, SD-LoRA uses CLIP ViT-B/16~\cite{radford2021learningclip}; additional ViT-B/32 results are in Appendix~\ref{app:vit_b32_results}. CLIP weights are frozen; only the LoRA parameters, cache head, and training-only teacher are optimized with AdamW (learning rate $1\times10^{-5}$, cosine decay). 
Further implementation details are in Appendix~\ref{app:implementation_details}, additional backbone and transfer results in Appendix~\ref{app:full_fewshot_results}, and hyperparameter sensitivity and computational complexity in Appendix~\ref{app:compute_complex}.

\vspace{-0.5em}
\subsection{Does Structural Distillation Improve Few-Shot Adaptation?}
\label{sec:fewshot_results}
\begin{figure*}[t]
\centering
\setlength{\tabcolsep}{0.5pt}
\renewcommand{\arraystretch}{0.9}

\begin{tabular}{@{}cccc@{}}
\includegraphics[width=0.247\textwidth]{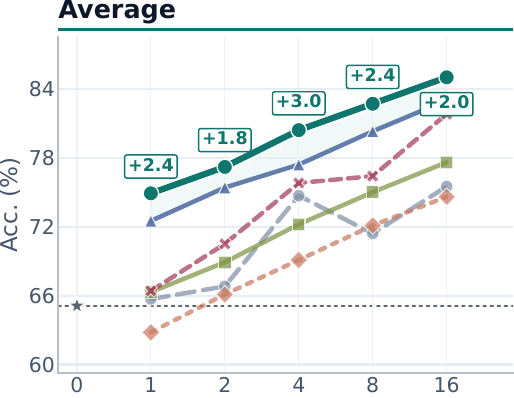} &
\includegraphics[width=0.247\textwidth]{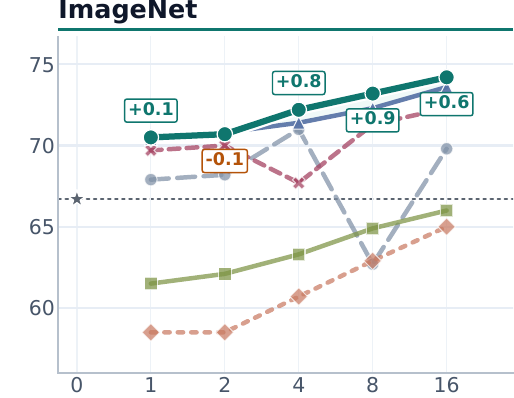} &
\includegraphics[width=0.247\textwidth]{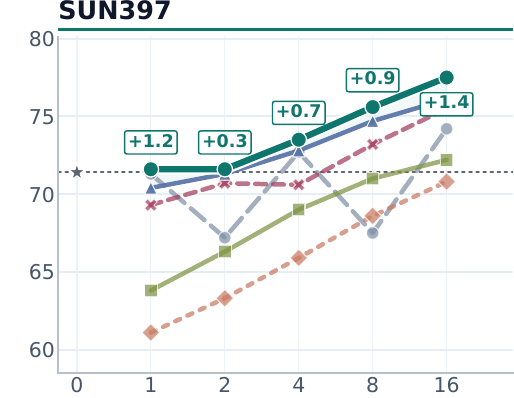} &
\includegraphics[width=0.247\textwidth]{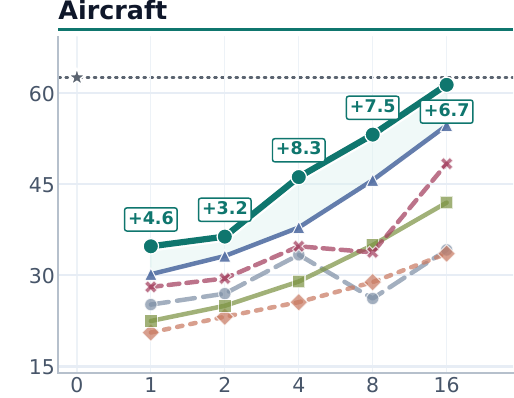} \\

\includegraphics[width=0.247\textwidth]{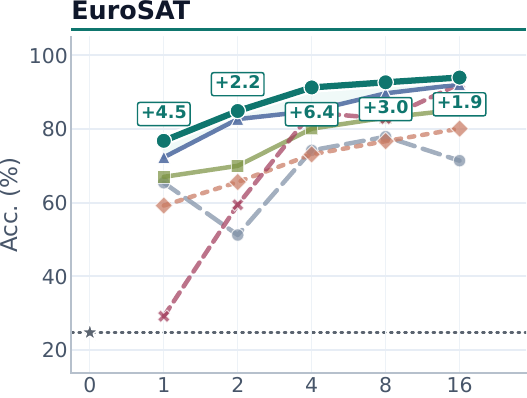} &
\includegraphics[width=0.247\textwidth]{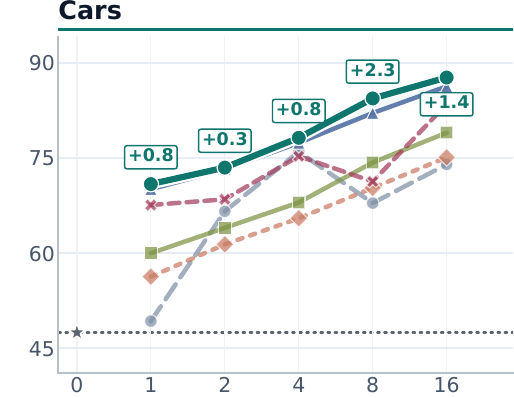} &
\includegraphics[width=0.247\textwidth]{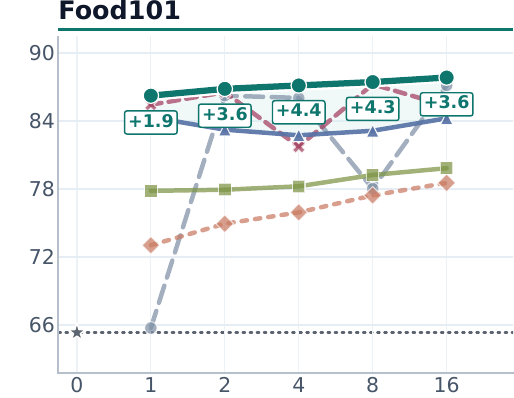} &
\includegraphics[width=0.247\textwidth]{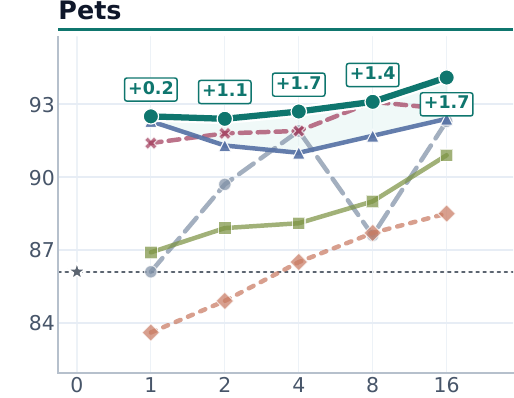} \\

\includegraphics[width=0.247\textwidth]{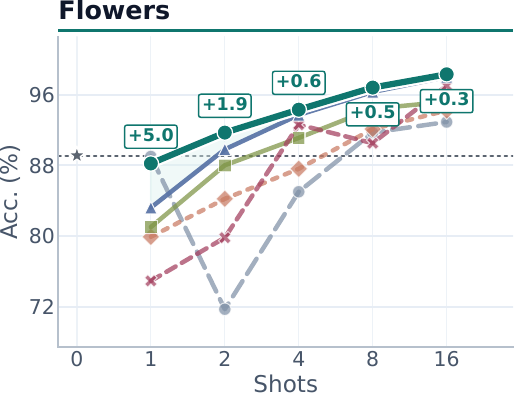} &
\includegraphics[width=0.247\textwidth]{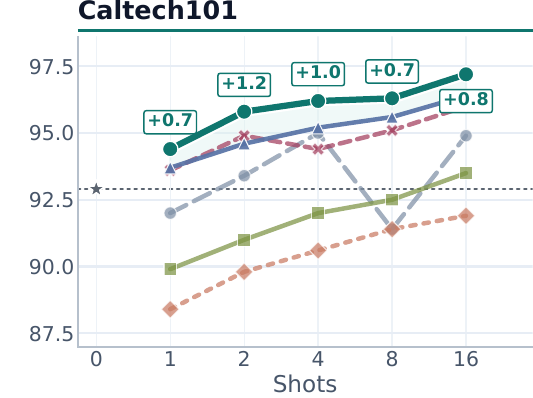} &
\includegraphics[width=0.247\textwidth]{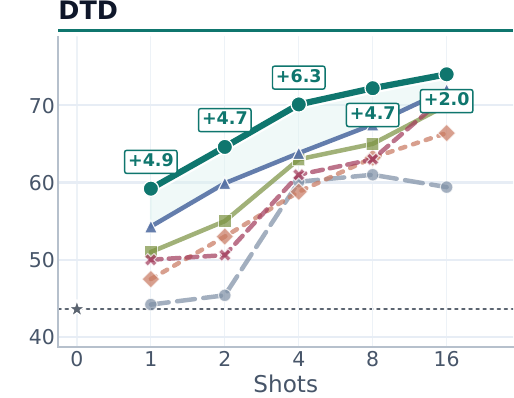} &
\includegraphics[width=0.247\textwidth]{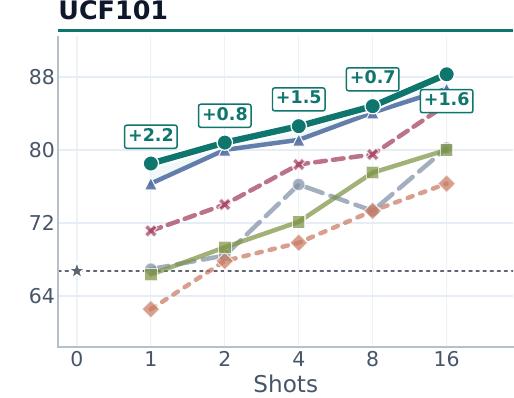} \\[-1.0ex]

\multicolumn{4}{c}{\includegraphics[width=0.92\textwidth]{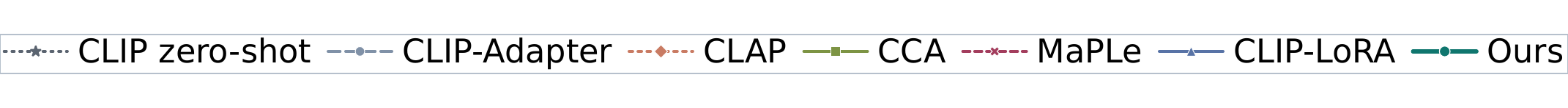}}
\end{tabular}

\caption{
\textbf{Few-shot accuracy trends across 11 benchmarks.}
Each panel shows top-1 accuracy for $K\in\{1,2,4,8,16\}$ shots on one dataset; the first panel reports the macro-average over all 11 benchmarks. SD-LoRA is compared to recent CLIP adaptation methods. 
Positive values mean structural distillation improves the deployed LoRA$+$cache model, with especially strong gains in low-shot and fine-grained scenarios.
}
\label{fig:main_lineplots_4x4}
\vspace{-1.2em}
\end{figure*}

\vspace{-0.5em}
Table~\ref{tab:main_avg_results} reports average few-shot performance over 11 benchmarks, and Figure~\ref{fig:main_lineplots_4x4} shows per-dataset trends. SD-LoRA consistently outperforms the strongest lightweight CLIP adaptation baselines across all shot regimes. Relative to CLIP-LoRA, which uses the same low-rank backbone without structural supervision, SD-LoRA improves average accuracy by 2.4 points at 1-shot, 1.8 at 2-shot, 3.0 at 4-shot, 2.4 at 8-shot, and 2.0 at 16-shot. These gains reflect the full SD-LoRA system. To analyze their source, Table~\ref{tab:main_component_ablation} separates the LoRA+cache student from the training-only graph branch: the student-only baseline isolates the effect of cache-based inference, while the additional improvement of SD-LoRA over LoRA+Cache captures the impact of structural supervision on the shared LoRA representation. 
Gains are largest on fine-grained, domain-specific datasets such as Aircraft, EuroSAT, DTD, and UCF101, where global CLIP embeddings can miss local or class-specific cues. This supports SD-LoRA’s hypothesis that token-level structural reasoning is most useful during adaptation, when it shapes the deployable representation, but need not be retained at inference. Full per-dataset comparisons with adapter- and prompt-based methods are in Appendix~\ref{app:full_fewshot_results} (Tables~\ref{tab:comparison_full_1_4_16} and \ref{tab:comparison_full_2_8}).

\textbf{Scalability with more support data.}
As shown in Figure~\ref{fig:main_lineplots_4x4}, increasing the number of shots improves all strong adaptation methods and narrows their performance gap, making large margins at higher shots less likely, particularly for recent LoRA- and adapter-based baselines. Nonetheless, SD-LoRA is best in all shot settings, outperforming the strongest prior baseline by $+2.4$, $+1.4$, $+1.5$, $+1.0$, and $+0.4$ points for 1-, 2-, 4-, 8-, and 16-shots, respectively (Table~\ref{tab:main_avg_results}). The smaller margin at 16 shots reflects that top baselines are already tightly clustered, yet SD-LoRA still achieves the highest average accuracy and clear gains on difficult datasets such as Aircraft, EuroSAT, DTD, and UCF101. This suggests that structural supervision is especially useful when local evidence is decisive, while remaining competitive as more labeled examples reduce ambiguity.
\begin{table*}[t]
\centering
\caption{
\textbf{Macro-average few-shot accuracy across 11 benchmarks.}
We report average top-1 accuracy (\%) over all datasets for top-10 representative adapter-, prompt-, and LoRA-based CLIP adaptation methods. Full per-dataset results are provided in Appendix~\ref{app:full_fewshot_results}. The best result is shown in \textbf{bold}, and the strongest previous baseline in each shot setting is underlined.
}
\label{tab:main_avg_results}
\scriptsize
\setlength{\tabcolsep}{4pt}
\begin{tabular}{lcccccccccc}
\toprule
Shots 
& TIP-F 
& TaskRes 
& CLIP-Adapt 
& CCA 
& PLOT++ 
& MaPLe 
& PromptSRC 
& CLIP-LoRA 
& CLIP-AST 
& \cellcolor{oursbg}\textbf{SD-LoRA} \\
\midrule
1  & 70.9$\pm$0.5 & 70.8$\pm$0.7 & 65.7$\pm$0.6 & 66.3$\pm$0.5 & 70.7$\pm$0.6 & 66.4$\pm$0.5 & 72.1$\pm$0.7 & \underline{72.5$\pm$0.5} & 72.3$\pm$0.2 & \cellcolor{oursbg}\textbf{74.9$\pm$0.1} \\
2  & 73.7$\pm$0.4 & 72.9$\pm$0.4 & 66.8$\pm$0.6 & 68.9$\pm$0.4 & 74.0$\pm$0.6 & 70.5$\pm$0.3 & 75.2$\pm$0.2 & 75.4$\pm$0.6 & \underline{75.8$\pm$0.3} & \cellcolor{oursbg}\textbf{77.2$\pm$0.3} \\
4  & 75.6$\pm$0.1 & 70.3$\pm$0.2 & 74.7$\pm$0.4 & 72.2$\pm$0.3 & 76.9$\pm$0.1 & 75.8$\pm$0.4 & 78.3$\pm$0.4 & 77.4$\pm$0.1 & \underline{78.9$\pm$0.2} & \cellcolor{oursbg}\textbf{80.4$\pm$0.1} \\
8  & 78.3$\pm$0.3 & 78.4$\pm$0.3 & 71.4$\pm$0.2 & 75.0$\pm$0.3 & 79.6$\pm$0.2 & 76.4$\pm$0.3 & 80.6$\pm$0.3 & 80.3$\pm$0.2 & \underline{81.7$\pm$0.1} & \cellcolor{oursbg}\textbf{82.7$\pm$0.1} \\
16 & 80.7$\pm$0.2 & 80.8$\pm$0.2 & 75.5$\pm$0.1 & 77.6$\pm$0.2 & 82.1$\pm$0.1 & 81.8$\pm$0.1 & 82.7$\pm$0.2 & 83.0$\pm$0.1 & \underline{84.6$\pm$0.1} & \cellcolor{oursbg}\textbf{85.0$\pm$0.2} \\
\bottomrule
\end{tabular}
\end{table*}
\vspace{-0.8em}\subsection{Does the Learned Representation Generalize?}
\label{sec:generalization}
\vspace{-0.5em}
Standard few-shot accuracy only measures adaptation to a fixed support set, not transfer to unseen classes or robustness to distribution shift. We therefore also evaluate SD-LoRA on base-to-novel generalization and ImageNet out-of-distribution robustness.


\paragraph{Base-to-novel generalization.}
Table~\ref{tab:base_to_novel_generalization} evaluates whether adaptation to base classes transfers to novel classes under the MaPLe protocol. SD-LoRA achieves the best average performance on all three metrics: 87.23\% on base classes, 86.59\% on novel classes, and an 86.91\% harmonic mean. Compared to the strongest baseline, it yields a small gain on base classes (+0.27) but clear gains on novel classes (+5.85) and harmonic mean (+3.18), indicating that SD-LoRA avoids overfitting to base classes and instead learns representations that transfer better to unseen categories. Improvements are strongest on fine-grained and domain-specific datasets: SD-LoRA improves novel-class accuracy over the strongest baseline by +9.33 on Cars, +14.79 on EuroSAT, +15.54 on Flowers, +10.16 on UCF101, and +9.31 on DTD. These gains indicate that token-level structural supervision preserves class-discriminative local evidence beyond the support classes. Although SD-LoRA is not always best on base-class accuracy, it consistently improves the base--novel trade-off, as shown by the harmonic mean.

\textbf{Out-of-distribution robustness.}
Table~\ref{tab:ood_results} reports robustness by adapting on ImageNet and testing on four OOD variants: ImageNet-R, ImageNet-A, ImageNet-V2, and ImageNet-Sketch. SD-LoRA achieves the highest average accuracy across all five datasets, surpassing the strongest structured baseline, GraphAdapter, by $0.41$ points. The margin is smaller than in the base-to-novel setting because all methods use a strong, shared ImageNet adaptation protocol, and the top baselines are closely matched. Notably, we achieve this without any graph- or token-level reasoning at inference.

SD-LoRA achieves the best accuracy on ImageNet, ImageNet-R, and ImageNet-Sketch, and remains competitive on ImageNet-A and ImageNet-V2, reflecting the impact of the training-time graph teacher. Structural distillation strengthens local texture, shape, and part-level cues, which particularly benefits rendition and sketch-style shifts. In contrast, ImageNet-A contains adversarially filtered natural images with strong object, pose, and context biases, where improving local evidence alone may not fully address the shift. ImageNet-V2 is close to the original ImageNet validation set, and several source-adapted baselines already perform similarly, leaving limited room for gains. Overall, SD-LoRA improves the average OOD trade-off without requiring test-time graph reasoning, with its advantage largest when token-level visual evidence is directly informative. It also preserves the lightweight CLIP-LoRA$+$cache deployment: after adaptation, the graph teacher and prototype head are removed. These results support our central claim that structural reasoning can enhance representations during training without increasing test-time complexity.

\textbf{Additional transfer and calibration results.}
Additional generalization results appear in Appendix~\ref{app:cross_dataset} and Appendix~\ref{app:calibration}. Adapted on ImageNet and evaluated on ten target datasets without target-domain training, SD-LoRA raises average cross-dataset accuracy over PromptKD from $71.33\%$ to $74.98\%$ $(+3.65)$, with gains on 9 of 10 datasets. Under the same setting, calibration also improves: average ECE drops from $5.57$ to $2.65$, NLL from $1.05$ to $0.62$, and Brier score from $0.38$ to $0.20$, indicating more accurate, better-calibrated representations under domain shift.
\begin{table*}[htbp]
\centering
\caption{
\textbf{Base-to-novel generalization under the MaPLe protocol.}
We report accuracy (\%) on base classes, novel classes, and their harmonic mean (HM). Avg.(11) is the unweighted mean over 11 benchmarks. The $\Delta$ row shows the improvement of SD-LoRA over the strongest baseline for each metric. Best results are in \textbf{bold}.
}
\label{tab:base_to_novel_generalization}
\vspace{-0.3em}
\tiny
\setlength{\tabcolsep}{2.6pt}
\renewcommand{\arraystretch}{1.12}
\resizebox{\textwidth}{!}{
\begin{tabular}{@{}lccc ccc ccc ccc ccc ccc@{}}
\toprule
\multirow{2}{*}{\textbf{Method}}
& \multicolumn{3}{c}{\textbf{Avg.(11)}}
& \multicolumn{3}{c}{\textbf{Cars}}
& \multicolumn{3}{c}{\textbf{EuroSAT}}
& \multicolumn{3}{c}{\textbf{Flowers}}
& \multicolumn{3}{c}{\textbf{UCF}}
& \multicolumn{3}{c}{\textbf{DTD}} \\
\cmidrule(lr){2-4}
\cmidrule(lr){5-7}
\cmidrule(lr){8-10}
\cmidrule(lr){11-13}
\cmidrule(lr){14-16}
\cmidrule(lr){17-19}
& Base & Novel & HM
& Base & Novel & HM
& Base & Novel & HM
& Base & Novel & HM
& Base & Novel & HM
& Base & Novel & HM \\
\midrule

CLIP
& 69.34 & 74.22 & 71.70
& 63.37 & 74.89 & 68.65
& 56.48 & 64.05 & 60.03
& 72.08 & 77.80 & 74.83
& 70.53 & 77.50 & 73.85
& 53.24 & 59.90 & 56.37 \\

CoOp
& 82.69 & 63.22 & 71.66
& 78.12 & 60.40 & 68.13
& 92.19 & 54.74 & 68.69
& 97.60 & 59.67 & 74.06
& 84.69 & 56.05 & 67.46
& 79.44 & 41.18 & 54.24 \\

CoCoOp
& 80.47 & 71.69 & 75.83
& 70.49 & 73.59 & 72.01
& 87.49 & 60.04 & 71.21
& 94.87 & 71.75 & 81.71
& 82.33 & 73.45 & 77.64
& 77.01 & 56.00 & 64.85 \\

ProGrad
& 82.48 & 70.75 & 76.16
& 77.68 & 68.63 & 72.88
& 90.11 & 60.89 & 72.67
& 95.54 & 71.87 & 82.03
& 84.33 & 74.94 & 79.35
& 77.35 & 52.35 & 62.45 \\

KgCoOp
& 80.73 & 73.60 & 77.00
& 71.76 & 75.04 & 73.36
& 85.64 & 64.34 & 73.48
& 95.00 & 74.73 & 83.65
& 82.89 & 76.67 & 79.65
& 77.55 & 54.99 & 64.35 \\

MaPLe
& 82.28 & 75.14 & 78.55
& 72.94 & 74.00 & 73.47
& 94.07 & 73.23 & 82.35
& 95.92 & 72.46 & 82.56
& 83.00 & 78.66 & 80.77
& 80.36 & 59.18 & 68.16 \\

PromptSRC
& 84.26 & 76.10 & 79.97
& 78.27 & 74.97 & 76.58
& 92.90 & 73.90 & 82.32
& 98.07 & 76.50 & 85.95
& 87.10 & 78.80 & 82.74
& 83.37 & 62.97 & 71.75 \\

CLIP-AST
& 85.94 & 76.99 & 81.06
& 84.21 & 74.05 & 78.80
& 95.90 & 81.72 & 88.24
& 97.91 & 77.73 & 86.66
& 87.38 & 79.12 & 83.05
& 84.03 & 65.34 & 73.52 \\

PromptKD
& 86.96 & 80.73 & 83.73
& 82.80 & 83.37 & 83.13
& \best{97.54} & 82.08 & 89.14
& \best{99.42} & 82.62 & 90.24
& 89.71 & 82.27 & 86.10
& \best{85.84} & 71.37 & 77.94 \\

\midrule

\rowcolor{oursrow}
\textbf{Ours}
& \best{87.23} & \best{86.59} & \best{86.91}
& \best{86.06} & \best{92.70} & \best{89.25}
& 97.29 & \best{96.87} & \best{97.08}
& 98.20 & \best{98.16} & \best{98.18}
& \best{90.07} & \best{92.43} & \best{91.24}
& 84.68 & \best{80.68} & \best{82.63} \\

\rowcolor{deltarow}
$\Delta$
& \pos{+0.27} & \pos{+5.85} & \pos{+3.18}
& \pos{+1.85} & \pos{+9.33} & \pos{+6.12}
& \negv{-0.25} & \pos{+14.79} & \pos{+7.94}
& \negv{-1.22} & \pos{+15.54} & \pos{+7.94}
& \pos{+0.36} & \pos{+10.16} & \pos{+5.14}
& \negv{-1.16} & \pos{+9.31} & \pos{+4.69} \\

\bottomrule
\end{tabular}%
}
\vspace{-2.0em}
\end{table*}
\begin{table}[t]
\centering
\caption{
\textbf{Out-of-distribution (OOD) generalization.}
Models are adapted on ImageNet and evaluated on ImageNet and four OOD variants. Top-1 accuracy (\%) is reported, and ``Avg.'' denotes the unweighted mean over all five datasets. Best results are shown in \textbf{bold}.
}
\label{tab:ood_results}
\vspace{0.5mm}
\scriptsize
\setlength{\tabcolsep}{2.5pt}
\begin{tabular}{@{}lcccccccccc@{}}
\toprule
\textbf{Dataset} 
& \textbf{CLIP} 
& \textbf{CoOp} 
& \textbf{CoCoOp} 
& \textbf{MaPLe} 
& \textbf{ProGrad} 
& \textbf{KgCoOp} 
& \textbf{GraphAdapter} 
& \textbf{TransCLIP} 
& \textbf{CLIP-AST} 
& \cellcolor{oursrow}\textbf{Ours} \\
\midrule
ImageNet 
& 66.73 & 71.51 & 71.02 & 70.72 & 72.24 & 71.20 & 73.68 & 70.30 & 73.87 
& \cellcolor{oursrow}\textbf{74.20} \\

ImageNet-R 
& 73.96 & 75.21 & 76.18 & 76.98 & 74.58 & 76.70 & 77.20 & 75.00 & --- 
& \cellcolor{oursrow}\textbf{77.32} \\

ImageNet-A 
& 47.77 & 49.71 & 50.63 & \textbf{50.90} & 49.39 & 50.69 & 49.23 & 49.50 & --- 
& \cellcolor{oursrow}49.63 \\

ImageNet-V2 
& 60.83 & 64.20 & 64.07 & 64.07 & 64.73 & 64.10 & 65.57 & 62.30 & \textbf{66.37} 
& \cellcolor{oursrow}65.78 \\

ImageNet-Sketch 
& 46.15 & 47.99 & 48.75 & 49.15 & 47.61 & 48.97 & 48.57 & 49.07 & 48.37 
& \cellcolor{oursrow}\textbf{49.38} \\

\midrule
Avg. 
& 59.09 & 61.72 & 62.13 & 62.23 & 61.71 & 62.33 & 62.85 & 61.23 & --- 
& \cellcolor{oursrow}\textbf{63.26} \\
\bottomrule
\end{tabular}%
\end{table}
\vspace{-0.8em}\subsection{What Components Drive the Improvement?}
\label{sec:ablation_analysis}
\vspace{-0.5em}
Table~\ref{tab:main_component_ablation} isolates the effects of the cache branch and the training-only graph teacher. LoRA$+$Cache is the deployed student without structural supervision, and LoRA$+$Graph removes the cache but retains graph-guided training of the shared LoRA backbone. Both components help, but together they work best. Across Aircraft, Caltech101, and EuroSAT, SD-LoRA outperforms LoRA$+$Cache by $+2.2$, $+3.0$, and $+2.2$ points at 1-, 4-, and 16-shot learning, and exceeds LoRA$+$Graph by $+1.5$, $+1.3$, and $+0.8$ points. Thus, the cache and graph are complementary: the cache offers efficient support-set memory at inference, while the graph enhances LoRA training. Because the graph is discarded after adaptation, these gains add no test-time graph cost.

\begin{figure}[htbp]
    \centering
    \includegraphics[width=\linewidth]{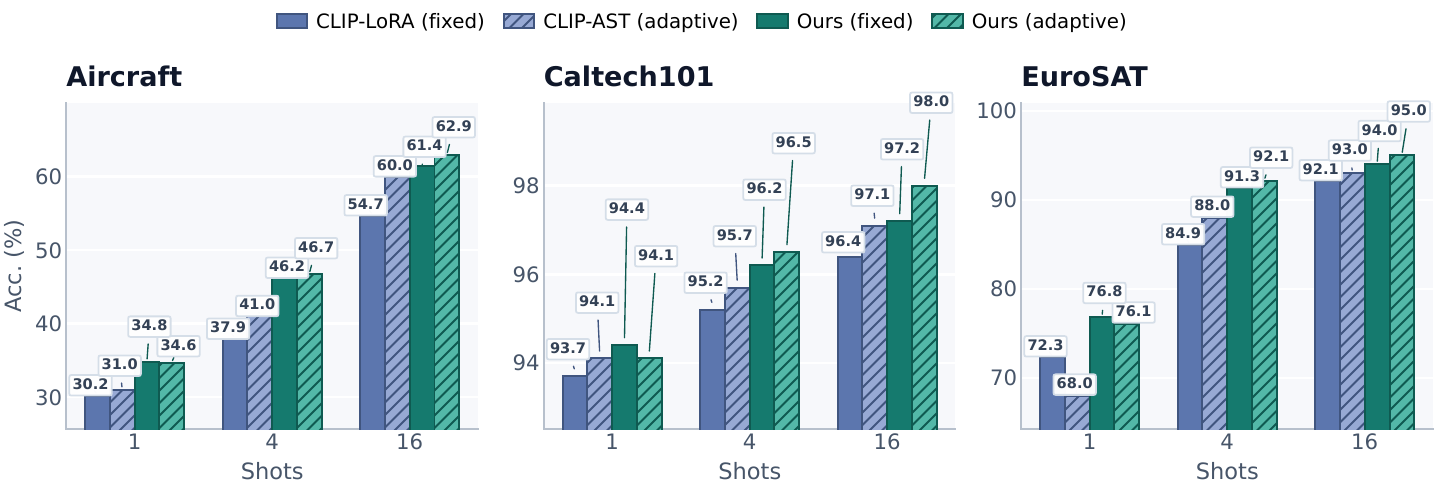}
    \caption{
\textbf{Effect of LoRA placement.}
We compare fixed LoRA placement and adaptive layer selection under 1/4/16-shot settings. SD-LoRA outperforms both fixed CLIP-LoRA and adaptive CLIP-AST, indicating that adaptive placement complements SD-LoRA: layer selection optimizes where LoRA is inserted, while structural distillation improves how its parameters are trained.
}
    \label{fig:lora_placement}
    \vspace{-1.0em}
\end{figure}

\textbf{Effect of LoRA placement.}
Figure~\ref{fig:lora_placement} shows that SD-LoRA consistently outperforms both fixed CLIP-LoRA and adaptive CLIP-AST across datasets and shot settings. This indicates that gains are not only from better LoRA placement: adaptive placement decides where to insert LoRA, while structural distillation improves how its parameters are trained.

More ablations are provided for graph edge families (Appx.~\ref{app:edge_type_ablation}, Table~\ref{tab:edge_type_ablation}), PPA (Appx.~\ref{app:ppa_graph_ablation}, Table~\ref{tab:ppa_graph_ablation}), token filtering (Appx.~\ref{app:token_filtering}, Table~\ref{tab:topk_pooling}), input resolution (Appx.~\ref{app:resolution_ablation}, Table~\ref{tab:resolution_ablation}), graph design (Appx.~\ref{app:graph_architecture_ablation}, Table~\ref{tab:components}), and loss terms (Appx.~\ref{app:loss_ablation}, Table~\ref{tab:loss_ablation_final}). Appx.~\ref{app:compute_complex} details computational complexity and hyperparameter selection, and Appx.~\ref{app:limitations} discusses scope, limitations, and responsible use.

\begin{table*}[t]
\centering
\caption{
\textbf{Main component ablation.}
LoRA is enabled in all variants. Cache is a deployed inference component, while the graph is a training-only teacher and is removed after adaptation. SD-LoRA combines both components and achieves the best performance across all datasets and shot settings.
}
\label{tab:main_component_ablation}
\vspace{-0.3em}
\scriptsize
\setlength{\tabcolsep}{4.2pt}
\begin{tabular}{@{}lcc ccc ccc ccc@{}}
\toprule
\multirow{2}{*}{\textbf{Variant}}
& \multicolumn{2}{c}{\textbf{Components}}
& \multicolumn{3}{c}{\textbf{Aircraft}}
& \multicolumn{3}{c}{\textbf{Caltech101}}
& \multicolumn{3}{c}{\textbf{EuroSAT}} \\
\cmidrule(lr){2-3}
\cmidrule(lr){4-6}
\cmidrule(lr){7-9}
\cmidrule(lr){10-12}
& \textbf{Cache} 
& \textbf{Graph}
& \textbf{1-Shot} 
& \textbf{4-shot} 
& \textbf{16-shot}
& \textbf{1-shot} 
& \textbf{4-shot} 
& \textbf{16-shot}
& \textbf{1-shot} 
& \textbf{4-shot} 
& \textbf{16-shot} \\
\midrule

LoRA + Cache
& \yes & \no
& 32.2 & 43.5 & 59.2
& 92.7 & 93.0 & 95.1
& 74.6 & 88.3 & 91.8 \\

LoRA + Graph
& \no & \yes
& 33.1 & 44.7 & 60.7
& 93.2 & 94.8 & 96.5
& 75.3 & 90.2 & 93.1 \\

\rowcolor{oursbg}
\textbf{SD-LoRA}
& \yes & \yes
& \best{34.8} & \best{46.2} & \best{61.4}
& \best{94.4} & \best{96.2} & \best{97.2}
& \best{76.8} & \best{91.3} & \best{94.0} \\

\bottomrule
\end{tabular}%
\end{table*}

\begin{minipage}{\linewidth}
\textbf{Matched controls and retained structural supervision.}
On Aircraft, Caltech101, EuroSAT, and Food101, we match support splits, optimization, and cache-based deployment (three runs per configuration). Table~\ref{tab:sd_matched_main} varies training resolution; all models deploy at $224\times224$. Structural training adds $+1.8/+1.7/+1.7/+1.7$ pp over the matched 336-trained cache baseline. After teacher removal, the deployed student's intra-class distance falls from 0.560 to 0.511 and inter-class distance rises from 0.713 to 0.890; prototype margin increases from 0.075 to 0.156. These results support a persistent structural-training effect beyond cache or resolution alone. Full controls and representation diagnostics appear in Appx.~\ref{app:matched_resolution_cache} and~\ref{app:deployed_features}.
\begin{center}
\centering
\captionsetup{hypcap=false}
\captionof{table}{\textbf{Resolution- and cache-matched controls.} Top-1 accuracy (\%) is averaged over four datasets, distinct from the 11-benchmark averages in Table~\ref{tab:main_avg_results}. @ denotes training resolution; inference is always $224\times224$.}
\label{tab:sd_matched_main}
\small
\setlength{\tabcolsep}{8pt}
\begin{tabular}{lcccc}
\toprule
Training configuration & 1-shot & 4-shot & 8-shot & 16-shot \\
\midrule
CLIP-LoRA+cache @224 & 70.8 & 77.2 & 79.8 & 82.9 \\
CLIP-LoRA+cache @336 & 71.3 & 78.5 & 80.7 & 83.4 \\
\rowcolor{oursrow}
\textbf{SD-LoRA @336} & \textbf{73.1} & \textbf{80.2} & \textbf{82.4} & \textbf{85.1} \\
\bottomrule
\end{tabular}
\end{center}
\end{minipage}

\section{Conclusion}
\label{sec:conclusion}
\vspace{-0.8em}
We proposed SD-LoRA, a training-time structural distillation framework for few-shot CLIP adaptation that decouples training from deployment. A relation-aware graph teacher is used only during optimization to reason over visual tokens and class-text embeddings, transferring structural supervision into a shared LoRA backbone. At test time, the teacher is removed and inference is based solely on a lightweight CLIP-LoRA$+$cache student. Unlike many efficient VLM adapters that supervise mainly global image embeddings, SD-LoRA injects token-level relational bias during training while keeping the inference cost of the standard LoRA$+$cache. Experiments on few-shot accuracy, base-to-novel generalization, OOD robustness, and calibration show consistent gains over strong baselines, and ablations confirm the complementarity of graph-guided training and cache-based inference. SD-LoRA demonstrates that structural reasoning can enhance parameter-efficient VLM adaptation without adding inference-time overhead.


\begin{ack}
\vspace{-0.8em}
This work was supported by UK Research and Innovation (UKRI) through
the Engineering and Physical Sciences Research Council (EPSRC) grant
ATRACT (\texttt{EP/X028631/1}), the Economic and Social Research Council (ESRC) grant SCAnDi (\texttt{ES/Y010655/1}), and Cancer Research UK (CRUK) grants \texttt{EDDISA-May21\textbackslash100001} and
\texttt{EDDPMA-May22\textbackslash100028}.
\end{ack}

{
\small
\bibliographystyle{plainnat}
\bibliography{main}

\begin{thebibliography}{46}
\providecommand{\natexlab}[1]{#1}
\providecommand{\url}[1]{\texttt{#1}}
\expandafter\ifx\csname urlstyle\endcsname\relax
  \providecommand{\doi}[1]{doi: #1}\else
  \providecommand{\doi}{doi: \begingroup \urlstyle{rm}\Url}\fi

\bibitem[Ahmad et~al.(2025)Ahmad, Sikdar, Pradhan, and Behera]{ahmad2025tip}
Tasweer Ahmad, Arindam Sikdar, Sandip Pradhan, and Ardhendu Behera.
\newblock Advancing cache-based few-shot classification via patch-driven relational gated graph attention.
\newblock \emph{arXiv preprint arXiv:2512.12498}, 2025.

\bibitem[Bendou et~al.(2025)Bendou, Ouasfi, Gripon, and Boukhayma]{bendou2025proker}
Yassir Bendou, Amine Ouasfi, Vincent Gripon, and Adnane Boukhayma.
\newblock Proker: A kernel perspective on few-shot adaptation of large vision-language models.
\newblock In \emph{Proceedings of the IEEE/CVF Conference on Computer Vision and Pattern Recognition (CVPR)}, pages 25092--25102, 2025.

\bibitem[Bossard et~al.(2014)Bossard, Guillaumin, and Van~Gool]{bossard2014food}
Lukas Bossard, Matthieu Guillaumin, and Luc Van~Gool.
\newblock Food-101--mining discriminative components with random forests.
\newblock In \emph{European conference on computer vision}, pages 446--461. Springer, 2014.

\bibitem[Chen et~al.(2022)Chen, Yao, Song, Li, Rao, and Zhang]{chen2022promptplot++}
Guangyi Chen, Weiran Yao, Xiangchen Song, Xinyue Li, Yongming Rao, and Kun Zhang.
\newblock Prompt learning with optimal transport for vision-language models.
\newblock 2022.

\bibitem[Cimpoi et~al.(2014)Cimpoi, Maji, Kokkinos, Mohamed, and Vedaldi]{cimpoi2014describing}
Mircea Cimpoi, Subhransu Maji, Iasonas Kokkinos, Sammy Mohamed, and Andrea Vedaldi.
\newblock Describing textures in the wild.
\newblock In \emph{Proceedings of the IEEE conference on computer vision and pattern recognition}, pages 3606--3613, 2014.

\bibitem[Deng et~al.(2009)Deng, Dong, Socher, Li, Li, and Fei-Fei]{deng2009imagenet}
Jia Deng, Wei Dong, Richard Socher, Li-Jia Li, Kai Li, and Li~Fei-Fei.
\newblock Imagenet: A large-scale hierarchical image database.
\newblock In \emph{2009 IEEE conference on computer vision and pattern recognition}, pages 248--255. Ieee, 2009.

\bibitem[Fei-Fei et~al.(2004)Fei-Fei, Fergus, and Perona]{fei2004learning}
Li~Fei-Fei, Rob Fergus, and Pietro Perona.
\newblock Learning generative visual models from few training examples: An incremental bayesian approach tested on 101 object categories.
\newblock In \emph{2004 conference on computer vision and pattern recognition workshop}, pages 178--178. IEEE, 2004.

\bibitem[Gao et~al.(2024)Gao, Geng, Zhang, Ma, Fang, Zhang, Li, and Qiao]{gao2024clip-adapter}
Peng Gao, Shijie Geng, Renrui Zhang, Teli Ma, Rongyao Fang, Yongfeng Zhang, Hongsheng Li, and Yu~Qiao.
\newblock Clip-adapter: Better vision-language models with feature adapters.
\newblock \emph{International Journal of Computer Vision}, 132\penalty0 (2):\penalty0 581--595, 2024.

\bibitem[Ghiasvand et~al.(2025)Ghiasvand, Oskouie, Alizadeh, and Pedarsani]{ghiasvand2025few}
Sajjad Ghiasvand, Haniyeh~Ehsani Oskouie, Mahnoosh Alizadeh, and Ramtin Pedarsani.
\newblock Few-shot adversarial low-rank fine-tuning of vision-language models.
\newblock \emph{arXiv preprint arXiv:2505.15130}, 2025.

\bibitem[Helber et~al.(2019)Helber, Bischke, Dengel, and Borth]{helber2019eurosat}
Patrick Helber, Benjamin Bischke, Andreas Dengel, and Damian Borth.
\newblock Eurosat: A novel dataset and deep learning benchmark for land use and land cover classification.
\newblock \emph{IEEE Journal of Selected Topics in Applied Earth Observations and Remote Sensing}, 12\penalty0 (7):\penalty0 2217--2226, 2019.

\bibitem[Hinton et~al.(2015)Hinton, Vinyals, and Dean]{hinton2015distillingkd}
Geoffrey Hinton, Oriol Vinyals, and Jeff Dean.
\newblock Distilling the knowledge in a neural network.
\newblock \emph{arXiv preprint arXiv:1503.02531}, 2015.

\bibitem[Hu et~al.(2022)Hu, Shen, Wallis, Allen-Zhu, Li, Wang, Wang, Chen, et~al.]{hu2022lora}
Edward~J Hu, Yelong Shen, Phillip Wallis, Zeyuan Allen-Zhu, Yuanzhi Li, Shean Wang, Lu~Wang, Weizhu Chen, et~al.
\newblock Lora: Low-rank adaptation of large language models.
\newblock \emph{ICLR}, 1\penalty0 (2):\penalty0 3, 2022.

\bibitem[Jia et~al.(2021)Jia, Yang, Xia, Chen, Parekh, Pham, Le, Sung, Li, and Duerig]{jia2021scaling}
Chao Jia, Yinfei Yang, Ye~Xia, Yi-Ting Chen, Zarana Parekh, Hieu Pham, Quoc Le, Yun-Hsuan Sung, Zhen Li, and Tom Duerig.
\newblock Scaling up visual and vision-language representation learning with noisy text supervision.
\newblock In \emph{International conference on machine learning}, pages 4904--4916. PMLR, 2021.

\bibitem[Jiang et~al.(2025)Jiang, Zhang, Liu, and Shi]{jiang2025causalcca}
Tianjiao Jiang, Zhen Zhang, Yuhang Liu, and Javen~Qinfeng Shi.
\newblock Causal disentanglement and cross-modal alignment for enhanced few-shot learning.
\newblock In \emph{Proceedings of the IEEE/CVF International Conference on Computer Vision}, pages 890--900, 2025.

\bibitem[Khattak et~al.(2023{\natexlab{a}})Khattak, Rasheed, Maaz, Khan, and Khan]{khattak2023maple}
Muhammad~Uzair Khattak, Hanoona Rasheed, Muhammad Maaz, Salman Khan, and Fahad~Shahbaz Khan.
\newblock Maple: Multi-modal prompt learning.
\newblock In \emph{Proceedings of the IEEE/CVF conference on computer vision and pattern recognition}, pages 19113--19122, 2023{\natexlab{a}}.

\bibitem[Khattak et~al.(2023{\natexlab{b}})Khattak, Wasim, Naseer, Khan, Yang, and Khan]{khattak2023self}
Muhammad~Uzair Khattak, Syed~Talal Wasim, Muzammal Naseer, Salman Khan, Ming-Hsuan Yang, and Fahad~Shahbaz Khan.
\newblock Self-regulating prompts: Foundational model adaptation without forgetting.
\newblock In \emph{Proceedings of the IEEE/CVF international conference on computer vision}, pages 15190--15200, 2023{\natexlab{b}}.

\bibitem[Krause et~al.(2013)Krause, Stark, Deng, and Fei-Fei]{krause20133d}
Jonathan Krause, Michael Stark, Jia Deng, and Li~Fei-Fei.
\newblock 3d object representations for fine-grained categorization.
\newblock In \emph{Proceedings of the IEEE international conference on computer vision workshops}, pages 554--561, 2013.

\bibitem[Kravets et~al.(2025)Kravets, Chen, and Namboodiri]{kravets2025rethinking}
Alexey Kravets, Da~Chen, and Vinay~P Namboodiri.
\newblock Rethinking few shot clip benchmarks: A critical analysis in the inductive setting.
\newblock In \emph{Proceedings of the IEEE/CVF International Conference on Computer Vision}, pages 1902--1911, 2025.

\bibitem[Li et~al.(2025)Li, Wang, Wang, Jiang, Peng, and Long]{li2025dpc}
Haoyang Li, Liang Wang, Chao Wang, Jing Jiang, Yan Peng, and Guodong Long.
\newblock Dpc: Dual-prompt collaboration for tuning vision-language models.
\newblock In \emph{Proceedings of the IEEE/CVF Conference on Computer Vision and Pattern Recognition (CVPR)}, pages 25623--25633, 2025.

\bibitem[Li et~al.(2022)Li, Zhang, Zhang, Yang, Li, Zhong, Wang, Yuan, Zhang, Hwang, et~al.]{li2022grounded}
Liunian~Harold Li, Pengchuan Zhang, Haotian Zhang, Jianwei Yang, Chunyuan Li, Yiwu Zhong, Lijuan Wang, Lu~Yuan, Lei Zhang, Jenq-Neng Hwang, et~al.
\newblock Grounded language-image pre-training.
\newblock In \emph{Proceedings of the IEEE/CVF conference on computer vision and pattern recognition}, pages 10965--10975, 2022.

\bibitem[Li et~al.(2023)Li, Lian, Lu, Bai, Chen, and Wang]{li2023graphadapter}
Xin Li, Dongze Lian, Zhihe Lu, Jiawang Bai, Zhibo Chen, and Xinchao Wang.
\newblock Graphadapter: Tuning vision-language models with dual knowledge graph.
\newblock \emph{Advances in Neural Information Processing Systems}, 36:\penalty0 13448--13466, 2023.

\bibitem[Li et~al.(2024)Li, Li, Fu, Zhang, Wang, Chen, and Yang]{li2024promptkd}
Zheng Li, Xiang Li, Xinyi Fu, Xin Zhang, Weiqiang Wang, Shuo Chen, and Jian Yang.
\newblock Promptkd: Unsupervised prompt distillation for vision-language models.
\newblock In \emph{Proceedings of the IEEE/CVF Conference on Computer Vision and Pattern Recognition (CVPR)}, pages 26617--26626, 2024.

\bibitem[Lin et~al.(2017)Lin, Goyal, Girshick, He, and Doll{\'a}r]{lin2017focal}
Tsung-Yi Lin, Priya Goyal, Ross Girshick, Kaiming He, and Piotr Doll{\'a}r.
\newblock Focal loss for dense object detection.
\newblock In \emph{Proceedings of the IEEE international conference on computer vision}, pages 2980--2988, 2017.

\bibitem[Maji et~al.(2013)Maji, Rahtu, Kannala, Blaschko, and Vedaldi]{maji2013fineaircraft}
Subhransu Maji, Esa Rahtu, Juho Kannala, Matthew Blaschko, and Andrea Vedaldi.
\newblock Fine-grained visual classification of aircraft.
\newblock \emph{arXiv preprint arXiv:1306.5151}, 2013.

\bibitem[Mohammad et~al.(2026)Mohammad, Behera, Pradhan, Kumar, and Ahmed]{Mohammad_2026_CVPR}
Mohammed Rahman Sherif~Khan Mohammad, Ardhendu Behera, Sandip Pradhan, Swagat Kumar, and Amr Ahmed.
\newblock Training-only heterogeneous image-patch-text graph supervision for advancing few-shot learning adapters.
\newblock In \emph{Proceedings of the IEEE/CVF Conference on Computer Vision and Pattern Recognition (CVPR)}, pages 19613--19622, June 2026.

\bibitem[Nilsback and Zisserman(2008)]{nilsback2008automatedflowers}
Maria-Elena Nilsback and Andrew Zisserman.
\newblock Automated flower classification over a large number of classes.
\newblock In \emph{2008 Sixth Indian conference on computer vision, graphics \& image processing}, pages 722--729. IEEE, 2008.

\bibitem[Parkhi et~al.(2012)Parkhi, Vedaldi, Zisserman, and Jawahar]{parkhi2012cats}
Omkar~M Parkhi, Andrea Vedaldi, Andrew Zisserman, and CV~Jawahar.
\newblock Cats and dogs.
\newblock In \emph{2012 IEEE conference on computer vision and pattern recognition}, pages 3498--3505. IEEE, 2012.

\bibitem[Radford et~al.(2021)Radford, Kim, Hallacy, Ramesh, Goh, Agarwal, Sastry, Askell, Mishkin, Clark, et~al.]{radford2021learningclip}
Alec Radford, Jong~Wook Kim, Chris Hallacy, Aditya Ramesh, Gabriel Goh, Sandhini Agarwal, Girish Sastry, Amanda Askell, Pamela Mishkin, Jack Clark, et~al.
\newblock Learning transferable visual models from natural language supervision.
\newblock In \emph{International conference on machine learning}, pages 8748--8763. PmLR, 2021.

\bibitem[Rao et~al.(2022)Rao, Zhao, Chen, Tang, Zhu, Huang, Zhou, and Lu]{rao2022denseclip}
Yongming Rao, Wenliang Zhao, Guangyi Chen, Yansong Tang, Zheng Zhu, Guan Huang, Jie Zhou, and Jiwen Lu.
\newblock Denseclip: Language-guided dense prediction with context-aware prompting.
\newblock In \emph{Proceedings of the IEEE/CVF conference on computer vision and pattern recognition}, pages 18082--18091, 2022.

\bibitem[Roy and Etemad(2024)]{roy2024coprompt}
Shuvendu Roy and Ali Etemad.
\newblock Consistency-guided prompt learning for vision-language models.
\newblock In \emph{The Twelfth International Conference on Learning Representations}, 2024.

\bibitem[Silva-Rodriguez et~al.(2024)Silva-Rodriguez, Hajimiri, Ben~Ayed, and Dolz]{silva2024closerCLAP}
Julio Silva-Rodriguez, Sina Hajimiri, Ismail Ben~Ayed, and Jose Dolz.
\newblock A closer look at the few-shot adaptation of large vision-language models.
\newblock In \emph{Proceedings of the IEEE/CVF Conference on Computer Vision and Pattern Recognition}, pages 23681--23690, 2024.

\bibitem[Soomro et~al.(2012)Soomro, Zamir, and Shah]{soomro2012ucf101}
Khurram Soomro, Amir~Roshan Zamir, and Mubarak Shah.
\newblock Ucf101: A dataset of 101 human actions classes from videos in the wild.
\newblock \emph{arXiv preprint arXiv:1212.0402}, 2012.

\bibitem[Xiao et~al.(2010)Xiao, Hays, Ehinger, Oliva, and Torralba]{xiao2010sun}
Jianxiong Xiao, James Hays, Krista~A Ehinger, Aude Oliva, and Antonio Torralba.
\newblock Sun database: Large-scale scene recognition from abbey to zoo.
\newblock In \emph{2010 IEEE computer society conference on computer vision and pattern recognition}, pages 3485--3492. IEEE, 2010.

\bibitem[Yao et~al.(2023)Yao, Zhang, and Xu]{yao2023visualKgCoCoOp}
Hantao Yao, Rui Zhang, and Changsheng Xu.
\newblock Visual-language prompt tuning with knowledge-guided context optimization.
\newblock In \emph{Proceedings of the IEEE/CVF conference on computer vision and pattern recognition}, pages 6757--6767, 2023.

\bibitem[Yoon et~al.(2024)Yoon, Yoon, Tee, Hasegawa-Johnson, Li, and Yoo]{yoon2024ctpt}
Hee~Suk Yoon, Eunseop Yoon, Joshua Tian~Jin Tee, Mark~A. Hasegawa-Johnson, Yingzhen Li, and Chang~D. Yoo.
\newblock C-{TPT}: Calibrated test-time prompt tuning for vision-language models via text feature dispersion.
\newblock In \emph{The Twelfth International Conference on Learning Representations (ICLR)}, 2024.

\bibitem[Yu et~al.(2023)Yu, Lu, Jin, Chen, and Wang]{yu2023task}
Tao Yu, Zhihe Lu, Xin Jin, Zhibo Chen, and Xinchao Wang.
\newblock Task residual for tuning vision-language models.
\newblock In \emph{Proceedings of the IEEE/CVF conference on computer vision and pattern recognition}, pages 10899--10909, 2023.

\bibitem[Zanella and Ben~Ayed(2024)]{zanella2024lowclip-lora}
Maxime Zanella and Ismail Ben~Ayed.
\newblock Low-rank few-shot adaptation of vision-language models.
\newblock In \emph{Proceedings of the IEEE/CVF Conference on Computer Vision and Pattern Recognition Workshops}, pages 1593--1603, 2024.

\bibitem[Zanella et~al.(2024)Zanella, G{\'e}rin, and Ayed]{zanella2024boosting}
Maxime Zanella, Beno{\^\i}t G{\'e}rin, and Ismail Ayed.
\newblock Boosting vision-language models with transduction.
\newblock \emph{Advances in Neural Information Processing Systems}, 37:\penalty0 62223--62256, 2024.

\bibitem[Zanella et~al.(2025)Zanella, Fuchs, De~Vleeschouwer, and Ben~Ayed]{zanella2025realistic}
Maxime Zanella, Cl{\'e}ment Fuchs, Christophe De~Vleeschouwer, and Ismail Ben~Ayed.
\newblock Realistic test-time adaptation of vision-language models.
\newblock In \emph{Proceedings of the Computer Vision and Pattern Recognition Conference}, pages 25103--25112, 2025.

\bibitem[Zhang et~al.(2022)Zhang, Zhang, Fang, Gao, Li, Dai, Qiao, and Li]{zhang2022tip}
Renrui Zhang, Wei Zhang, Rongyao Fang, Peng Gao, Kunchang Li, Jifeng Dai, Yu~Qiao, and Hongsheng Li.
\newblock Tip-adapter: Training-free adaption of clip for few-shot classification.
\newblock In \emph{European conference on computer vision}, pages 493--510. Springer, 2022.

\bibitem[Zhang et~al.(2025)Zhang, Deng, Guo, and Hu]{zhang2025adaptiveclip-ast}
Yi~Zhang, Yi-Xuan Deng, Meng-Hao Guo, and Shi-Min Hu.
\newblock Adaptive parameter selection for tuning vision-language models.
\newblock In \emph{Proceedings of the Computer Vision and Pattern Recognition Conference}, pages 4280--4290, 2025.

\bibitem[Zhao et~al.(2024)Zhao, Jiang, Wang, Xu, and Tang]{zhao2024hegraphadapter}
Yumiao Zhao, Bo~Jiang, Xiao Wang, Qin Xu, and Jin Tang.
\newblock Hegraphadapter: Tuning multi-modal vision-language models with heterogeneous graph adapter.
\newblock \emph{arXiv preprint arXiv:2410.07854}, 2024.

\bibitem[Zhong et~al.(2022)Zhong, Yang, Zhang, Li, Codella, Li, Zhou, Dai, Yuan, Li, et~al.]{zhong2022regionclip}
Yiwu Zhong, Jianwei Yang, Pengchuan Zhang, Chunyuan Li, Noel Codella, Liunian~Harold Li, Luowei Zhou, Xiyang Dai, Lu~Yuan, Yin Li, et~al.
\newblock Regionclip: Region-based language-image pretraining.
\newblock In \emph{Proceedings of the IEEE/CVF conference on computer vision and pattern recognition}, pages 16793--16803, 2022.

\bibitem[Zhou et~al.(2022{\natexlab{a}})Zhou, Yang, Loy, and Liu]{zhou2022conditionalCoCoOp}
Kaiyang Zhou, Jingkang Yang, Chen~Change Loy, and Ziwei Liu.
\newblock Conditional prompt learning for vision-language models.
\newblock In \emph{Proceedings of the IEEE/CVF conference on computer vision and pattern recognition}, pages 16816--16825, 2022{\natexlab{a}}.

\bibitem[Zhou et~al.(2022{\natexlab{b}})Zhou, Yang, Loy, and Liu]{zhou2022learningcoop}
Kaiyang Zhou, Jingkang Yang, Chen~Change Loy, and Ziwei Liu.
\newblock Learning to prompt for vision-language models.
\newblock \emph{International Journal of Computer Vision}, 130\penalty0 (9):\penalty0 2337--2348, 2022{\natexlab{b}}.

\bibitem[Zhu et~al.(2023)Zhu, Niu, Han, Wu, and Zhang]{zhu2023promptProGrad}
Beier Zhu, Yulei Niu, Yucheng Han, Yue Wu, and Hanwang Zhang.
\newblock Prompt-aligned gradient for prompt tuning.
\newblock In \emph{Proceedings of the IEEE/CVF international conference on computer vision}, pages 15659--15669, 2023.

\end{thebibliography}
}
\clearpage


\appendix

\newpage
\appendix
\onecolumn
Additional results, ablation studies, analyzes, visualization, and computational complexity that could not be accommodated in the main paper due to space constraints are provided in this Appendix.
\section{Additional Implementation Details}
\label{app:implementation_details}

This appendix provides implementation details for reproducing SD-LoRA. The main paper focuses on the experimental findings and analysis; here we describe support-set construction, preprocessing, LoRA adaptation, cache construction, the training-time graph teacher, and optimization.

\paragraph{Support-set construction.}
For each dataset, we follow the standard inductive few-shot protocol used in prior CLIP adaptation work~\cite{zhou2022learningcoop,zhang2022tip,zanella2024lowclip-lora}. Given a $C$-class dataset, we randomly sample $K\in\{1,2,4,8,16\}$ labeled examples per class from the training split to form the support set $\mathcal{S}=\{(x_i,y_i)\}_{i=1}^{CK}$. The resulting support set has $CK$ labeled images. No class-manifold search or unlabeled test-time information is used. For each shot setting, we follow the standard inductive few-shot protocol and repeat the experiment over three independently sampled support splits, and report the mean top-1 accuracy.

\paragraph{Preprocessing and prompts.}
We use the standard CLIP image preprocessing pipeline. During adaptation, images are resized and augmented following common few-shot CLIP practice~\cite{zhang2022tip,zanella2024lowclip-lora}; at evaluation time, deterministic resizing and center cropping are used. Text embeddings are produced from class-name prompts using the CLIP text encoder. Unless otherwise stated, we use the same prompt templates as prior CLIP few-shot adaptation methods to ensure a fair comparison.

\paragraph{LoRA adaptation.}
The backbone is CLIP ViT-B/16~\cite{radford2021learningclip}. The pretrained CLIP parameters are frozen, and adaptation is performed using low-rank updates. For a pretrained projection matrix $W_0$, LoRA represents the adapted weight as
\begin{equation}
    W = W_0 + \Delta W,
    \qquad
    \Delta W = BA,
\end{equation}
where $A$ and $B$ are trainable low-rank matrices. Only the LoRA parameters, the cache head, and the training-only teacher parameters are optimized during adaptation. The vision and text encoder remains frozen throughout. This keeps the deployed model lightweight while allowing the visual representation to be shaped by structural supervision during training.

\paragraph{Cache construction.}
For each support image $x_i$, we compute the normalized LoRA-adapted image feature
\begin{equation}
    z_{\theta}(x_i)=
    \frac{f_v^{\theta}(x_i)}
    {\|f_v^{\theta}(x_i)\|_2}.
\end{equation}
The support features form the cache keys, while their one-hot class labels form the cache values. For a query image $x$, the cache branch computes similarities between the query feature $z_{\theta}(x)$ and the support keys, followed by an affinity transformation controlled by the cache sharpness parameter. The resulting cache logits are combined with the zero-shot CLIP logits to obtain the deployed student prediction. After adaptation, the cache keys are recomputed once using the final LoRA-adapted encoder and then kept fixed for inference.

\paragraph{Training-time structural teacher.}
The graph teacher is used only during adaptation. It takes high-resolution ViT visual tokens from the same LoRA-adapted visual backbone as the deployed student, along with frozen class-text embeddings. Positional embeddings are interpolated for higher spatial resolutions. The teacher builds a heterogeneous graph with visual nodes (\texttt{[CLS]} and patch tokens) and class-text nodes (class-level text embeddings).

The graph has three relation types: \emph{patch--patch}, \emph{global--local}, and \emph{visual--text}. Patch--patch edges capture local spatial interactions among neighboring tokens. Global--local edges link the \texttt{[CLS]} token to patch tokens, allowing image-level context to influence local evidence. Visual--text edges connect visual tokens to class-text embeddings, enabling class-conditioned structural reasoning. The graph is refined via relation-aware message passing, with node-type-specific transformations controlled by $\phi(i)$ and relation-specific interactions by $r=\mathrm{rel}(j,i)$.

\paragraph{Structural readout.}
After graph refinement, we obtain updated visual and text node representations. To prevent background tokens from overwhelming local evidence, we apply a saliency-based readout over patch nodes: each refined patch node receives a learnable score, we keep the top-$\kappa$ fraction, average them, and combine this with the refined \texttt{[CLS]} representation to form the graph-conditioned representation $h_g$. This representation is used only for training losses and is not computed at inference.

\paragraph{Training objective.}
The overall objective combines three terms:
\[
\mathcal{L}_{\mathrm{train}}
=
\mathcal{L}_{\mathrm{Student}}
+
\lambda_1 \mathcal{L}_{\mathrm{Teacher}}
+
\lambda_2 \mathcal{L}_{\mathrm{PPA}}.
\]
The student loss uses cross-entropy on the deployed CLIP-LoRA$+$cache logits. The teacher loss applies focal loss to the graph logits, down-weighting confident examples and emphasizing low-confidence or misclassified ones. Prototype Predictive Alignment (PPA) encourages the graph-conditioned representation to predict the support-derived prototype of the ground-truth class. Because the student and teacher share the LoRA-adapted backbone, gradients from the teacher and PPA losses update the same LoRA parameters used at inference.

\paragraph{Evaluation.}
All aggregate values are computed directly from the per-dataset results using unweighted arithmetic means. For base-to-novel evaluation, harmonic mean is computed as
$\mathrm{HM}=2BN/(B+N)$ from the reported base and novel accuracies.


\paragraph{Optimization.}
We optimize with AdamW, an initial learning rate of $1\times10^{-5}$, and cosine learning-rate decay, using the same settings across datasets unless noted. To prevent test-set leakage, all hyperparameters are chosen only from training/validation data or via support-split cross-validation; test labels are never used. Hyperparameter sensitivity is reported in Appendix~\ref{app:compute_complex}.

\paragraph{Training-time structural transfer and deployment.}
Algorithm~\ref{alg:sd_lora} summarizes the asymmetric train--deploy design. The structural branch and student jointly update the same LoRA parameters; there is no independently pretrained frozen teacher or teacher--student logit-matching loss.
\begin{algorithm}[H]
\caption{SD-LoRA: Training-Time Structural Transfer and Lightweight Deployment}
\label{alg:sd_lora}
\small
\begin{algorithmic}[1]
\Require Labeled support set $\mathcal S$, pretrained CLIP backbone, frozen class-text embeddings $T$, selected training and cache settings
\Ensure Adapted $224\times224$ CLIP-LoRA+cache predictor
\State Initialize shared LoRA parameters $\theta$, cache parameters $\eta$, and auxiliary teacher/PPA parameters
\For{each support-set optimization step}
    \State Compute student logits using the LoRA backbone and support-derived cache \Comment{Eq.~\ref{eq:student_logits}}
    \State Extract visual tokens and construct patch--patch, global--local, and visual--text relations
    \State Compute graph-conditioned representations and teacher logits
    \State Align graph-conditioned evidence with support-derived class prototypes \Comment{Eq.~\ref{eq:ppa_prototype}}
    \State $\mathcal L_{\rm struct}\gets\lambda_1\mathcal L_{\rm Teacher}+\lambda_2\mathcal L_{\rm PPA}$
    \State Optimize $\mathcal L_{\rm Student}+\mathcal L_{\rm struct}$, updating shared $\theta$ \Comment{Eqs.~\ref{eq:implicit_gradient},~\ref{eq:training_objective}}
\EndFor
\State Discard the graph teacher, saliency/readout, PPA head, and high-resolution path
\State Recompute the support cache once using the final $224\times224$ LoRA encoder
\State \Return the fixed LoRA+cache predictor $s_{\theta,\eta}(x)=L_{\rm ZS}(x)+\omega L_{\rm Cache}(x)$
\end{algorithmic}
\end{algorithm}
The structurally supervised updates to $\theta$ remain in the deployed encoder even though auxiliary parameters are discarded. Each test image is processed independently by a single LoRA image-encoder pass and the cache classifier, without graph construction or high-resolution processing. Code repository: \url{https://github.com/MR-Sherif/SD-LoRA}.

\paragraph{Inference.}
At test time, all teacher-specific components are removed, including the relation-aware graph transformer, high-resolution graph-token processing, structural readout, and prototype prediction head. Inference uses only the LoRA-adapted CLIP image encoder, frozen CLIP text embeddings, and the cache-based classifier:
\[
    s_{\theta,\eta}(x)
    =
    L_{\mathrm{ZS}}(x)
    +
    \omega L_{\mathrm{Cache}}(x).
\]
Thus, SD-LoRA introduces structural reasoning only during training while keeping inference lightweight, with no graph computation at deployment.



\section{Comparison to the State-of-the-art (SotA): Complete Table}
\label{app:full_fewshot_results}

\paragraph{Benchmarks.} We evaluate our approach on 11 standard few-shot image classification benchmarks widely used in CLIP research: ImageNet~\cite{deng2009imagenet}, Caltech101~\cite{fei2004learning}, OxfordPets~\cite{parkhi2012cats}, StanfordCars~\cite{krause20133d}, Flowers102~\cite{nilsback2008automatedflowers}, Food101~\cite{bossard2014food}, FGVC-Aircraft~\cite{maji2013fineaircraft}, SUN397~\cite{xiao2010sun}, DTD~\cite{cimpoi2014describing}, EuroSAT~\cite{helber2019eurosat}, and UCF101~\cite{soomro2012ucf101}, covering general object recognition, fine-grained categorization, textures, scene understanding, remote sensing, and action recognition.

\paragraph{Baselines.} We compare our method with two main streams of state-of-the-art vision-language adaptation techniques. The first includes adapter-based methods: Tip-Adapter~\cite{zhang2022tip}, TaskRes~\cite{yu2023task}, GraphAdapter~\cite{li2023graphadapter}, CLIP-Adapter~\cite{gao2024clip-adapter}, CLAP~\cite{silva2024closerCLAP}, and CCA~\cite{jiang2025causalcca}. The second includes prompt-based and fine-tuning methods: CoOp~\cite{zhou2022learningcoop}, CoCoOp~\cite{zhou2022conditionalCoCoOp}, PLOT++~\cite{chen2022promptplot++}, KgCoOp~\cite{yao2023visualKgCoCoOp}, MaPLe~\cite{khattak2023maple}, ProGrad~\cite{zhu2023promptProGrad}, PromptSRC~\cite{khattak2023self}, CLIP-LoRA~\cite{zanella2024lowclip-lora}, and CLIP-AST~\cite{zhang2025adaptiveclip-ast}.

\paragraph{Discussion of full few-shot results.}
Tables~\ref{tab:comparison_full_1_4_16} and~\ref{tab:comparison_full_2_8} present per-dataset results corresponding to the average comparison in Table~\ref{tab:main_avg_results}. SD-LoRA consistently improves performance across shot settings and dataset types, with especially strong gains on fine-grained and domain-specific benchmarks such as Aircraft, EuroSAT, DTD, Flowers102, and UCF101. These datasets rely on localized shape, texture, or class-specific cues, highlighting the benefit of the training-only graph teacher in shaping the deployed LoRA representation. Relative to CLIP-LoRA, SD-LoRA preserves the same lightweight inference design but achieves higher accuracy via structural supervision during adaptation. The results also show that SD-LoRA is complementary to adaptive LoRA placement methods such as CLIP-AST: layer selection controls where LoRA is inserted, whereas SD-LoRA controls how its parameters are trained. We include the full tables here for completeness; the main paper reports only compact averages and per-dataset trends to save space.

\begin{table*}[!ht]
\centering
\caption{
\textbf{Full few-shot comparison on 11 benchmarks.}
We report top-1 accuracy (\%) for 1-, 2-, and 4-shot classification against representative adapter- and prompt-based CLIP adaptation methods. The best result in each shot group is shown in \textbf{bold}. SD-LoRA is reported in the final row of each shot group. Dataset abbreviations: INet: ImageNet; SUN: SUN397; Air: FGVC-Aircraft; Euro: EuroSAT; Cars: StanfordCars; Food: Food101; Pets: OxfordPets; Flow: Flowers102; Cal: Caltech101; DTD: Describable Textures; UCF: UCF101.
}
\label{tab:comparison_full_1_4_16}
\scriptsize 
\setlength{\tabcolsep}{4pt} 
\begin{tabular}{@{}l c l l r r r r r r r r r r r r@{}}
\toprule
Shots & Type & Method & Venue & INet & SUN & Air & Euro & Cars & Food & Pets & Flow & Cal & DTD & UCF & Avg \\
\midrule
0 & & CLIP & ICML'22 & 66.7 & 71.4 & 62.6 & 24.7 & 47.5 & 65.3 & 86.1 & 89.1 & 92.9 & 43.6 & 66.7 & 65.1 \\
\midrule
\multirow{17}{*}{1}
& \multirow{6}{*}{\rotatebox{90}{\textbf{Adapter}}}
& TIP-Adapter-F & ECCV'22 & 69.4 & 67.2 & 28.8 & 67.8 & 67.1 & 85.8 & 90.6 & 83.8 & 94.0 & 51.6 & 73.4 & 70.9 \\
& & TaskRes & CVPR'23 & 69.6 & 68.1 & 31.3 & 65.4 & 68.8 & 84.6 & 90.2 & 81.7 & 93.6 & 53.8 & 71.7 & 70.8 \\
& & GraphAdapter & NeurIPS'23 & 61.5 & 61.9 & 20.9 & 63.3 & 59.7 & 75.4 & 84.4 & 80.0 & 88.9 & 51.8 & 64.9 & 64.8 \\
& & CLIP-Adapter & IJCV'24 & 67.9 & 71.3 & 25.2 & 65.4  & 49.3 & 65.7 & 86.1 & 89.0 & 92.0 & 44.2 & 66.9 & 65.7 \\
& & CLAP & CVPR'24 & 58.5 & 61.1 & 20.6 & 59.2 & 56.3 & 73.0 & 83.6 & 79.9 & 88.4 & 47.5 & 62.5 & 62.8 \\
& & CCA & ICCV'25 & 61.5 & 63.8 & 22.5 & 67.0 & 60.0 & 77.8 & 86.9 & 81.0 & 89.9 & 51.0 & 66.3 & 66.3 \\
\cmidrule(lr){2-16}

& \multirow{11}{*}{\rotatebox{90}{\textbf{Prompt}}}
& CoOp (4) & IJCV'22 & 68.0 & 67.3 & 26.2 & 50.9 & 67.1 & 82.6 & 90.3 & 72.7 & 93.2 & 50.1 & 70.7 & 67.2 \\
& & CoOp (16) & IJCV'22 & 65.7 & 67.0 & 20.8 & 56.4 & 67.5 & 84.3 & 90.2 & 78.3 & 92.5 & 50.1 & 71.2 & 67.6 \\
& & CoCoOp & CVPR'22 & 69.4 & 68.7 & 28.1 & 55.4 & 67.6 & 84.9 & 91.9 & 73.4 & 94.1 & 52.6 & 70.4 & 68.8 \\
& & PLOT++ & ICLR'23 & 66.4 & 66.7 & 28.6 & 65.4 & 68.8 & 86.1 & 91.8 & 80.4 & 94.3 & 54.5 & 74.3 & 70.7 \\
& & KgCoOp & CVPR'23 & 68.9 & 68.4 & 26.8 & 61.9 & 66.7 & 86.4 & 92.1 & 74.7 & 94.2 & 52.7 & 72.8 & 69.6 \\
& & MaPLe & CVPR'23 & 69.7 & 69.3 & 28.1 & 29.1 & 67.6 & 85.4 & 91.4 & 74.9 & 93.6 & 50.0 & 71.1 & 66.4 \\
& & ProGrad & ICCV'23 & 67.0 & 67.0 & 28.8 & 57.0 & 68.2 & 84.9 & 91.4 & 80.9 & 93.5 & 52.8 & 73.3 & 69.5 \\
& & PromptSRC & ICCV'23 & 68.5 & 68.5 & 27.4 & 72.8 & 68.7 & 85.3 & 90.5 & 85.9 & 93.8 & 56.7 & 74.8 & 72.1 \\
& & CLIP-LoRA & CVPR'24 & 70.4 & 70.4 & 30.2 & 72.3 & 70.1 & 84.3 & 92.3 & 83.2 & 93.7 & 54.3 & 76.3 & 72.5 \\
& & CLIP-AST & CVPR'25 & 70.1 & 70.3 & 31.0 & 68.0 & 70.7 & 86.0 & 92.3 & 83.0 & 94.1 & 56.0 & 74.0 & 72.3 \\

\cmidrule(lr){3-16}
& & \textbf{SD-LoRA (Ours)} & --- & \cellcolor{blue!15} \textbf{70.5} & \cellcolor{blue!15}\textbf{71.6} & \cellcolor{blue!15}\textbf{34.8} & \cellcolor{blue!15}\textbf{76.8} & \cellcolor{blue!15}\textbf{70.9} & \cellcolor{blue!15}86.2 & \cellcolor{blue!15}92.5 & \cellcolor{blue!15}\textbf{88.2} & \cellcolor{blue!15}\textbf{94.4} & \cellcolor{blue!15}\textbf{59.2} & \cellcolor{blue!15}\textbf{78.5} & \cellcolor{blue!15}\textbf{74.9} \\

\midrule

\multirow{17}{*}{2}
& \multirow{6}{*}{\rotatebox{90}{\textbf{Adapter}}}
& TIP-Adapter-F & ECCV'22 & 70.0 & 68.6 & 32.8 & 73.2 & 70.8 & 86.0 & 91.6 & 90.1 & 93.9 & 57.8 & 76.2 & 73.7 \\
& & TaskRes & CVPR'23 & 70.2 & 70.5 & 32.7 & 70.2 & 72.1 & 85.6 &  90.7 & 84.4 & 94.3 & 55.6 & 75.2 & 72.9 \\
& & GraphAdapter & NeurIPS'23 & 62.3 & 64.6 & 23.8 & 67.3 & 63.2 & 76.3 & 86.3 & 85.6 & 90.2 & 55.7 & 69.5 & 67.7 \\
& & CLIP-Adapter & IJCV'24 & 68.2 & 67.2 & 27.0 & 51.2 & 66.6 & 86.2 & 89.7 & 71.7 & 93.4 & 45.4 & 68.4 & 66.8 \\
& & CLAP & CVPR'24 & 58.5 & 63.3 & 23.2 & 65.6 & 61.4 & 74.9 & 84.9 & 84.2 & 89.8 & 53.0 & 67.8 & 66.1 \\
& & CCA & ICCV'25 & 62.1 & 66.3 & 25.0 & 70.0 & 64.0 & 77.9 & 87.9 & 88.0 & 91.0 & 55.0 & 69.3 & 68.9 \\
\cmidrule(lr){2-16}

& \multirow{11}{*}{\rotatebox{90}{\textbf{Prompt}}}
& CoOp (4) & IJCV'22 & 68.7 & 68.0 & 28.1 & 66.2 & 70.5 & 82.6 & 89.9 & 80.9 & 93.0 & 53.7 & 73.5 & 70.5 \\
& & CoOp (16) & IJCV'22 & 67.0 & 67.0 & 25.9 & 65.1 & 70.4 & 84.4 & 89.9 & 88.0 & 93.1 & 54.1 & 74.1 & 70.8 \\
& & CoCoOp & CVPR'22 & 70.1 & 69.4 & 29.3 & 61.8 & 68.4 & 85.9 & 91.9 & 77.8 & 94.4 & 52.3 & 73.4 & 70.4 \\
& & PLOT++ & ICLR'23 & 68.2 & 68.0 & 31.1 & 76.8 & 73.1 & 86.3 & 92.2 & 89.8 & 94.6 & 56.7 & 76.7 & 74.0 \\
& & KgCoOp & CVPR'23 & 69.6 & 69.6 & 28.0 & 69.2 & 68.2 & 86.6 & 92.3 & 79.8 & 94.5 & 55.3 & 74.6 & 71.6 \\
& & MaPLe & CVPR'23 & 70.0 & 70.7 & 29.5 & 59.4 & 68.5 & 86.5 & 91.8 & 79.8 & 94.9 & 50.6 & 74.0 & 70.5 \\
& & ProGrad & ICCV'23 & 69.1 & 69.0 & 31.1 & 66.3 & 72.4 & 84.8 & 91.5 & 87.5 & 93.6 & 56.0 & 75.6 & 72.4 \\
& & PromptSRC & ICCV'23 & 69.8 & 71.3 & 31.5 & 79.2 & 72.9 & 86.0 & 92.3 & 91.1 & 94.6 & 60.0 & 78.6 & 75.2 \\
& & CLIP-LoRA & CVPR'24 & 70.8 & 71.3 & 33.2 & 82.7 & 73.2 & 83.2 & 91.3 & 89.8 & 94.6 & 59.9 & 80.0 & 75.4 \\
& & CLIP-AST & CVPR'25 & \textbf{71.1} & 71.5 & 35.0 & 82.0 & 73.5 & 86.1 & \textbf{92.9} & 90.0 & \textbf{95.9} & 58.0 & 78.0 & 75.8 \\

\cmidrule(lr){3-16}
& & \textbf{SD-LoRA (Ours)} & --- & \cellcolor{blue!15} 70.7 & \cellcolor{blue!15}\textbf{71.6} & \cellcolor{blue!15}\textbf{36.4} & \cellcolor{blue!15}\textbf{84.9} & \cellcolor{blue!15}\textbf{73.5} & \cellcolor{blue!15}86.8 & \cellcolor{blue!15}92.4 & \cellcolor{blue!15}\textbf{91.7} & \cellcolor{blue!15}95.8 & \cellcolor{blue!15}\textbf{64.6} & \cellcolor{blue!15}\textbf{80.8} & \cellcolor{blue!15}\textbf{77.2} \\

\midrule
\multirow{17}{*}{4}
& \multirow{6}{*}{\rotatebox{90}{\textbf{Adapter}}}
& TIP-Adapter-F & ECCV'22 & 70.7 & 70.8 & 35.7 & 76.8 & 74.1 & 86.5 & 91.9 & 92.1 & 94.8 & 59.8 & 78.1 & 75.6 \\
& & TaskRes & CVPR'23 & 63.6 & 67.3 & 25.7 & 73.8 & 67.4 & 76.1 & 86.3 & 90.2 & 91.0 & 60.7 & 70.9 & 70.3 \\
& & GraphAdapter & NeurIPS'23 & 63.1 & 66.7 & 27.0 & 75.2 & 66.5 & 76.8 & 86.6 & 89.9 & 91.0 & 59.6 & 71.5 & 70.3 \\
& & CLIP-Adapter & IJCV'24 & 71.0 & 72.7 & 33.4 & 74.2 & 76.0 &  86.0 & 91.9 &85.0 & 95.0 & 60.1 & 76.2 & 74.7 \\
& & CLAP & CVPR'24 & 60.7 & 65.9 & 25.6 & 73.1 & 65.5 & 75.9 & 86.5 & 87.6 & 90.6 & 58.8 & 69.8 & 69.1 \\
& & CCA & ICCV'25 & 63.3 & 69.0 & 29.0 & 80.1 & 68.0 & 78.2 & 88.1 & 91.1 & 92.0 & 63.0 & 72.1 & 72.2 \\

\cmidrule(lr){2-16}

& \multirow{11}{*}{\rotatebox{90}{\textbf{Prompt}}}
& CoOp (4) & IJCV'22 & 69.7 & 70.6 & 29.7 & 65.8 & 73.4 & 83.5 & 92.3 & 86.6 & 94.5 & 58.5 & 78.1 & 73.0 \\
& & CoOp (16) & IJCV'22 & 70.6 & 69.7 & 30.9 & 69.7 & 74.4 & 84.5 & 92.5 & 92.2 & 94.5 & 59.5 & 77.6 & 74.2 \\
& & CoCoOp & CVPR'22 & 70.8 & 70.4 & 30.6 & 61.7 & 69.5 & 86.3 & 92.7 & 81.5 & 94.8 & 55.7 & 75.3 & 71.8 \\
& & PLOT++ & ICLR'23 & 70.4 & 71.7 & 35.2 & 83.2 & 76.2 & 86.4 & 92.5 & 92.9 & 95.1 & 62.4 & 79.7 & 76.9 \\
& & KgCoOp & CVPR'23 & 70.2 & 71.5 & 32.2 & 71.8 & 69.5 & 86.9 & 92.6 & 87.0 & 95.0 & 58.7 & 77.6 & 73.9 \\
& & MaPLe & CVPR'23 & 67.7 & 70.6 & 34.8 & 84.5 & 75.3 & 81.7 & 91.9 & 92.6 & 94.4 & 61.0 & 78.4 & 75.8 \\
& & ProGrad & ICCV'23 & 71.3 & 71.7 & 34.1 & 69.6 & 75.0 & 85.4 & 92.1 & 91.1 & 94.4 & 59.7 & 77.9 & 73.9 \\
& & PromptSRC & ICCV'23 & 70.8 & 73.8 & 37.4 & 86.3 & 77.0 & 86.6 & 93.1 & 93.8 & 95.3 & 65.4 & 81.6 & 78.3 \\
& & CLIP-LoRA & CVPR'24 & 71.4 & 72.8 & 37.9 & 84.9 & 77.4 & 82.7 & 91.0 & 93.7 & 95.2 & 63.8 & 81.1 & 77.4 \\
& & CLIP-AST & CVPR'25 & 71.1 & 73.4 & 41.0 & 88.0 & 78.0 & 86.8 & \textbf{93.5} & 94.0 & 95.7 & 65.0 & 81.0 & 78.9 \\

\cmidrule(lr){3-16}
& & \textbf{SD-LoRA (Ours)} & --- & \cellcolor{blue!15}\textbf{72.2} & \cellcolor{blue!15}\textbf{73.5} & \cellcolor{blue!15}\textbf{46.2} & \cellcolor{blue!15}\textbf{91.3} & \cellcolor{blue!15}\textbf{78.2} & \cellcolor{blue!15}\textbf{87.1} & \cellcolor{blue!15}92.7 & \cellcolor{blue!15}\textbf{94.3} & \cellcolor{blue!15}\textbf{96.2} & \cellcolor{blue!15}\textbf{70.1} & \cellcolor{blue!15}\textbf{82.6} & \cellcolor{blue!15}\textbf{80.4} \\
\bottomrule
\end{tabular}
\end{table*}
\begin{table*}[ht]
\centering
\caption{
\textbf{Full 8- and 16-shot comparison on 11 benchmarks.}
We report top-1 accuracy (\%) against representative adapter- and prompt-based CLIP adaptation methods. The best result in each shot group is shown in \textbf{bold}. SD-LoRA is reported in the final row of each shot group. Dataset abbreviations follow Table~\ref{tab:comparison_full_1_4_16}.
}
\label{tab:comparison_full_2_8}
\scriptsize 
\setlength{\tabcolsep}{4pt} 
\begin{tabular}{@{}l c l l r r r r r r r r r r r r@{}}
\toprule
Shots & Type & Method & Venue & INet & SUN & Air & Euro & Cars & Food & Pets & Flow & Cal & DTD & UCF & Avg \\
\midrule

\multirow{17}{*}{8}
& \multirow{6}{*}{\rotatebox{90}{\textbf{Adapter}}}
& TIP-Adapter-F & ECCV'22 & 71.7 & 73.5 & 39.5 & 81.3 & 78.3 & 86.9 & 91.8 & 94.3 & 95.2 & 66.7 & 82.0 & 78.3 \\
& & TaskRes & CVPR'23 & 72.3 & 74.6 & 40.3 & 77.5 & 79.6 & 86.4 & 92.0 & 96.0 & 95.3 & 66.7 & 81.6 & 78.4 \\
& & GraphAdapter & NeurIPS'23 & 64.2 & 68.9 & 31.4 & 80.2 & 70.5 & 77.7 & 87.6 & 94.1 & 92.4 & 64.5 & 75.7 & 73.4 \\
& & CLIP-Adapter & IJCV'24 & 62.7 & 67.5 & 26.2 & 77.9 & 67.9 & 78.0 & 87.6 & 91.7 & 91.4 & 61.0 & 73.3 & 71.4 \\
& & CLAP & CVPR'24 & 62.9 & 68.6 & 28.9 & 76.7 & 70.3 & 77.4 & 87.7 & 92.1 & 91.4 & 63.2 & 73.3 & 72.1 \\
& & CCA & ICCV'25 & 64.9 & 71.0 & 35.0 & 83.1 & 74.3 & 79.2 & 89.0 & 94.4 & 92.5 & 65.0 & 77.5 & 75.0 \\

\cmidrule(lr){2-16}

& \multirow{11}{*}{\rotatebox{90}{\textbf{Prompt}}}
& CoOp (4) & IJCV'22 & 70.8 & 72.4 & 37.0 & 74.7 & 76.8 & 83.3 & 92.1 & 95.0 & 94.7 & 63.7 & 79.8 & 76.4 \\
& & CoOp (16) & IJCV'22 & 70.6 & 71.9 & 38.5 & 77.1 & 79.0 & 82.7 & 91.3 & 94.9 & 94.5 & 64.8 & 80.0 & 76.8 \\
& & CoCoOp & CVPR'22 & 70.8 & 71.5 & 32.4 & 69.1 & 70.4 & 87.0 & 93.3 & 86.3 & 94.9 & 60.1 & 75.9 & 73.8 \\
& & PLOT++ & ICLR'23 & 71.3 & 73.9 & 41.4 & 88.3 & 81.2 & 86.5 & 93.0 & 95.4 & 95.5 & 66.4 & 82.8 & 79.6 \\
& & KgCoOp & CVPR'23 & 70.2 & 72.6 & 34.8 & 73.9 & 72.8 & 87.0 & 93.0 & 91.5 & 95.1 & 65.6 & 80.0 & 76.0 \\
& & MaPLe & CVPR'23 & 71.3 & 73.2 & 33.8 & 82.8 & 71.3 & 87.2 & 93.1 & 90.5 & 95.1 & 63.0 & 79.5 & 76.4 \\
& & ProGrad & ICCV'23 & 71.3 & 73.0 & 37.7 & 77.8 & 78.7 & 86.1 & 92.2 & 95.0 & 94.8 & 63.9 & 80.5 & 77.4 \\
& & PromptSRC & ICCV'23 & 71.7 & 75.8 & 43.0 & 88.8 & 80.9 & 86.9 & 93.2 & 96.1 & 95.7 & 69.8 & 84.5 & 80.6 \\
& & CLIP-LoRA & CVPR'24 & 72.3 & 74.7 & 45.7 & 89.7 & 82.1 & 83.1 & 91.7 & 96.3 & 95.6 & 67.5 & 84.1 & 80.3 \\
& & CLIP-AST & CVPR'25 & 72.9 & 75.4 & 50.0 & 90.0 & 83.0 & 87.0 & \textbf{94.5} & 96.5 & 96.1 & 70.1 & 84.0 & 81.7 \\
\cmidrule(lr){3-16}

& & \textbf{SD-LoRA (Ours)} & --- & \cellcolor{blue!15} \textbf{73.2} & \cellcolor{blue!15}\textbf{75.6} & \cellcolor{blue!15}\textbf{53.2} & \cellcolor{blue!15}\textbf{92.7} & \cellcolor{blue!15}\textbf{84.4} & \cellcolor{blue!15}\textbf{87.4} & \cellcolor{blue!15}93.1 & \cellcolor{blue!15}\textbf{96.8} & \cellcolor{blue!15}\textbf{96.3} & \cellcolor{blue!15}\textbf{72.2} & \cellcolor{blue!15}\textbf{84.8} & \cellcolor{blue!15}\textbf{82.7} \\

\midrule
\multirow{17}{*}{16}
& \multirow{6}{*}{\rotatebox{90}{\textbf{Adapter}}}
& TIP-Adapter-F & ECCV'22 & 73.4 & 76.0 & 44.6 & 85.9 & 82.3 & 86.8 & 92.6 & 96.2 & 95.7 & 70.8 & 83.9 & 80.7 \\
& & TaskRes & CVPR'23 & 73.0 & 76.1 & 44.9 & 82.7 & 83.5 & 86.9 & 92.4 & 97.5 & 95.8 & 71.5 & 84.0 & 80.8 \\
& & GraphAdapter & NeurIPS'23 & 65.7 & 71.2 & 36.9 & 85.3 & 76.2 & 78.6 & 88.6 & 96.2 & 93.3 & 67.6 & 78.8 & 76.2 \\
& & CLIP-Adapter & IJCV'24 & 69.8 & 74.2 & 34.2 & 71.4 & 74.0 & 87.1 & 92.3 & 92.9 & 94.9 & 59.4 & 80.2 & 75.5 \\
& & CLAP & CVPR'24 & 65.0 & 70.8 & 33.6 & 80.1 & 75.1 & 78.5 & 88.5 & 94.2 & 91.9 & 66.4 & 76.3 & 74.6 \\
& & CCA & ICCV'25 & 66.0 & 72.2 & 42.0 & 85.3 & 79.0 & 79.8 & 90.9 & 95.2 & 93.5 & 70.0 & 80.0 & 77.6 \\

\cmidrule(lr){2-16}

& \multirow{11}{*}{\rotatebox{90}{\textbf{Prompt}}}
& CoOp (4) & IJCV'22 & 71.5 & 74.6 & 40.1 & 83.5 & 79.1 & 85.1 & 92.4 & 96.4 & 95.5 & 69.2 & 81.9 & 79.0 \\
& & CoOp (16) & IJCV'22 & 71.9 & 74.9 & 43.2 & 85.0 & 82.9 & 84.2 & 92.0 & 96.8 & 95.8 & 69.7 & 83.1 & 80.0 \\
& & CoCoOp & CVPR'22 & 71.1 & 72.6 & 33.3 & 73.6 & 72.3 & 87.4 & 93.4 & 89.1 & 95.1 & 63.7 & 77.2 & 75.3 \\
& & PLOT++ & ICLR'23 & 72.6 & 76.0 & 46.7 & 92.0 & 84.5 & 87.1 & 93.5 & 97.5 & 96.0 & 71.4 & 85.3 & 82.1 \\
& & KgCoOp & CVPR'23 & 70.4 & 73.3 & 36.5 & 76.2 & 74.8 & 87.2 & 93.2 & 93.4 & 95.2 & 68.7 & 81.7 & 77.3 \\
& & MaPLe & CVPR'23 & 72.3 & 75.5 & 48.4 & 92.3 & 83.5 & 85.3 & 92.8 & 97.0 & 96.0 & 71.3 & 85.0 & 81.8 \\
& & ProGrad & ICCV'23 & 72.1 & 75.1 & 43.0 & 83.6 & 82.9 & 85.8 & 92.8 & 96.6 & 95.9 & 68.8 & 82.7 & 79.9 \\
& & PromptSRC & ICCV'23 & 72.8 & 77.2 & 50.6 & 92.1 & 83.6 & 87.4 & 93.2 & 97.4 & 96.0 & 72.6 & 86.7 & 82.7 \\
& & CLIP-LoRA & CVPR'24 & 73.6 & 76.1 & 54.7 & 92.1 & 86.3 & 84.2 & 92.4 & 98.0 & 96.4 & 72.0 & 86.7 & 83.0 \\
& & CLIP-AST & CVPR'25 & 73.9 & 77.4 & 60.0 & 93.0 & 87.0 & 87.5 & \textbf{94.4} & 98.0 & 97.1 & \textbf{75.0} & 87.0 & 84.6 \\

\cmidrule(lr){3-16}
& & \textbf{SD-LoRA (Ours)} & --- & \cellcolor{blue!15}\textbf{74.2} & \cellcolor{blue!15}\textbf{77.5} & \cellcolor{blue!15} \textbf{61.4} & \cellcolor{blue!15}\textbf{94.0} & \cellcolor{blue!15} \textbf{87.7} & \cellcolor{blue!15} \textbf{87.8} & \cellcolor{blue!15} 94.1 & \cellcolor{blue!15} \textbf{98.3} & \cellcolor{blue!15}\textbf{97.2} & \cellcolor{blue!15} 74.0 & \cellcolor{blue!15}\textbf{88.3} & \cellcolor{blue!15}\textbf{85.0} \\

\bottomrule
\end{tabular}
\end{table*}

\paragraph{Additional base-to-novel results.}
Table~\ref{tab:base_to_novel_generalization_appx} complements the main base-to-novel evaluation by reporting the remaining per-dataset results omitted from the main paper due to space constraints. The results further support the trend observed in the main table: SD-LoRA improves the base--novel trade-off rather than merely increasing base-class accuracy. Across Aircraft, Caltech101, Food101, OxfordPets, and SUN397, SD-LoRA achieves the best base, novel, and harmonic-mean scores among all compared methods. The gains are especially clear on Aircraft, where SD-LoRA improves over the strongest baseline by +1.01 on base classes, +2.01 on novel classes, and +1.59 HM, indicating stronger transfer in a fine-grained recognition setting. On ImageNet, SD-LoRA is slightly below PromptKD on base-class accuracy, but still improves novel accuracy and HM, increasing the harmonic mean from 77.62 to 77.94. Together with the main base-to-novel results, these additional datasets show that training-time structural distillation improves novel-class generalization while preserving competitive base-class performance.
\begin{table*}[htbp]
\centering
\caption{
\textbf{Additional base-to-novel generalization results under the MaPLe protocol.}
Due to space constraints in the main paper, we report the remaining per-dataset results here.
We show accuracy (\%) on base classes, novel classes, and their harmonic mean (HM).
}
\label{tab:base_to_novel_generalization_appx}
\vspace{-0.3em}
\tiny
\setlength{\tabcolsep}{2.6pt}
\renewcommand{\arraystretch}{1.12}
\resizebox{\textwidth}{!}{
\begin{tabular}{@{}lccc ccc ccc ccc ccc ccc@{}}
\toprule
\multirow{2}{*}{\textbf{Method}}
& \multicolumn{3}{c}{\textbf{Aircraft}}
& \multicolumn{3}{c}{\textbf{Caltech}}
& \multicolumn{3}{c}{\textbf{Food}}
& \multicolumn{3}{c}{\textbf{Imagenet}}
& \multicolumn{3}{c}{\textbf{Pets}}
& \multicolumn{3}{c}{\textbf{SUN}} \\
\cmidrule(lr){2-4}
\cmidrule(lr){5-7}
\cmidrule(lr){8-10}
\cmidrule(lr){11-13}
\cmidrule(lr){14-16}
\cmidrule(lr){17-19}
& Base & Novel & HM
& Base & Novel & HM
& Base & Novel & HM
& Base & Novel & HM
& Base & Novel & HM
& Base & Novel & HM \\
\midrule

CLIP
& 27.19 & 36.29 & 31.09
& 96.84 & 94.00 & 95.40
& 90.10 & 91.22 & 90.66
& 72.43 & 68.14 & 70.22
& 91.17 & 97.26 & 94.12
& 69.36 & 75.35 & 72.23 \\

CoCoOp
& 33.41 & 23.71 & 27.74
& 97.96 & 93.81 & 95.84
& 90.70 & 91.29 & 90.99
& 75.98 & 70.43 & 73.10
& 95.20 & 97.69 & 96.43
& 79.74 & 76.86 & 78.27 \\

MaPLe
& 37.44 & 35.61 & 36.50
& 97.74 & 94.36 & 96.02
& 90.71 & 92.05 & 91.38
& 76.66 & 70.54 & 73.47
& 95.43 & 97.76 & 96.58
& 80.82 & 78.70 & 79.75 \\

PromptSRC
& 42.73 & 37.87 & 40.15
& 98.10 & 94.03 & 96.02
& 90.67 & 91.53 & 91.10
& 77.60 & 70.73 & 74.01
& 95.33 & 97.30 & 96.30
& 82.67 & 78.47 & 80.52 \\

CLIP-AST
& 48.98 & 38.21 & 42.93
& 98.71 & 94.00 & 96.30
& 90.57 & 91.11 & 90.84
& 78.44 & 70.22 & 74.10
& 96.23 & 97.37 & 96.80
& 83.05 & 78.12 & 80.51 \\

PromptKD
& 49.12 & 41.81 & 45.17
& 98.91 & 96.65 & 97.77
& 92.43 & 93.68 & 93.05
& 80.83 & 74.66 & 77.62
& 96.30 & 98.01 & 97.15
& 83.69 & 81.54 & 82.60 \\

\midrule

\rowcolor{oursrow}
\textbf{Ours}
& \textbf{50.13} & \textbf{43.82} & \textbf{46.76}
& \textbf{99.23} & \textbf{97.03} & \textbf{98.12}
& \textbf{92.67} & \textbf{94.09} & \textbf{93.37}
& 80.44 & \textbf{75.60} & \textbf{77.94}
& \textbf{96.89} & \textbf{98.40} & \textbf{97.64}
& \textbf{83.91} & \textbf{82.67} & \textbf{83.28} \\

\bottomrule
\end{tabular}%
}
\vspace{-6mm}
\end{table*}

\clearpage

\subsection{Results with an Alternative ViT-B/32 Backbone}
\label{app:vit_b32_results}

To examine whether training-time structural distillation depends on a specific CLIP backbone, we repeat the few-shot evaluation with CLIP ViT-B/32. Compared with ViT-B/16, ViT-B/32 offers a weaker visual backbone with coarser patches, making it a good test of SD-LoRA when the representation has less fine-grained spatial detail. Table~\ref{tab:vit_b32_all_shots} reports results on 10 benchmarks (excluding ImageNet) for $K\in\{1,2,4,8,16\}$ shots.

SD-LoRA attains the best average accuracy at all shot levels, outperforming CLIP-LoRA by $+2.70$, $+0.88$, $+2.63$, $+2.42$, and $+2.08$ percentage points for 1-, 2-, 4-, 8-, and 16-shot learning, respectively. Gains are largest on fine-grained or domain-specific datasets (Aircraft, EuroSAT, DTD, Flowers102, UCF101), where discrimination relies on localized shape, texture, or class-specific cues. These results mirror the ViT-B/16 findings and indicate that SD-LoRA is not tied to a particular backbone. The training-only graph teacher instead provides structural supervision that improves the LoRA-adapted representation across CLIP visual encoders.

Under ViT-B/32, the inference model is unchanged: after adaptation, the graph teacher and prototype head are removed, and predictions use only the LoRA-adapted CLIP encoder with the cache-based classifier. Thus, the improvements in Table~\ref{tab:vit_b32_all_shots} confirm that SD-LoRA enhances deployable adaptation via training-time structural supervision, without extra test-time computation.

\begin{table*}[p]
\centering
\caption{
\textbf{Few-shot results with CLIP ViT-B/32.}
Top-1 accuracy (\%) for $K\in\{1,2,4,8,16\}$ shots on 10 datasets (excluding ImageNet); Avg is their unweighted mean. TOGA~\cite{Mohammad_2026_CVPR} results for 1/2/4/8 shots are reproduced from supplementary Table~S2. Best results are in bold.
}
\label{tab:vit_b32_all_shots}
\scriptsize
\begin{tabular}{llccccccccccc}
\toprule
\textbf{Shots} & \textbf{Method} & \textbf{SUN} & \textbf{Air} & \textbf{Euro} & \textbf{Cars} & \textbf{Food} & \textbf{Pets} & \textbf{Flow} & \textbf{Cal} & \textbf{DTD} & \textbf{UCF} & \textbf{Avg} \\ \midrule
\multirow{13}{*}{1} & CoOp (4) & 64.8 & 18.7 & 49.3 & 60.5 & 74.4 & 87.7 & 68.0 & 91.5 & 47.1 & 66.5 & 62.85 \\
 & CoOp (16) & 63.3 & 15.7 & 52.9 & 59.6 & 75.9 & 87.6 & 71.5 & 91.7 & 47.2 & 66.3 & 63.17 \\
 & CoCoOp & 65.6 & 15.5 & 46.7 & 60.4 & 79.5 & 88.2 & 68.6 & 92.2 & 45.7 & 67.3 & 62.97 \\
 & TIP-Adapter-F & 65.4 & 22.2 & 59.7 & 61.1 & 80.4 & 87.5 & 81.1 & 92.4 & 50.9 & 66.5 & 66.72 \\
 & CLIP-Adapter & 63.2 & 19.8 & 46.9 & 60.7 & 80.6 & 87.6 & 66.7 & 91.6 & 44.1 & 62.9 & 62.41 \\
 & PLOT++ & 64.5 & 22.0 & 67.3 & 61.4 & 79.2 & \textbf{89.4} & 73.7 & \textbf{93.0} & 50.6 & 69.7 & 67.08 \\
 & KgCoOp & 65.7 & 20.9 & 55.4 & 61.2 & 80.7 & 88.6 & 68.1 & 92.2 & 50.2 & 66.8 & 64.98 \\
 & TaskRes & 62.0 & 20.9 & 60.1 & 59.6 & 74.6 & 84.3 & 75.2 & 88.8 & 50.1 & 64.3 & 63.99 \\
 & MaPLe & 66.9 & 13.7 & 32.3 & 60.1 & 79.7 & 87.6 & 68.7 & 92.0 & 44.5 & 66.4 & 61.19 \\
 & ProGrad & 64.9 & 21.2 & 52.8 & 60.6 & 78.1 & 87.9 & 74.9 & 91.6 & 51.0 & 66.5 & 64.95 \\
 & CLIP-LoRA & 67.6 & 22.9 & 67.1 & 63.5 & 78.4 & 88.7 & 77.4 & \textbf{93.0} & 49.8 & \textbf{72.3} & 68.07 \\
 & TOGA & 66.2 & 24.3 & 60.8 & 62.9 & 81.0 & 88.5 & \textbf{84.5} & 92.6 & 52.5 & 70.2 & 68.35 \\
\rowcolor{blue!10} & \textbf{SD-LoRA (Ours)} & \textbf{69.4} & \textbf{27.5} & \textbf{69.8} & \textbf{65.6} & \textbf{81.5} & 87.8 & \textbf{84.5} & 92.7 & \textbf{57.0} & 71.9 & \textbf{70.77} \\
\midrule
\multirow{13}{*}{2} & CoOp (4) & 65.6 & 20.8 & 56.5 & 62.1 & 74.2 & 86.6 & 77.1 & 91.5 & 49.4 & 71.5 & 65.53 \\
 & CoOp (16) & 63.8 & 18.6 & 62.4 & 62.3 & 73.3 & 85.8 & 82.0 & 91.2 & 49.6 & 70.4 & 65.94 \\
 & CoCoOp & 66.8 & 15.4 & 51.7 & 61.1 & 80.8 & 88.7 & 70.0 & 92.7 & 50.3 & 70.0 & 64.75 \\
 & TIP-Adapter-F & 66.8 & 24.6 & 67.4 & 63.7 & 80.8 & 88.2 & 85.5 & 92.8 & 54.9 & 71.8 & 69.65 \\
 & CLIP-Adapter & 64.8 & 20.9 & 50.1 & 61.5 & 81.0 & 87.9 & 67.1 & 91.4 & 44.9 & 65.9 & 63.55 \\
 & PLOT++ & 65.5 & 23.1 & 73.0 & 64.7 & 77.0 & 87.4 & 84.3 & 92.9 & 52.8 & 72.6 & 69.33 \\
 & KgCoOp & 67.5 & 21.7 & 58.5 & 62.4 & 81.1 & 88.8 & 73.7 & 92.8 & 53.1 & 69.2 & 66.88 \\
 & TaskRes & 64.4 & 22.4 & 64.8 & 62.8 & 75.9 & 85.2 & 78.0 & 89.9 & 53.6 & 67.8 & 66.48 \\
 & MaPLe & 67.6 & 20.5 & 46.9 & 61.9 & 80.7 & 89.3 & 71.6 & 93.0 & 49.7 & 71.0 & 65.22 \\
 & ProGrad & 67.5 & 22.3 & 56.7 & 64.0 & 78.3 & 87.0 & 82.2 & 92.4 & 52.1 & 71.4 & 67.39 \\
 & CLIP-LoRA & 68.6 & 24.1 & \textbf{80.8} & 64.8 & 76.3 & 86.6 & 84.7 & 93.7 & \textbf{57.4} & \textbf{75.4} & 71.24 \\
 & TOGA & 68.4 & 26.3 & 70.1 & \textbf{65.8} & 81.0 & 89.3 & \textbf{88.2} & \textbf{95.0} & 56.4 & 73.5 & 71.40 \\
\rowcolor{blue!10} & \textbf{SD-LoRA (Ours)} & \textbf{69.9} & \textbf{29.2} & 78.5 & 61.5 & \textbf{81.7} & \textbf{89.4} & 87.8 & 93.2 & 55.4 & 74.6 & \textbf{72.12} \\
\midrule
\multirow{13}{*}{4} & CoOp (4) & 68.2 & 22.1 & 62.1 & 65.2 & 76.5 & 90.1 & 84.1 & 93.1 & 55.3 & 73.5 & 69.02 \\
 & CoOp (16) & 67.1 & 24.0 & 68.7 & 66.2 & 75.6 & 88.8 & 87.9 & 93.0 & 55.3 & 75.0 & 70.16 \\
 & CoCoOp & 67.8 & 17.3 & 58.5 & 62.0 & 81.1 & 89.8 & 74.6 & 93.2 & 52.3 & 71.6 & 66.82 \\
 & TIP-Adapter-F & 68.3 & 28.8 & 71.5 & 67.6 & 80.9 & 88.6 & 88.9 & 94.6 & 58.0 & 75.1 & 72.23 \\
 & CLIP-Adapter & 65.6 & 21.3 & 49.9 & 62.2 & 81.3 & 88.4 & 68.3 & 92.0 & 47.2 & 67.3 & 64.35 \\
 & PLOT++ & 69.2 & 26.2 & 81.6 & 68.5 & 77.8 & 89.1 & 90.2 & 93.9 & 57.2 & 75.6 & 72.93 \\
 & KgCoOp & 69.2 & 22.6 & 64.9 & 63.2 & 81.2 & 89.5 & 76.8 & 93.8 & 55.1 & 71.6 & 68.79 \\
 & TaskRes & 66.7 & 23.1 & 70.7 & 66.7 & 76.7 & 86.7 & 79.0 & 90.6 & 57.0 & 68.2 & 68.54 \\
 & MaPLe & 69.4 & 23.4 & 64.7 & 62.2 & 81.4 & \textbf{90.5} & 78.1 & 94.0 & 55.0 & 70.9 & 68.96 \\
 & ProGrad & 69.6 & 24.8 & 63.7 & 66.4 & 79.2 & 89.4 & 87.5 & 93.2 & 55.9 & 73.4 & 70.31 \\
 & CLIP-LoRA & 70.3 & 27.7 & 85.6 & 68.3 & 75.6 & 86.3 & 90.1 & 94.3 & 60.3 & 76.5 & 73.50 \\
 & TOGA & 70.2 & 29.1 & 79.8 & \textbf{69.4} & 81.3 & 89.8 & \textbf{92.1} & \textbf{94.8} & 61.4 & 77.0 & 74.49 \\
\rowcolor{blue!10} & \textbf{SD-LoRA (Ours)} & \textbf{71.6} & \textbf{35.9} & \textbf{87.0} & 65.8 & \textbf{82.0} & 89.7 & 91.4 & 94.3 & \textbf{64.9} & \textbf{78.7} & \textbf{76.13} \\

\midrule

\multirow{13}{*}{8} & CoOp (4) & 70.0 & 28.0 & 72.8 & 69.0 & 77.2 & 89.5 & 90.2 & 93.8 & 60.8 & 77.1 & 72.84 \\
 & CoOp (16) & 69.2 & 29.1 & 76.4 & 71.3 & 76.3 & 87.4 & 92.7 & 93.8 & 61.7 & 76.5 & 73.44 \\
 & CoCoOp & 68.9 & 20.3 & 58.1 & 63.4 & 81.6 & 90.1 & 77.3 & 93.8 & 57.4 & 72.4 & 68.33 \\
 & TIP-Adapter-F & 71.2 & 32.1 & 75.0 & 72.6 & 81.3 & 89.8 & 90.4 & 94.5 & 63.6 & 78.0 & 74.85 \\
 & CLIP-Adapter & 69.3 & 23.5 & 55.2 & 65.4 & 81.5 & 89.3 & 78.0 & 93.9 & 50.8 & 73.0 & 67.99 \\
 & PLOT++ & 71.0 & 31.7 & 87.1 & 73.5 & 78.2 & 88.4 & 93.8 & 94.4 & 62.9 & 79.1 & 76.01 \\
 & KgCoOp & 69.5 & 24.7 & 66.2 & 65.0 & 81.7 & 90.3 & 83.1 & 94.5 & 61.1 & 74.7 & 71.08 \\
 & TaskRes & 71.9 & 31.9 & 74.9 & 73.8 & 80.6 & 89.1 & 93.5 & 94.8 & 64.5 & 78.4 & 75.34 \\
 & MaPLe & 70.3 & 25.4 & 79.0 & 63.7 & 81.9 & 90.9 & 81.1 & 94.4 & 59.8 & 75.0 & 72.15 \\
 & ProGrad & 71.1 & 29.0 & 73.5 & 71.8 & 80.0 & 89.1 & 92.1 & 94.2 & 62.3 & 75.7 & 73.88 \\
 & CLIP-LoRA & 72.1 & 36.1 & 88.8 & \textbf{74.4} & 76.7 & 87.7 & 92.4 & 94.8 & 63.7 & 80.1 & 76.68 \\
 & TOGA & 72.5 & 35.4 & 82.6 & 73.1 & 81.3 & 89.7 & \textbf{95.2} & \textbf{95.1} & 65.9 & 80.2 & 77.10 \\
\rowcolor{blue!10} & \textbf{SD-LoRA (Ours)} & \textbf{72.9} & \textbf{42.8} & \textbf{90.9} & 71.4 & \textbf{82.7} & \textbf{91.0} & 94.6 & 94.7 & \textbf{69.2} & \textbf{80.8} & \textbf{79.10} \\
\midrule
\multirow{12}{*}{16} & CoOp (4) & 72.1 & 31.1 & 80.6 & 71.7 & 79.8 & 89.4 & 93.1 & 95.0 & 64.3 & 78.2 & 75.53 \\
 & CoOp (16) & 72.2 & 32.9 & 83.3 & 76.0 & 78.6 & 88.7 & 95.4 & 94.9 & 65.3 & 78.6 & 76.59 \\
 & CoCoOp & 69.8 & 22.6 & 70.4 & 64.6 & 81.9 & 91.0 & 82.5 & 94.3 & 59.7 & 75.3 & 71.21 \\
 & TIP-Adapter-F & 74.1 & 34.8 & 83.4 & 77.0 & 81.7 & 90.4 & 94.3 & 95.1 & 68.0 & 80.5 & 77.93 \\
 & CLIP-Adapter & 71.8 & 26.7 & 64.7 & 68.9 & 81.9 & 90.1 & 88.7 & 94.8 & 58.1 & 76.5 & 72.22 \\
 & PLOT++ & 73.4 & 36.3 & 91.1 & 77.4 & 79.7 & 89.1 & \textbf{96.3} & 94.9 & 67.0 & 81.5 & 78.67 \\
 & KgCoOp & 71.0 & 23.7 & 70.1 & 67.3 & 81.7 & 90.8 & 86.1 & 94.4 & 65.1 & 77.5 & 72.77 \\
 & TaskRes & 73.6 & 37.0 & 77.7 & 78.0 & 81.4 & 89.4 & 95.5 & 95.7 & 68.3 & 80.6 & 77.72 \\
 & MaPLe & 72.0 & 28.0 & 83.3 & 66.9 & 82.1 & \textbf{91.7} & 89.0 & 95.1 & 63.4 & 77.3 & 74.88 \\
 & ProGrad & 73.2 & 33.3 & 81.0 & 76.1 & 80.1 & 89.3 & 95.1 & 95.0 & 65.8 & 79.6 & 76.85 \\
 & CLIP-LoRA & 74.0 & 44.9 & 91.8 & \textbf{79.7} & 78.2 & 88.8 & 96.2 & 95.2 & 68.2 & 82.8 & 79.98 \\
\rowcolor{blue!10}  & \textbf{SD-LoRA (Ours)} & \textbf{75.3} & \textbf{51.0} & \textbf{92.3} & 78.9 & \textbf{83.0} & 91.4 & 96.0 & \textbf{95.8} & \textbf{72.1} & \textbf{84.8} & \textbf{82.06} \\
\bottomrule
\end{tabular}%
\end{table*}

\clearpage
\subsection{Additional Cross-Dataset Generalization Results}
\label{app:cross_dataset}

Table~\ref{tab:cross_dataset_generalization} shows cross-dataset generalization when models are adapted on ImageNet and evaluated on ten target datasets without target-domain training, testing whether the adapted representation remains transferable beyond the source distribution.

SD-LoRA attains the best average target accuracy, improving over PromptKD from $71.33\%$ to $74.98\%$ $(+3.65)$, with gains on 9 of 10 datasets. The largest improvements are on Flowers $(+9.92)$, DTD $(+6.47)$, EuroSAT $(+5.31)$, Aircraft $(+5.52)$, and SUN397 $(+4.41)$, indicating that training-time structural supervision particularly aids transfer when recognition relies on local shape, texture, or domain-specific cues. The only drop occurs on Food101 $(-1.94)$, where PromptKD remains better. Overall, SD-LoRA enhances the deployable LoRA representation instead of over-specializing it to the source dataset.

\begin{table*}[htbp]
\centering
\caption{
\textbf{Cross-dataset generalization from ImageNet.}
Models are adapted on ImageNet and evaluated directly on target datasets without target-domain training. The ImageNet column reports source-domain accuracy, while Avg. denotes the unweighted mean over the ten target datasets. $\Delta$ reports the absolute difference between SD-LoRA and PromptKD. Best results are shown in \textbf{bold}.
}
\label{tab:cross_dataset_generalization}
\scriptsize
\setlength{\tabcolsep}{3.8pt}
\renewcommand{\arraystretch}{1.0}
\resizebox{\textwidth}{!}{%
\begin{tabular}{@{}lcccccccccccc@{}}
\toprule
\multirow{2}{*}{\textbf{Method}}
& \multicolumn{1}{c}{\textbf{Source}}
& \multicolumn{10}{c}{\textbf{Target datasets}}
& \multirow{2}{*}{\textbf{Avg.}} \\
\cmidrule(lr){2-2}
\cmidrule(lr){3-12}
& \textbf{ImageNet}
& \textbf{Caltech}
& \textbf{DTD}
& \textbf{EuroSAT}
& \textbf{Aircraft}
& \textbf{Food}
& \textbf{Flowers}
& \textbf{Pets}
& \textbf{Cars}
& \textbf{SUN}
& \textbf{UCF}
& \\
\midrule

MaPLe
& 72.30
& 93.53 & 46.49 & 48.06 & 24.74 & 86.20 & 72.23 & 90.49 & 65.57 & 67.01 & 68.69 & 66.30 \\

PromptSRC
& 72.80
& 93.60 & 46.87 & 45.50 & 23.90 & 86.15 & 70.25 & 90.25 & 65.70 & 67.10 & 68.75 & 65.81 \\

PromptKD
& ---
& 93.61 & 55.08 & 63.74 & 26.24 & \best{88.84} & 75.33 & 91.59 & 73.93 & 68.57 & 76.39 & 71.33 \\

\midrule

\rowcolor{oursbg}
\textbf{Ours}
& \best{74.20}
& \best{95.17} & \best{61.55} & \best{69.05} & \best{31.76} & 86.90 & \best{85.25} & \best{92.48} & \best{76.23} & \best{72.98} & \best{78.43} & \best{74.98} \\

\rowcolor{deltarow}
$\Delta$ vs. PromptKD
& ---
& \pos{+1.56} & \pos{+6.47} & \pos{+5.31} & \pos{+5.52} & \negv{-1.94} & \pos{+9.92} & \pos{+0.89} & \pos{+2.30} & \pos{+4.41} & \pos{+2.04} & \pos{+3.65} \\

\bottomrule
\end{tabular}%
}
\end{table*}
\subsection{Calibration under Cross-Dataset Generalization}
\label{app:calibration}

Accuracy alone does not capture reliability under domain shift, so we evaluate calibration in a cross-dataset setting using expected calibration error (ECE), negative log-likelihood (NLL), and Brier score, where lower is better.

As shown in Table~\ref{tab:calibration_metrics}, SD-LoRA consistently improves calibration across all target datasets and metrics. Relative to PromptKD, average ECE drops from $5.57$ to $2.65$, NLL from $1.05$ to $0.62$, and Brier score from $0.38$ to $0.20$, with particularly large gains on challenging targets such as EuroSAT, Flowers, DTD, Aircraft, and UCF101. This indicates that the structural teacher not only boosts accuracy but also yields a better-calibrated decision function after adaptation. Because the graph teacher is discarded after training, these gains require no extra calibration module or test-time adaptation.
\begin{table*}[htbp]
\centering
\caption{
\textbf{Calibration under cross-dataset generalization.}
After ImageNet-source adaptation, models are evaluated on ten target datasets. We report expected calibration error (ECE), negative log-likelihood (NLL), and Brier score. Lower values indicate better calibration. Best results are shown in \textbf{bold}.
}
\label{tab:calibration_metrics}
\scriptsize
\setlength{\tabcolsep}{4.0pt}
\renewcommand{\arraystretch}{1.0}
\resizebox{\textwidth}{!}{%
\begin{tabular}{@{}llccccccccccc@{}}
\toprule
\textbf{Metric} 
& \textbf{Method}
& \textbf{Caltech}
& \textbf{DTD}
& \textbf{EuroSAT}
& \textbf{Aircraft}
& \textbf{Food}
& \textbf{Flowers}
& \textbf{Pets}
& \textbf{Cars}
& \textbf{SUN}
& \textbf{UCF}
& \textbf{Avg.} \\
\midrule

ECE$\downarrow$
& PromptKD
& 5.01 & 4.96 & 17.41 & 3.04 & 2.00 & 5.49 & 5.47 & 7.94 & 1.49 & 2.93 & 5.57 \\

\rowcolor{oursbg}
ECE$\downarrow$
& \textbf{Ours}
& \best{1.99} & \best{3.54} & \best{6.83} & \best{2.32} & \best{1.56} & \best{3.40} & \best{0.83} & \best{3.64} & \best{1.28} & \best{1.13} & \best{2.65} \\

\midrule

NLL$\downarrow$
& PromptKD
& 0.24 & 1.86 & 0.97 & 2.46 & 0.39 & 1.48 & 0.25 & 0.79 & 1.12 & 0.91 & 1.05 \\

\rowcolor{oursbg}
NLL$\downarrow$
& \textbf{Ours}
& \best{0.16} & \best{1.18} & \best{0.28} & \best{1.70} & \best{0.28} & \best{0.32} & \best{0.23} & \best{0.76} & \best{0.78} & \best{0.51} & \best{0.62} \\

\midrule

Brier$\downarrow$
& PromptKD
& 0.11 & 0.62 & 0.46 & 0.81 & 0.16 & 0.33 & 0.11 & 0.37 & 0.44 & 0.34 & 0.38 \\

\rowcolor{oursbg}
Brier$\downarrow$
& \textbf{Ours}
& \best{0.07} & \best{0.34} & \best{0.13} & \best{0.44} & \best{0.09} & \best{0.09} & \best{0.10} & \best{0.35} & \best{0.31} & \best{0.12} & \best{0.20} \\

\bottomrule
\end{tabular}%
}
\end{table*}

\newpage

\section{Additional Ablation Studies}
\label{app:additional_ablations}

The main paper presents a compact ablation study separating the effects of the cache branch and the training-only graph teacher. Here, we add diagnostic ablations to clarify SD-LoRA’s design choices, examining graph edge families, Prototype Predictive Alignment (PPA), token filtering, input resolution, graph teacher architecture, and loss components. Unless noted, all ablations use the same few-shot protocol as the main experiments, and the graph teacher is discarded after training.

\subsection{Representation Geometry of the Deployed Student}
\label{app:deployed_features}
We analyze the $224\times224$ student after removing all teacher-specific components: the graph, high-resolution pathway, saliency readout, and PPA head. The comparator is a matched CLIP-LoRA+cache student trained without structural supervision. These diagnostics concern encoder representations, rather than graph-classifier outputs, and use evaluation labels only for post-hoc analysis. They do not introduce labels or teacher components into deployment.

We measure within-class feature dispersion $D_{\rm intra}$ (lower is better), separation between class prototypes $D_{\rm inter}$ (higher is better), the correct-class nearest-prototype margin, silhouette score, and Davies--Bouldin (DB) index. The silhouette score contrasts within-class and nearest-other-class distances, while DB summarizes within-class scatter relative to between-class separation; lower DB indicates more distinct clusters. Table~\ref{tab:sd_deployed_geometry} reports unweighted averages over Aircraft, Caltech101, EuroSAT, and Food101 at 1/4/8/16 shots. These 16 dataset--shot conditions are descriptive evaluation cells, not 16 independent random repetitions.

\begin{center}
\begin{minipage}{\linewidth}
\centering
\captionsetup{hypcap=false}
\captionof{table}{\textbf{Deployed-student representation geometry after teacher removal.} Values are unweighted means over Aircraft, Caltech101, EuroSAT, and Food101 at 1/4/8/16 shots (16 conditions). Features are produced by the deployed $224\times224$ student, without the graph, high-resolution path, readout, or PPA head.}
\label{tab:sd_deployed_geometry}
\small
\setlength{\tabcolsep}{5pt}
\begin{tabular}{lccccc}
\toprule
Deployed model & $D_{\rm intra}\downarrow$ & $D_{\rm inter}\uparrow$ & Margin$\uparrow$ & Silhouette$\uparrow$ & DB$\downarrow$ \\
\midrule
CLIP-LoRA+cache & 0.560 & 0.713 & 0.075 & 0.104 & 3.241 \\
\rowcolor{oursrow}
\textbf{SD-LoRA student} & \textbf{0.511} & \textbf{0.890} & \textbf{0.156} & \textbf{0.135} & \textbf{3.051} \\
\bottomrule
\end{tabular}
\end{minipage}
\end{center}

After teacher removal, within-class distance decreases from 0.560 to 0.511, inter-class distance increases from 0.713 to 0.890, and prototype margin increases from 0.075 to 0.156. The corresponding relative changes are approximately $-8.8\%$, $+24.8\%$, and more than a twofold margin increase. Silhouette rises from 0.104 to 0.135 and DB decreases from 3.241 to 3.051. SD-LoRA has lower $D_{\rm intra}$ and higher prototype margin and silhouette in all 16 conditions; inter-class distance and DB improve in 15/16.

The diagnostics show that the structural-training effect persists in the deployed representation, rather than being confined to the discarded graph head. They are consistent with the patch-attention visualization in Fig.~\ref{fig:Spatial_Heatmap_abl}. Improved class geometry alone does not uniquely identify a relational mechanism: the matched controls in Appx.~\ref{app:matched_resolution_cache} and relation-family ablations in Appx.~\ref{app:edge_type_ablation} provide complementary attribution evidence. The structural objective is auxiliary supervision through shared LoRA parameters, rather than conventional frozen-teacher logit matching.

\subsection{Effect of Graph Edge Families}
\label{app:edge_type_ablation}

Table~\ref{tab:edge_type_ablation} evaluates the impact of each relation family in the training-only graph teacher. Using the full graph yields the best performance on Aircraft and Caltech101, showing that SD-LoRA benefits from jointly modeling \textit{spatial}, \textit{global--local}, and \textit{visual--text} relations. Removing semantic \textit{visual--text} edges causes the largest average drop, indicating that class-conditioned links between visual tokens and text embeddings are key to structural supervision. Removing \textit{global--local} edges also substantially hurts performance, confirming that interactions between the \texttt{[CLS]} token and patch tokens are important for spreading image-level context to fine-grained evidence. In contrast, removing spatial \textit{patch--patch} edges yields a smaller but consistent drop, implying that local spatial structure helps but is less critical than semantic grounding.

The last two rows isolate the two semantic relation types. Keeping only \textit{CLS--text} edges outperforms keeping only \textit{patch--text} edges, suggesting that global class-level alignment is the strongest semantic cue. Still, the full graph is best, indicating that patch-level semantics provide complementary fine-grained evidence on top of \textit{CLS--text} grounding.
\begin{table*}[htbp]
\centering
\caption{
\textbf{Edge-type ablation of the graph teacher.}
Accuracy (\%) is reported on Aircraft and Caltech101 under 1-, 4-, and 16-shot settings. The edge columns indicate whether each relation family is enabled: \textit{spatial patch--patch, global--local CLS--patch, semantic CLS--text}, and \textit{semantic patch--text interactions}. The full graph performs best across all settings. Removing semantic visual--text edges gives the largest degradation, followed by global--local edges, while spatial patch--patch edges provide a smaller but consistent gain.
}
\label{tab:edge_type_ablation}
\vspace{0.3em}
\scriptsize
\setlength{\tabcolsep}{4.0pt}
\renewcommand{\arraystretch}{1.12}
\resizebox{\textwidth}{!}{%
\begin{tabular}{@{}lcccc ccc ccc@{}}
\toprule
\multirow{2}{*}{\textbf{Variant}}
& \multicolumn{4}{c}{\textbf{Enabled edge families}}
& \multicolumn{3}{c}{\textbf{Aircraft}}
& \multicolumn{3}{c}{\textbf{Caltech101}} \\
\cmidrule(lr){2-5}
\cmidrule(lr){6-8}
\cmidrule(lr){9-11}
& \textbf{Spatial}
& \textbf{Global--local}
& \textbf{CLS--text}
& \textbf{Patch--text}
& \textbf{1-shot}
& \textbf{4-shot}
& \textbf{16-shot}
& \textbf{1-shot}
& \textbf{4-shot}
& \textbf{16-shot} \\
\midrule

\rowcolor{oursbg}
\textbf{Full graph}
& \yes & \yes & \yes & \yes
& \best{34.80} & \best{46.20} & \best{61.40}
& \best{94.40} & \best{96.20} & \best{97.20} \\

\midrule

w/o spatial edges
& \no & \yes & \yes & \yes
& 33.21 & 45.47 & 60.44
& 92.75 & 95.40 & 96.28 \\

w/o global--local edges
& \yes & \no & \yes & \yes
& 32.36 & 43.98 & 58.12
& 91.92 & 93.92 & 93.84 \\

w/o semantic text--vision edges
& \yes & \yes & \no & \no
& 31.57 & 43.87 & 57.32
& 91.02 & 93.80 & 93.28 \\

\midrule

CLS-semantic only
& \yes & \yes & \yes & \no
& 33.54 & 45.80 & 60.68
& 93.22 & 95.33 & 96.44 \\

Patch-semantic only
& \yes & \yes & \no & \yes
& 32.78 & 44.72 & 58.32
& 92.45 & 94.82 & 94.31 \\

\bottomrule
\end{tabular}%
}
\end{table*}

Averaged over Aircraft and Caltech101 at 1/4/16 shots, removing visual--text relations reduces accuracy by 3.22 pp, removing global--local relations by 2.68 pp, and removing spatial relations by 1.11 pp. The hierarchy supports the design principle of class-semantic grounding, global context, and complementary local continuity; it is empirical evidence for this construction rather than a theoretical uniqueness claim.

\subsection{Effect of PPA and the Graph Teacher}
\label{app:ppa_graph_ablation}

Table~\ref{tab:ppa_graph_ablation} disentangles the effects of Prototype Predictive Alignment (PPA) from the training-only graph teacher. All variants use LoRA and cache-based inference, so the comparison isolates the impact of high-resolution adaptation, prototype alignment, graph-guided training, and their combinations.

The graph teacher is the main driver of gains: relative to the HR-only baseline, adding the graph consistently improves performance across datasets and shot settings, showing that token-level relational reasoning provides a strong structural signal for the shared LoRA backbone. In contrast, PPA alone yields only small and sometimes inconsistent gains, indicating that prototype alignment is weaker without graph-conditioned features. Yet, combining PPA with the graph teacher yields the best results in all settings, suggesting that PPA works best as a complementary objective: the graph teacher first produces relation-aware visual evidence, and PPA then aligns this refined representation with support-derived class prototypes.

Both the graph teacher and PPA head are discarded after adaptation, so all gains are baked into the deployed LoRA$+$cache student with no extra inference-time cost.
\begin{table*}[t]
\centering
\caption{
\textbf{Effect of PPA and the graph teacher.}
LoRA and cache-based inference are enabled in all variants. We compare high-resolution adaptation alone (HR only), adding Prototype Predictive Alignment (HR + PPA), adding the training-only graph teacher (HR + Graph), and the full model with both graph and PPA. Accuracy (\%) is reported under 1-, 4-, and 16-shot settings on Aircraft, Caltech101, and EuroSAT. The graph teacher provides the dominant gain, while PPA is most effective when combined with graph-conditioned representations.
}
\label{tab:ppa_graph_ablation}
\vspace{0.3em}
\scriptsize
\setlength{\tabcolsep}{4.0pt}
\renewcommand{\arraystretch}{1.12}
\resizebox{\textwidth}{!}{%
\begin{tabular}{@{}lcc ccc ccc ccc@{}}
\toprule
\multirow{2}{*}{\textbf{Variant}}
& \multicolumn{2}{c}{\textbf{Enabled components}}
& \multicolumn{3}{c}{\textbf{Aircraft}}
& \multicolumn{3}{c}{\textbf{Caltech101}}
& \multicolumn{3}{c}{\textbf{EuroSAT}} \\
\cmidrule(lr){2-3}
\cmidrule(lr){4-6}
\cmidrule(lr){7-9}
\cmidrule(lr){10-12}
& \textbf{PPA} & \textbf{Graph}
& \textbf{1-shot} & \textbf{4-Shot} & \textbf{16-Shot}
& \textbf{1-shot} & \textbf{4-Shot} & \textbf{16-Shot}
& \textbf{1-shot} & \textbf{4-Shot} & \textbf{16-Shot} \\
\midrule

HR only
& \no & \no
& 32.2 & 43.5 & 59.2
& 92.7 & 93.0 & 95.1
& 74.6 & 88.3 & 91.8 \\

HR + PPA
& \yes & \no
& 31.7 & 43.4 & 59.5
& 92.5 & 93.0 & 95.3
& 73.9 & 86.4 & 92.1 \\

HR + Graph
& \no & \yes
& 33.6 & 45.1 & 60.2
& 94.2 & 96.1 & 97.1
& 76.1 & 89.9 & 93.2 \\

\rowcolor{oursbg}
\textbf{Full model}
& \yes & \yes
& \best{34.8} & \best{46.2} & \best{61.4}
& \best{94.4} & \best{96.2} & \best{97.2}
& \best{76.8} & \best{91.3} & \best{94.0} \\

\bottomrule
\end{tabular}%
}
\end{table*}

\paragraph{Quantifying the conditional role of PPA.}
Across the nine dataset--shot cells in Table~\ref{tab:ppa_graph_ablation} (Aircraft, Caltech101, and EuroSAT at 1/4/16 shots), PPA without graph-conditioned representations changes the HR-only baseline by $-0.29$ pp on average. The graph without PPA contributes $+1.68$ pp, and adding PPA on top of the graph contributes a further $+0.76$ pp; the full model improves over HR-only by approximately $+2.43$ pp. Thus, PPA is a complementary regularizer of graph-conditioned evidence, not a consistently beneficial standalone branch. The small difference between sums of rounded component gains and the rounded total reflects rounding.

\subsection{Effect of Token Filtering}
\label{app:token_filtering}

Table~\ref{tab:topk_pooling} evaluates the readout strategy of the training-only graph teacher. The results show that \emph{which} tokens are aggregated matters, not just how information propagates through the graph. Global pooling over all tokens consistently underperforms saliency-based Top-$M$ filtering, supporting our hypothesis that indiscriminate averaging mixes discriminative object regions with background or weakly informative tokens, weakening the structural signal transferred to the shared LoRA backbone.

A moderate number of high-saliency tokens works best. Top-$M{=}50$ yields the strongest and most stable performance across datasets and shot settings. On EuroSAT, Top-$M{=}50$ improves 4-shot accuracy from $89.5\%$ (global pooling) to $91.3\%$, and also performs best at 8- and 16-shot. Similar but smaller gains appear on Caltech101 and Food101, which are less affected by background clutter or strong texture ambiguity.

The ablation reveals a selection trade-off. Too few tokens (Top-$M{=}25$) remove useful context, while too many (Top-$M{=}75$) reintroduce low-saliency, noisy tokens. On EuroSAT, Top-$M{=}75$ drops to $88.8\%$ at 4-shot versus $91.3\%$ for Top-$M{=}50$. Overall, the graph teacher benefits from foreground-focused aggregation: it preserves discriminative local evidence while avoiding the dilution of global averaging. Because the readout is used only during training, this structural guidance is distilled into the deployed CLIP-LoRA backbone without adding inference-time overhead.
\begin{table}[h!]
\centering
\caption{
\textbf{Effect of token filtering in the graph-teacher readout.}
We compare global pooling over all graph-refined tokens with saliency-based Top-$M$ filtering. Top-$M$ retains only the highest-scoring patch tokens before aggregation, reducing the influence of background or weakly informative tokens. A moderate selection size, Top-$M{=}50$, provides the best overall trade-off across datasets and shot settings.
}
\label{tab:topk_pooling}
\setlength{\tabcolsep}{5pt} 
\begin{tabular}{llccccc}
\toprule
 & Pooling Strategy & 1-shot & 2-shot & 4-shot & 8-shot & 16-shot \\ 
\midrule

\multirow{4}{*}{\rotatebox{90}{\textbf{EuroSAT}}} 
 & $\sum_{\text{All}}$ (Global) & 75.9 & 80.1 & 89.5 & 92.2 & 93.6 \\
 & $\sum_{\text{Top-}M=25}$ & 77.0 & 81.2 & 90.0 & 92.6 & 93.9 \\
 & \cellcolor{oursbg}$\sum_{\text{Top-}M= 50}$ & \cellcolor{oursbg}\textbf{76.8} & \cellcolor{oursbg}\textbf{84.9} & \cellcolor{oursbg}\textbf{91.3} & \cellcolor{oursbg}\textbf{92.7} & \cellcolor{oursbg}\textbf{94.0} \\
 & $\sum_{\text{Top-}M= 75}$ & 76.7 & 81.0 & 88.8 & 92.4 & 93.2 \\ 
\midrule

\multirow{4}{*}{\rotatebox{90}{\textbf{Caltech}}} 
 & $\sum_{\text{All}}$ (Global) & 94.1 & 94.8 & 94.9 & 96.0 & 96.8 \\
 & $\sum_{\text{Top-}M= 25}$ & 94.2 & 95.0 & 95.0 & 96.2 & 96.9 \\
 & \cellcolor{oursbg}$\sum_{\text{Top-}M= 50}$ & \cellcolor{oursbg}\textbf{94.4} & \cellcolor{oursbg}\textbf{95.8} & \cellcolor{oursbg}\textbf{96.2} & \cellcolor{oursbg}\textbf{96.3} & \cellcolor{oursbg}\textbf{97.2} \\
 & $\sum_{\text{Top-}M= 75}$ & 94.0 & 94.8 & 95.3 & 96.0 & 97.0 \\ 
\midrule

\multirow{4}{*}{\rotatebox{90}{\textbf{Food}}} 
 & $\sum_{\text{All}}$ (Global) & 86.1 & 86.7 & 87.0 & 87.2 & 87.5 \\
 & $\sum_{\text{Top-}M= 25}$ & 86.0 & 86.8 & 86.9 & 87.3 & 87.6 \\
 & \cellcolor{oursbg}$\sum_{\text{Top-}M=50}$ & \cellcolor{oursbg}\textbf{86.2} & \cellcolor{oursbg}\textbf{86.8} & \cellcolor{oursbg}\textbf{87.1} & \cellcolor{oursbg}\textbf{87.4} & \cellcolor{oursbg}\textbf{87.8} \\
 & $\sum_{\text{Top-}M= 75}$ & 86.1 & 86.7 & 86.8 & 87.2 & 87.4 \\ 
\bottomrule
\end{tabular}
\end{table}

\subsection{Effect of Input Resolution}
\label{app:resolution_ablation}

Table~\ref{tab:resolution_ablation} analyzes the effect of input resolution on structure-aware adaptation. Because the graph teacher operates on ViT tokens, higher resolution yields a denser token grid and finer local cues for relational reasoning. Relative to the CLIP pretraining resolution of $224\times224$, $336\times336$ gives modest but non-uniform changes: some dataset--shot conditions improve while others decrease. The matched controls in Appx.~\ref{app:matched_resolution_cache} isolate this resolution effect from cache and structural supervision. At 16-shot, accuracy rises from $96.9\%$ to $97.2\%$ on Caltech101, from $93.9\%$ to $94.0\%$ on EuroSAT, and from $87.1\%$ to $87.8\%$ on Food101.

Further increasing the resolution to $512\times512$ consistently hurts performance, indicating that excessive upscaling mismatches the CLIP pretraining regime and destabilizes token-level interactions during graph propagation. The drop is most pronounced on Caltech101, where 16-shot accuracy falls from $97.2\%$ at $336\times336$ to $95.9\%$ at $512\times512$, and on Food101, where it falls from $87.8\%$ to $85.9\%$. Overall, $336\times336$ offers the best trade-off: it increases spatial granularity for the training-only graph teacher while preserving the stability of pretrained CLIP features.

\begin{table}[ht]
\centering
\caption{
\textbf{Effect of input resolution.}
We compare the standard CLIP input resolution, $224\times224$, with higher resolutions used for training-time graph reasoning. The $336\times336$ setting generally provides the best trade-off between finer token granularity and pretrained feature stability, while $512\times512$ introduces a stronger resolution shift and degrades performance.
}
\label{tab:resolution_ablation}
\setlength{\tabcolsep}{5pt}
\begin{tabular}{llccccc}
\toprule
 & Resolution ($H' \times W'$) & 1 Shot & 2 Shot & 4 Shot & 8 Shot & 16 Shot \\ 
\midrule

\multirow{3}{*}{\rotatebox{90}{\textbf{Caltech}}} 
 & $224 \times 224$ (Standard) & 94.1 & 95.6 & 95.4 & \textbf{96.8} & 96.9 \\
 & \cellcolor{oursbg}$336 \times 336$  & \cellcolor{oursbg}\textbf{94.4} & \cellcolor{oursbg}\textbf{95.8} & \cellcolor{oursbg}\textbf{96.2} & \cellcolor{oursbg}96.3 & \cellcolor{oursbg}\textbf{97.2} \\
 & $512 \times 512$ & 91.5 & 92.6 & 93.7 & 94.8 & 95.9 \\ 
\midrule

\multirow{3}{*}{\rotatebox{90}{\textbf{Euro}}} 
 & $224 \times 224$ (Standard) & 76.5 & \textbf{86.6} & \textbf{91.5} & 92.4 & 93.9 \\
 & \cellcolor{oursbg}$336 \times 336$  & \cellcolor{oursbg}\textbf{76.8} & \cellcolor{oursbg}84.9 & \cellcolor{oursbg}91.3 & \cellcolor{oursbg}\textbf{92.7} & \cellcolor{oursbg}\textbf{94.0} \\
 & $512 \times 512$ & 75.1 & 82.7 & 90.2 & 91.5 & 92.4 \\ 
\midrule

\multirow{3}{*}{\rotatebox{90}{\textbf{Food}}} 
 & $224 \times 224$ (Standard) & 86.0 & 86.5 & 86.7 & 87.1 & 87.1 \\
 & \cellcolor{oursbg}$336 \times 336$  & \cellcolor{oursbg}\textbf{86.2} & \cellcolor{oursbg}\textbf{86.8} & \cellcolor{oursbg}\textbf{87.1} & \cellcolor{oursbg}\textbf{87.4} & \cellcolor{oursbg}\textbf{87.8} \\
 & $512 \times 512$ & 84.3 & 85.1 & 85.3 & 85.6 & 85.9 \\ 
\bottomrule
\end{tabular}
\end{table}

%

\subsection{Matched Resolution, Cache, and Structural-Training Controls}
\label{app:matched_resolution_cache}
To separate the contributions of higher-resolution adaptation, cache-based prediction, and structural supervision, we evaluate six configurations on Aircraft, Caltech101, EuroSAT, and Food101 at 1/4/8/16 shots. Each cell is the unweighted mean across these four datasets, using identical support splits and three runs per configuration. Backbone, data partitions, support images, preprocessing, training schedule, and shared optimization settings are matched. The labels @224 and @336 denote adaptation resolution, not deployment resolution: all configurations use the original $224\times224$ inference encoder after adaptation. The SD-LoRA inference path retains the same cache as the corresponding LoRA+cache control.

\begin{center}
\begin{minipage}{\linewidth}
\centering
\captionsetup{hypcap=false}
\captionof{table}{\textbf{Full matched comparison of training resolution, cache, and structural supervision.} Top-1 accuracy (\%) is averaged over four datasets. The graph teacher and PPA head are absent from every deployed SD-LoRA model.}
\label{tab:sd_matched_full}
\small
\setlength{\tabcolsep}{9pt}
\begin{tabular}{lcccc}
\toprule
Training configuration & 1-shot & 4-shot & 8-shot & 16-shot \\
\midrule
CLIP-LoRA @224 & 70.1 & 75.2 & 78.5 & 81.9 \\
CLIP-LoRA @336 & 70.7 & 76.8 & 79.7 & 82.8 \\
CLIP-LoRA+cache @224 & 70.8 & 77.2 & 79.8 & 82.9 \\
CLIP-LoRA+cache @336 & 71.3 & 78.5 & 80.7 & 83.4 \\
SD-LoRA @224 & 72.6 & 79.3 & 81.9 & 84.8 \\
\rowcolor{oursrow}
\textbf{SD-LoRA @336} & \textbf{73.1} & \textbf{80.2} & \textbf{82.4} & \textbf{85.1} \\
\bottomrule
\end{tabular}
\end{minipage}
\end{center}
Increasing LoRA+cache training resolution from 224 to 336 changes accuracy by $+0.5/+1.3/+0.9/+0.5$ pp. At matched 336 resolution and cache, the training-only structural branch adds $+1.8/+1.7/+1.7/+1.7$ pp. At 336, adding the cache to vanilla CLIP-LoRA contributes $+0.6/+1.7/+1.0/+0.6$ pp. Finally, SD-LoRA trained at 224 exceeds the cache-equipped 336 control by $+1.3/+0.8/+1.2/+1.4$ pp. These controlled differences support an effect of structural training beyond either the cache or a denser token grid. They concern the four-dataset study above and should not be conflated with the 11-benchmark complete-system gains.

All 336-resolution variants use the same bicubic interpolation of the pretrained patch positional grid from $14\times14$ to $21\times21$, retaining the original \texttt{[CLS]} positional embedding. Higher-resolution adaptation provides more patch nodes for the teacher, rather than an additional inference path. Table~\ref{tab:resolution_ablation} further shows that resolution is not uniformly beneficial: 336 sometimes reduces accuracy, and 512 degrades performance in the evaluated settings. Thus, the matched controls isolate interpolation as well as resolution, without claiming that upscaling alone explains the structural-training gains.

\subsection{Graph Teacher Architecture Ablation}
\label{app:graph_architecture_ablation}

Table~\ref{tab:components} ablates the main components of the training-only graph teacher: the visual refinement Transformer (Trans), the relation-aware graph Transformer (RGT), and discriminative node filtering. The full teacher consistently performs best on Aircraft, Caltech101, and EuroSAT, showing that all three provide complementary structural supervision.

Removing Trans hurts performance, especially at higher shots on Aircraft and EuroSAT, indicating that early token refinement stabilizes local visual features before relation-aware message passing. Removing RGT causes a larger drop, confirming that explicit relation-aware propagation is central: without typed visual--visual and visual--text interactions, the model struggles to encode structured evidence. Removing discriminative node filtering also reduces accuracy, indicating that suppressing weak or background-dominated tokens improves the graph-level training signal.

Combined ablations reinforce this view: variants with only one or two components are consistently weaker than the full architecture, with the largest drops when RGT or filtering is missing. Thus, SD-LoRA gains from the interaction of token refinement, relation-aware reasoning, and foreground-focused readout, rather than any single addition. Because the graph teacher is used only during adaptation, these designs strengthen supervision of the shared LoRA backbone without increasing inference-time complexity.

\begin{table*}[t]
\centering
\scriptsize
\setlength{\tabcolsep}{3.6pt}
\renewcommand{\arraystretch}{1.08}
\caption{
\textbf{Graph teacher architecture ablation.}
We evaluate the contributions of the visual refinement Transformer (Trans), the relation-aware graph Transformer (RGT), and discriminative node filtering. Accuracy (\%) is reported on Aircraft, Caltech101, and EuroSAT under 1-, 4-, and 16-shot settings. The full architecture performs best across all datasets and shot settings, showing that token refinement, relation-aware message passing, and foreground-focused filtering provide complementary structural supervision.
}
\label{tab:components}
\resizebox{\textwidth}{!}{
\begin{tabular}{ccc ccc ccc ccc}
\toprule
\multicolumn{3}{c}{\textbf{Components}}
& \multicolumn{3}{c}{\textbf{Aircraft}}
& \multicolumn{3}{c}{\textbf{Caltech101}}
& \multicolumn{3}{c}{\textbf{EuroSAT}} \\
\cmidrule(lr){1-3}
\cmidrule(lr){4-6}
\cmidrule(lr){7-9}
\cmidrule(lr){10-12}
\textbf{Trans} & \textbf{RGT} & \textbf{Filtering}
& \textbf{1-shot} & \textbf{4-shot} & \textbf{16-shot}
& \textbf{1-shot} & \textbf{4-shot} & \textbf{16-shot}
& \textbf{1-shot} & \textbf{4-shot} & \textbf{16-shot} \\
\midrule
\no  & \yes & \yes & 34.1 & 45.1 & 58.8 & 93.2 & 94.8 & 95.4 & 74.2 & 89.8 & 91.1 \\
\yes & \no  & \yes & 33.5 & 45.3 & 57.6 & 92.8 & 93.7 & 95.5 & 73.1 & 88.6 & 89.8 \\
\yes & \yes & \no  & 33.4 & 44.2 & 57.4 & 92.4 & 94.2 & 95.7 & 73.7 & 89.1 & 91.6 \\
\no  & \yes & \no  & 32.7 & 44.9 & 56.5 & 91.5 & 93.3 & 94.5 & 72.8 & 88.2 & 90.8 \\
\yes & \no  & \no  & 33.2 & 44.0 & 55.3 & 91.5 & 92.5 & 93.3 & 72.1 & 87.5 & 90.0 \\
\no  & \no  & \yes & 32.7 & 41.7 & 54.2 & 89.7 & 91.8 & 92.3 & 72.3 & 86.8 & 89.3 \\
\rowcolor{oursbg}
\yes & \yes & \yes
& \textbf{34.8} & \textbf{46.2} & \textbf{61.4}
& \textbf{94.4} & \textbf{96.2} & \textbf{97.2}
& \textbf{76.8} & \textbf{91.3} & \textbf{94.0} \\
\bottomrule
\end{tabular}
}
\end{table*}

\subsection{Loss Component Ablation}
\label{app:loss_ablation}

Table~\ref{tab:loss_ablation_final} evaluates the impact of the three training losses: deployed student loss $\mathcal{L}_{\mathrm{Student}}$, graph-teacher loss $\mathcal{L}_{\mathrm{Teacher}}$, and Prototype Predictive Alignment $\mathcal{L}_{\mathrm{PPA}}$. The student-only objective is a strong baseline because it directly optimizes the LoRA$+$cache predictor used at inference, but its supervision operates mainly at the image-logit level and does not explicitly enforce token-level structure.

Adding graph-teacher loss improves performance in several cases (e.g., Caltech101 and Food101), showing that structure-aware supervision benefits the shared LoRA backbone. The conditional contribution of PPA is quantified separately in Appx.~\ref{app:ppa_graph_ablation}; it is most effective as a complement to graph-conditioned supervision.

Prototype Predictive Alignment offers a more stable complementary signal. Relative to the student-only objective, adding $\mathcal{L}_{\mathrm{PPA}}$ improves EuroSAT from $93.4\%$ to $93.8\%$ at 8-shot, Caltech101 from $93.9\%$ to $95.5\%$ at 4-shot, and Food101 from $86.9\%$ to $87.6\%$ at 8-shot. This indicates that prototype-level supervision better anchors the adapted representation to the support-induced class structure, beyond individual instance predictions.

The full objective, combining $\mathcal{L}_{\mathrm{Student}}$, $\mathcal{L}_{\mathrm{Teacher}}$, and $\mathcal{L}_{\mathrm{PPA}}$, yields the best performance in all settings. This validates implicit structural distillation in SD-LoRA: the graph teacher provides relation-aware signals, and PPA stabilizes them by enforcing prototype-consistent geometry in the shared CLIP-LoRA feature space. Because all teacher components are discarded after training, the gains fully transfer to the deployed LoRA$+$cache student without logit matching or test-time graph computation.
\begin{table*}[htbp]
\centering
\scriptsize
\setlength{\tabcolsep}{4.0pt}
\renewcommand{\arraystretch}{1.08}
\caption{
\textbf{Loss component ablation.}
We evaluate the contributions of the student loss $\mathcal{L}_{\mathrm{Student}}$, the graph-teacher loss $\mathcal{L}_{\mathrm{Teacher}}$, and Prototype Predictive Alignment $\mathcal{L}_{\mathrm{PPA}}$. Accuracy (\%) is reported on EuroSAT, Caltech101, and Food101 under 1-, 4-, and 8-shot settings. The full objective performs best across all settings, showing that graph-based structural supervision and prototype-level alignment provide complementary training signals.
}
\label{tab:loss_ablation_final}
\resizebox{\textwidth}{!}{
\begin{tabular}{ccc ccc ccc ccc}
\toprule
\multicolumn{3}{c}{\textbf{Loss Components}}
& \multicolumn{3}{c}{\textbf{EuroSAT}}
& \multicolumn{3}{c}{\textbf{Caltech101}}
& \multicolumn{3}{c}{\textbf{Food101}} \\
\cmidrule(lr){1-3}
\cmidrule(lr){4-6}
\cmidrule(lr){7-9}
\cmidrule(lr){10-12}
$\mathcal{L}_{\text{Student}}$
& $\mathcal{L}_{\text{Teacher}}$
& $\mathcal{L}_{\text{PPA}}$
& \textbf{1-shot} & \textbf{4-shot} & \textbf{8-shot}
& \textbf{1-shot} & \textbf{4-shot} & \textbf{8-shot}
& \textbf{1-shot} & \textbf{4-shot} & \textbf{8-shot} \\
\midrule
\yes & \no  & \no  & 74.6 & 88.3 & 91.8 & 92.7 & 93.0 & 95.1 & 85.2 & 86.3 & 86.9 \\
\yes & \yes & \no  & 76.2 & 89.7 & 93.5 & 94.2 & 94.9 & 96.8 & 85.7 & 86.7 & 87.2 \\
\yes & \no  & \yes & 76.4 & 91.0 & 93.8 & 94.3 & 95.5 & 97.0 & 86.0 & 86.9 & 87.6 \\
\rowcolor{oursbg}
\yes & \yes & \yes
& \textbf{76.8} & \textbf{91.3} & \textbf{94.0}
& \textbf{94.4} & \textbf{96.2} & \textbf{97.2}
& \textbf{86.2} & \textbf{87.1} & \textbf{87.8} \\
\bottomrule
\end{tabular}
}
\end{table*}

\subsection{Cross-modal Interaction Frequency}
\begin{figure}
    \centering
    \includegraphics[width=0.8\linewidth]{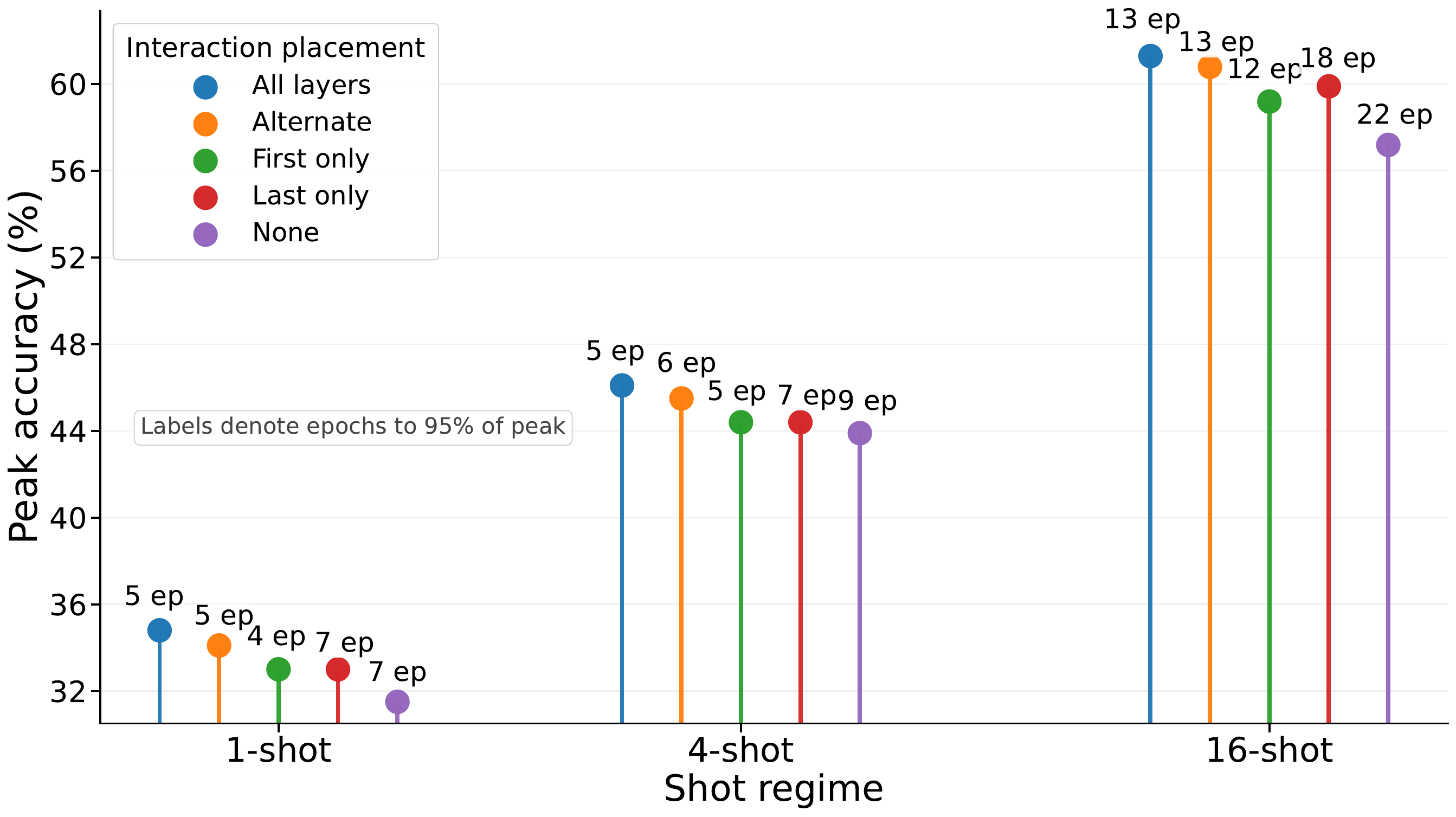}
    \caption{
\textbf{Effect of cross-modal interaction frequency.}
We ablate where semantic text--visual interaction is applied across graph layers on Aircraft under 1/4/16-shot settings. Each marker reports the peak accuracy, while the text label indicates the number of epochs required to reach 95\% of the peak. Applying interaction in \emph{all} layers consistently achieves the best peak accuracy across shot regimes and also converges efficiently in the 16-shot setting, reaching 95\% of peak in 13 epochs compared with 18 epochs for \emph{last only} and 22 epochs for \emph{none}. This suggests that frequent cross-modal interaction improves both final adaptation quality and optimization efficiency.
}
    \label{fig:cross-modal-frequency}
\end{figure}

\paragraph{Why cross-modal interaction frequency matters?}
The SD-LoRA graph teacher refines local visual evidence using both visual structure and class semantics. 
\emph{Interaction placement} specifies at which message-passing layers visual–text edges are active: \emph{all layers} activates them at every layer; \emph{alternate} at alternating layers; \emph{first only} only in the first layer; \emph{last only} only in the final layer; and \emph{none} removes visual–text interaction, leaving purely visual message passing. 
Figure~\ref{fig:cross-modal-frequency} reports Aircraft performance under 1/4/16-shot settings, where subtle local evidence is crucial. 
Marker height gives the best accuracy for each strategy; the label above each marker shows the epochs needed to reach 95\% of that peak, separating final performance from optimization speed.

Applying text–visual interaction in \emph{all layers} is consistently best across all shot counts, indicating that gains come from repeated semantic–visual exchange during message passing, not just adding text nodes. 
\emph{First-only} interaction injects semantics before visual features are well refined, while \emph{last-only} waits too long to ground them. 
The \emph{none} variant performs worst and converges slowest in the 16-shot case, showing that visual-only structure is inadequate for this fine-grained task. 
Thus, the ablation supports SD-LoRA’s central choice: frequent class-conditioned interaction during training yields a stronger LoRA, even though the graph is discarded at inference.

\section{Computational Complexity Analysis and Hyperparameter Optimization}
\label{app:compute_complex}
Table~\ref{tab:compute_analysis} separates the cost of high-resolution structural training from that of the graph branch. SD-LoRA uses a 336$\times$336 CLIP forward pass only during adaptation, letting the training-only teacher see a denser ViT token grid and reason over finer local evidence. This addresses few-shot fine-grained tasks, where discriminative cues often come from small parts or textures that may vanish when adaptation relies only on a standard 224$\times$224 global representation. Instead of multiple manual crops or repeated region encodings, SD-LoRA obtains richer spatial evidence with a single high-resolution ViT-B/16 pass, then applies graph reasoning over the resulting patch tokens.

The main extra training-time cost is high-resolution token extraction, which increases the backbone from 17.50 GFLOPs at 224$\times$224 to 46.73 GFLOPs at 336$\times$336. On top of this, the graph reasoning teacher adds only 0.61 GFLOPs (1.31\% of the 336$\times$336 CLIP pass), so the relation-aware graph module remains lightweight; the dominant cost is the one-time extraction of denser tokens during adaptation. This matters because the graph branch is never deployed; it serves only as a training-time source of token-level structural supervision for the shared LoRA parameters.

\begin{figure}
    \centering
    \includegraphics[width=\linewidth]{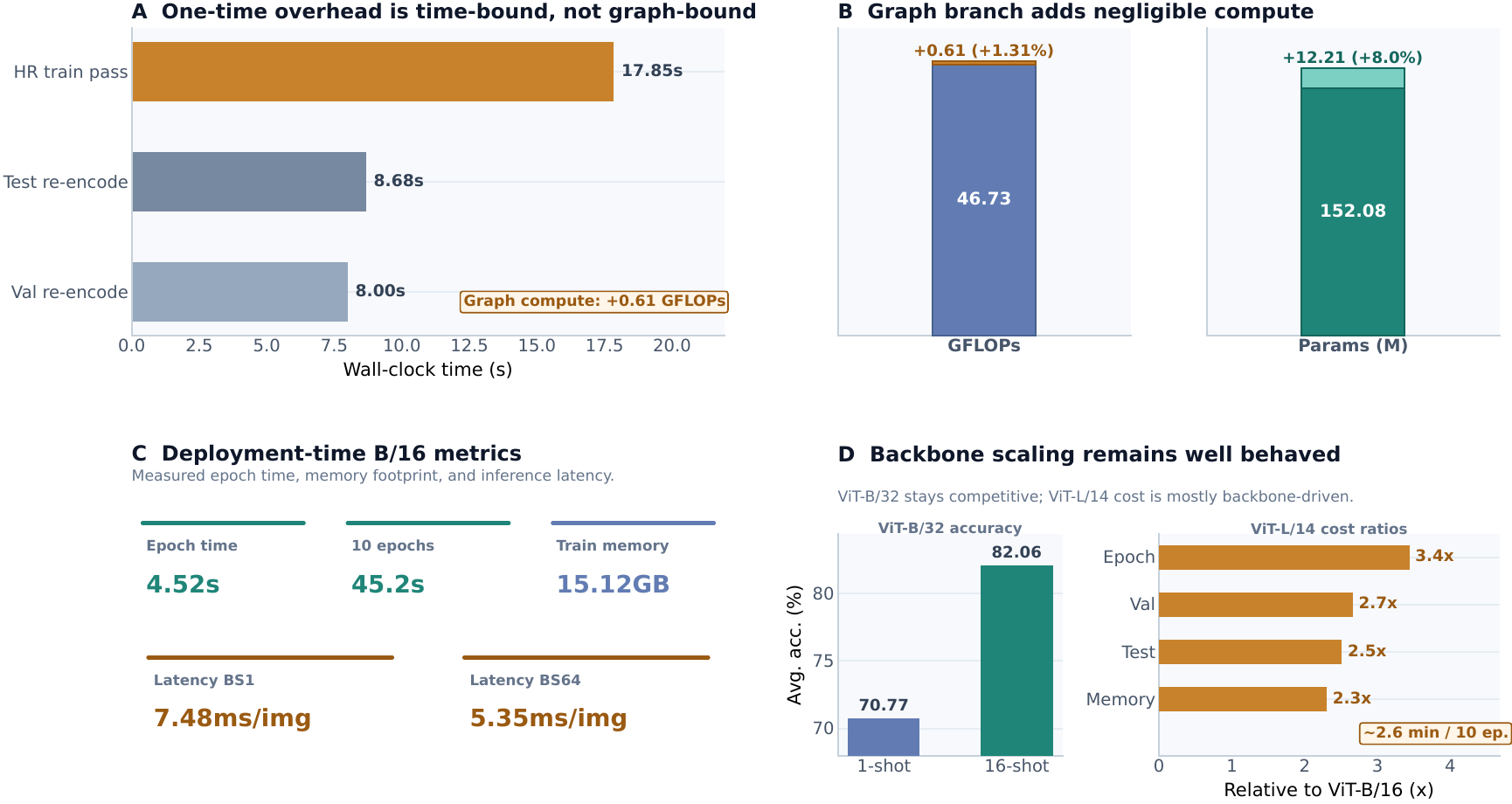}
\caption{\textbf{Computational and deployment analysis of our method.}
\textbf{A} shows the one-time wall-clock overhead associated with the additional high-resolution training and re-encoding passes.
\textbf{B} reports the computational overhead relative to the CLIP-LoRA training system, showing that the added graph branch increases computation by only $0.61$ GFLOPs ($+1.31\%$) while keeping parameter growth modest ($+8.0\%$).
\textbf{C} summarizes practical deployment-time metrics on ViT-B/16, including training time, memory footprint, and inference latency.
\textbf{D} shows that the method remains well behaved across backbone choices: ViT-B/32 retains competitive average few-shot accuracy, while the higher cost of ViT-L/14 is primarily attributable to the larger backbone rather than to additional method-specific steps.}

    \label{fig:efficiency_overview}
\end{figure}

\begin{table}[htbp]
    \centering
    \scriptsize
    \caption{
    \textbf{Computational analysis.}
    GFLOPs are reported per image using the same profiler. 
    SD-LoRA separates training-time structural supervision from deployment. 
    The 336$\times$336 pass is used only during adaptation to expose a denser ViT token grid for fine-grained graph supervision. 
    The graph branch itself adds 0.61 GFLOPs on top of this high-resolution pass, corresponding to 1.31\% overhead relative to the 336$\times$336 backbone pass. 
    At inference, the graph reasoning module, saliency scorer, PPA head, and high-resolution teacher path are removed, reducing deployment to the 224$\times$224 CLIP-LoRA+cache student.
    }
    \label{tab:compute_analysis}
    \vspace{0.2cm}
    \scriptsize
    \setlength{\tabcolsep}{3.8pt}
    \renewcommand{\arraystretch}{1.12}
    \resizebox{\linewidth}{!}{%
    \begin{tabular}{l c c c c}
        \toprule
        \textbf{Component / Setting} 
        & \textbf{Params (M)} 
        & \textbf{Trainable (M)} 
        & \textbf{GFLOPs} 
        & \textbf{Role} \\
        \midrule
        \textit{Standard adaptation reference} & & & & \\
        CLIP-LoRA, 224$\times$224 
        & 152.08 
        & 1.23 
        & 17.50 
        & Standard train/infer path \\
        \midrule
        \textit{SD-LoRA training path} & & & & \\
        Frozen CLIP backbone, 336$\times$336 
        & 150.85 
        & 0.00 
        & 46.73 
        & HR token extraction \\
        \quad + LoRA 
        & 1.23 
        & 1.23 
        & $\approx$0.00 
        & Shared deployed update \\
        Cache head 
        & 0.05 
        & 0.05 
        & $\approx$0.00 
        & Deployed classifier \\
        Graph reasoning teacher 
        & 12.16 
        & 12.16 
        & 0.61 
        & Training only \\
        \quad -- Token projection 
        & \textit{2.10} 
        & \textit{2.10} 
        & \textit{0.47} 
        & \textit{Training only} \\
        \quad -- Relation-aware graph layers 
        & \textit{6.89} 
        & \textit{6.89} 
        & \textit{0.14} 
        & \textit{Training only} \\
        \quad -- Saliency/readout/projection heads 
        & \textit{3.17} 
        & \textit{3.17} 
        & \textit{$<$0.01} 
        & \textit{Training only} \\
        \midrule
        \textbf{Total SD-LoRA training system} 
        & \textbf{164.29} 
        & \textbf{13.44} 
        & \textbf{47.34} 
        & \textbf{Adaptation only} \\
        Graph-only overhead vs. 336$\times$336 CLIP 
        & -- 
        & -- 
        & \textbf{+1.31\%} 
        & Teacher branch cost \\
        Full training cost vs. 224$\times$224 CLIP-LoRA 
        & -- 
        & -- 
        & \textbf{2.71$\times$} 
        & HR training cost \\
        \midrule
        \textbf{Deployed SD-LoRA inference path} 
        & \textbf{152.13} 
        & \textbf{1.28 retained} 
        & \textbf{17.50 + cache} 
        & \textbf{No graph / no HR teacher} \\
        \bottomrule
    \end{tabular}%
    }
\end{table}

The full training system has 164.29M parameters, with 13.44M trainable. Most trainable parameters lie in the training-only graph teacher (12.16M), while the deployed LoRA and cache use just 1.28M trainable parameters. After adaptation, the high-resolution teacher path, graph module, saliency scorer, and PPA head are discarded. Inference uses only the standard 224$\times$224 CLIP-LoRA encoder plus the cache head, with no graph construction or high-resolution computation. SD-LoRA thus trades a higher one-time adaptation cost for fine-grained structural supervision in the deployable LoRA backbone, while keeping inference lightweight for practical use.



Table~\ref{tab:search_space} summarizes the SD-LoRA hyperparameter search space, showing that adaptation uses a compact set of interpretable knobs. Most parameters follow standard optimization and PEFT practice (learning rate, weight decay, and a LoRA rank candidate set $r\in\{2,4,8,16\}$ with $\alpha_{\text{LoRA}}\in\{8,16,32\}$). The remaining freedom lies in balancing three training signals: graph supervision ($\lambda_1$), prototype alignment ($\lambda_2$), and hard-example emphasis via focal loss ($\gamma$) plus a single inference-time cache weight ($\omega$). The ranges are chosen for stability: $\lambda_2\in[0.1,1.0]$ regularizes rather than dominates learning, while $\lambda_1$ spans a wider range to accommodate dataset granularity and structural cue strength. Overall, the search is tractable (few hyperparameters, narrow PEFT range) and cleanly decouples training-time structural guidance from deployment, since only $\omega$ affects inference.

\begin{table}[ht!]
    \centering
    \caption{\textbf{Hyperparameter Optimization Space.} The search selects one global configuration on validation data; selected values are reported in Table~\ref{tab:sd_selected_settings} and are fixed across datasets and shots.}
    \label{tab:search_space}
    \begin{tabular}{l c c c}
        \toprule
        \textbf{Hyperparameter} & \textbf{Symbol} & \textbf{Search Range} & \textbf{Scale} \\
        \midrule
        \multicolumn{4}{l}{\textit{Optimizer \& Backbone}} \\
        Learning Rate & $\eta$ & $[10^{-5}, 5 \times 10^{-2}]$ & Log-Uniform \\
        Weight Decay & $wd$ & $[10^{-6}, 10^{-3}]$ & Log-Uniform \\
        LoRA Rank & $r$ & $[2, 4, 8, 16]$ & Integer \\
        LoRA Alpha & $\alpha_{\text{LoRA}}$ & $\{8,16,32\}$ & Discrete \\
        \midrule
        \multicolumn{4}{l}{\textit{Training Objectives}} \\
        Graph Loss Weight & $\lambda_1$ & $[0.1, 5.0]$ & Uniform \\
        PPA Alignment Weight & $\lambda_2$ & $[0.1, 1.0]$ & Uniform \\
        Focal Loss Gamma & $\gamma$ & $[0.5, 3.0]$ & Uniform \\
        \midrule
        \multicolumn{4}{l}{\textit{Inference}} \\
        Inference Cache Weight & $\omega$ & $[1.0, 10.0]$ & Uniform \\
        \bottomrule
    \end{tabular}
\end{table}

\subsection{Selected Configuration and Model-Selection Protocol}
\label{app:sd_selected_settings}
The search space in Table~\ref{tab:search_space} is distinct from the selected configuration in Table~\ref{tab:sd_selected_settings}. Bayesian optimization uses mean validation top-1 accuracy over a predefined set of six representative datasets and all shot regimes to choose one global configuration. That configuration is then used unchanged across all 11 datasets and shot settings; there is no per-dataset or per-shot hyperparameter selection. The wide learning-rate search interval does not imply that the reported models use large learning rates.

\begin{center}
\begin{minipage}{\linewidth}
\centering
\captionsetup{hypcap=false}
\captionof{table}{\textbf{Selected SD-LoRA hyperparameters.} These values are fixed globally after validation selection. Search ranges describe the selection procedure, not uncertainty in the final evaluation settings.}
\label{tab:sd_selected_settings}
\small
\setlength{\tabcolsep}{7pt}
\begin{tabular}{lcc}
\toprule
Hyperparameter & Selected value & Candidates / search range \\
\midrule
Shared LoRA learning rate & $1\times10^{-5}$ & $[10^{-5},5\times10^{-2}]$, log-uniform \\
Teacher/head learning rate & $1\times10^{-5}$ & $[10^{-5},5\times10^{-2}]$, log-uniform \\
Weight decay & $1\times10^{-4}$ & $[10^{-6},10^{-3}]$, log-uniform \\
LoRA rank $r$ & 4 & $\{2,4,8,16\}$ \\
LoRA alpha $\alpha_{\rm LoRA}$ & 8 & $\{8,16,32\}$ \\
Graph-loss weight $\lambda_1$ & 1.0 & $[0.1,5.0]$, uniform \\
PPA weight $\lambda_2$ & 0.5 & $[0.1,1.0]$, uniform \\
Focal-loss parameter $\gamma$ & 2.0 & $[0.5,3.0]$, uniform \\
Cache weight $\omega$ & 5.0 & $[1.0,10.0]$, uniform \\
\bottomrule
\end{tabular}
\end{minipage}
\end{center}
Support images are sampled only from training partitions. Model selection uses validation data, and test labels are not used for optimization or selection. Controlled comparisons, including CLIP-LoRA, CLIP-LoRA+cache, and SD-LoRA, share the same support samples, partitions, preprocessing, training schedules, validation criterion, tuning budget, and fixed shared hyperparameters; only the component being tested changes. Each shot setting uses three independently sampled support sets, with each controlled variant adapted separately on the same sets.

Published external baselines are taken from their original papers, rather than rerun or retuned. They provide literature-level comparisons under the reported benchmark protocols; we do not claim identical optimization settings or an equal search budget for those external results. The matched controls, rather than the full literature table alone, support attribution to structural training.

We report main few-shot means and SDs over the three support-split runs. The matched-control and deployed-feature tables report the aggregate values available for those studies; they do not imply new significance tests or per-condition confidence intervals. Protocol-specific differences in support-label access, validation access, and inference-time adaptation must be considered when interpreting comparisons.

\paragraph{Reproducibility and statistical reporting.}
For the main few-shot experiments, each result is the mean accuracy over three independently sampled support splits. Uncertainty values are the standard deviation across these runs, reflecting variation from support-set choice and stochastic optimization under a fixed evaluation protocol. Aggregate results are unweighted arithmetic means over datasets. For base-to-novel, cross-dataset, and ImageNet-OOD evaluations, we follow the standard fixed protocols from prior CLIP adaptation work and report point estimates unless noted otherwise. Hyperparameters are chosen using only training/validation data or support-split cross-validation; test labels are never used for model selection.


\section{Scope, Limitations, and Responsible Use}
\label{app:limitations}

SD-LoRA targets inductive few-shot adaptation of CLIP-like vision–language models with a small labeled support set and lightweight inference. It is best suited to settings where deployment must be efficient but training-time cost can be moderately higher. Its key benefit is a relation-aware graph teacher used only during adaptation and discarded afterward, improving the deployed LoRA$+$cache student without adding graph computation at test time; this also limits its applicability.

\paragraph{Training-time overhead.}
Despite no graph cost at inference, SD-LoRA is more expensive to adapt than standard CLIP-LoRA or cache-only methods because the teacher processes high-resolution visual tokens and class-text nodes. This is acceptable for offline few-shot adaptation on a small support set, but less suitable for frequent on-device updates or rapid continual adaptation.

\paragraph{Dependence on support-set quality.}
SD-LoRA relies on a high-quality few-shot support set. If labels are noisy, imbalanced, or visually unrepresentative, both the cache and the prototype-alignment objective can encode misleading class structure. The graph teacher refines local evidence but cannot compensate for unrepresentative support examples.

\paragraph{Backbone and modality scope.}
We study CLIP-style image--text models on standard few-shot benchmarks. The structural distillation idea is general, but our implementation relies on ViT token representations and class-text embeddings. Applying it to dense prediction, video, open-vocabulary detection, or multimodal temporal reasoning may require new graph constructions and task-specific readouts.

\paragraph{Resolution and graph design sensitivity.}
The Appendix shows that moderately higher-resolution tokenization helps, but excessively high resolution hurts due to mismatch with CLIP pretraining. Graph edge choices and token filtering also influence performance. Thus, structural supervision is beneficial but constrained: effective graph design must balance local detail, semantic grounding, and pretrained feature stability.

\paragraph{Responsible use.}
SD-LoRA improves adaptation efficiency and transferability, aiding applications with limited labeled data. Like other CLIP-based methods, it inherits biases and failure modes from the pretrained vision--language model and the support data. In sensitive domains (e.g., medical, forensic, public safety), SD-LoRA must not serve as an autonomous decision-maker; it should be evaluated with domain-specific protocols, carefully calibrated, and used only with human oversight and safeguards.

\paragraph{Existing assets, licenses, and data access.}
SD-LoRA relies on public benchmark datasets and pretrained CLIP-style vision--language models. We cite the original sources for all datasets, pretrained models, and baselines. We do not redistribute datasets or model weights; users must obtain them from official sources and follow the respective licenses or terms of use. Any released implementation will provide instructions for acquiring or preparing public datasets but will not repackage restricted or third-party assets.

\clearpage



\end{document}